\documentclass[11pt]{article}

\usepackage[preprint]{acl}

\usepackage{times}
\usepackage{latexsym}

\usepackage[T1]{fontenc}

\usepackage[utf8]{inputenc}

\usepackage{microtype}

\usepackage{inconsolata}

\usepackage{graphicx}

\usepackage[utf8]{inputenc} 
\usepackage[T1]{fontenc}    
\usepackage{hyperref}       
\usepackage{url}            
\usepackage{booktabs}       
\usepackage{amsfonts}       
\usepackage{nicefrac}       
\usepackage{microtype}      
\usepackage{xcolor}         
\usepackage{booktabs}
\usepackage{tabularx}
\usepackage{amsmath}
\usepackage{amssymb}
\usepackage{booktabs}
\usepackage[table]{xcolor}
\usepackage{multirow}
\usepackage{rotating}
\usepackage{caption}
\usepackage[utf8]{inputenc}
\usepackage{geometry}
\usepackage{multicol}
\usepackage{booktabs}
\usepackage{array}
\usepackage{caption}
\usepackage{float}
\usepackage{graphicx}
\usepackage{longtable}   
\usepackage{array}     
\usepackage{siunitx}   
\usepackage{rotating}
\usepackage{pdflscape}
\usepackage{makecell}
\usepackage{amsmath}
\usepackage{titletoc}
\usepackage{minitoc}
\usepackage{tocloft}
\usepackage{etoolbox}
\usepackage{hyperref}
\usepackage{comment}
\usepackage{todonotes}
\usepackage{amsmath}

\title{Generative Personality Simulation via Theory-Informed Structured Interview}

\author{
    Pengda Wang$^{1}$, Huiqi Zou$^{3}$, Han Jiang$^{4}$, \\
    \textbf{Hanjie Chen$^{2}$, Tianjun Sun$^{1}$$^{*}$, Xiaoyuan Yi$^{5}$, Ziang Xiao$^{4}$$^{*}$, and Frederick L. Oswald$^{1}$$^{*}$} \\
    $^1$Department of Psychological Sciences \& $^2$Department of Computer Science, Rice University \\
    $^3$Department of Electrical and Computer Engineering, Northeastern University \\ 
    $^4$Department of Computer Science, Johns Hopkins University \\
    $^5$Microsoft Research Asia\\ 
    \texttt{\{pw32,ts110,fo3\}@rice.edu;}
    \texttt{\{ziang.xiao\}@jhu.edu}
    \\ 
}

\begin{document}
\maketitle
\begin{abstract}
Despite their potential as human proxies, LLMs often fail to generate heterogeneous data with human-like diversity, thereby diminishing their value in advancing social science research.
To address this gap, we propose a novel method to incorporate psychological insights into LLM simulation through the Personality Structured Interview (PSI\footnote{Code and data: \url{https://github.com/isle-dev/PSI}.}). 
PSI leverages psychometric scale-development procedures to capture personality-related linguistic information from a formal psychological perspective.
To systematically evaluate simulation fidelity, we developed a measurement theory grounded evaluation procedure that considers the latent construct nature of personality and evaluates its reliability, structural validity, and external validity.
Results from three experiments demonstrate that PSI effectively improves human-like heterogeneity in LLM-simulated personality data and predicts personality-related behavioral outcomes.
We further offer a theoretical framework for designing theory-informed structured interviews to enhance the reliability and effectiveness of LLMs in simulating human-like data for broader psychometric research. 
\end{abstract}

\def\thefootnote{*}\makeatletter\def\Hy@Warning#1{}\makeatother\footnotetext{Co-corresponding authors.}

\def\thefootnote{\arabic{footnote}}

\section{Introduction}

The discipline of personality psychology seeks to understand how individual differences shape significant life outcomes and trajectories.
Decades of empirical work have demonstrated that personality traits predict a wide range of significant domains, including career success, mental and physical health, interpersonal relationships, and overall well-being (e.g.,~\citealp{judge2002personality, roberts2007power, robins2002s}).
Therefore, personality assessment has often been used to inform personnel selection and clinical interventions.
Moreover, the relevance of personality research is rapidly expanding into the field of Artificial Intelligence (AI), where it informs the design of more adaptive, human-centered, and personalized systems.


\begin{figure*}
    \centering
        \includegraphics[width=1.0\linewidth]{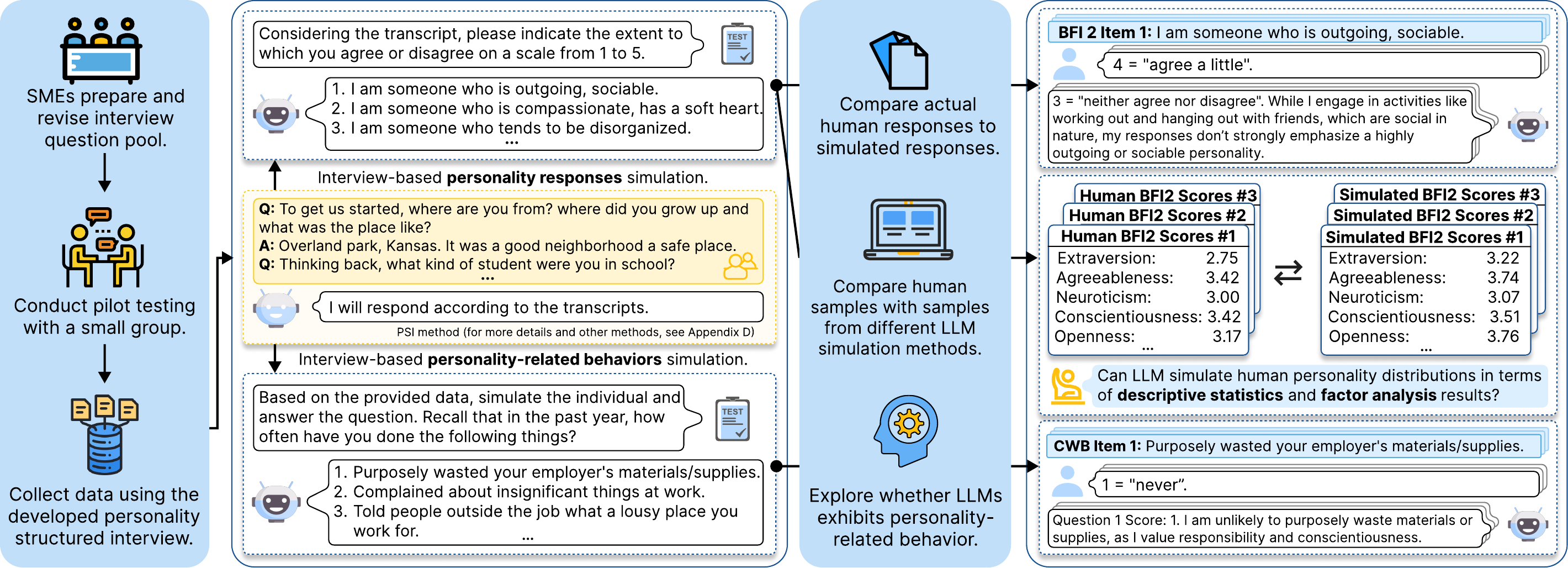}
    \caption{Overview of the development of the personality structured interview and experimental implementation.}
\label{fig:overview}
\vspace{-4mm}
\end{figure*}



However, studying personality requires large amounts of psychological data, such as personality trait scores and behavioral indicators. 
These data collection demands pose substantial logistical and financial challenges for researchers.
Moreover, certain types of data are particularly difficult to obtain. 
For example, studies focusing on individuals with elevated levels of psychopathy or narcissism often struggle to obtain adequate sample sizes due to the rarity and sensitivity of these traits (e.g.,~\citealp{lynam2001using}). 
Other research questions require longitudinal designs to capture personality development or dynamic patterns of social interaction over time, further compounding the complexity and resource demands (e.g.,~\citealp{damian2019sixteen,roberts2006patterns}). 
If a method could accurately simulate human personality distributions, it would accelerate personality research by offering a more scalable and cost-effective way to support data collection and experimentation~\citep{messeri2024artificial}.

In this context, the current work investigates the potential of LLMs to simulate human-like responses in psychometric research. 
Although prior studies have shown such promise, indicating that LLMs can generate responses for personality scales that reflect personality traits resembling those of humans (e.g., \citealp{huang2023revisiting}, \citealp{lee2024llms}), those approaches exhibit notable limitations, particularly in their ability to capture individual differences at the item-level and to reproduce the \textbf{heterogeneity} observed in human responses (e.g., \citealp{wang2024not}).
Item-level data provides a more granular understanding of how traits manifest in specific individuals rather than relying solely on broad trait averages.
Furthermore, modeling response heterogeneity is crucial for reflecting the variability and complexity of real-world human behavior.
Additionally, current evaluation practices are often misaligned with the standards of psychological research. 
For example, prior research has predominantly assessed personality simulation data at the aggregated level, neglecting item- and facet-level analyses as well as the psychometric properties of the data. 
The evaluation gap undermines psychologists’ confidence in the validity of such methods.

To address these research gaps, we investigated the use of \underline{P}ersonality \underline{S}tructured \underline{I}nterview (\textbf{PSI}) transcripts as a means of guiding LLMs in simulating personality data (see Figure~\ref{fig:overview}). 
The PSI comprises theory-driven questions paired with responses, designed to elicit narrative-based information pertinent to the personality constructs targeted in LLM simulations. 
This theory-informed approach enables the expression of personality in a manner that enhances the human-like heterogeneity and representativeness of the simulated data.
To bridge the evaluation gap, we developed a comprehensive framework for assessing personality simulation data grounded in measurement theory (e.g.,~\citealp{cronbach1951coefficient, cronbach1955construct}). 
This evaluation framework explicitly accounts for the hierarchical structure of personality measurement, wherein observed responses at the item-level map onto latent traits at the domain-level. 
It further supports a wide range of analytic strategies, from basic descriptive statistics to more advanced assessments of psychometric validity and reliability.

With this framework, we conducted three experiments to evaluate the efficacy of the PSI method in: (1) replicating individual-level personality data, (2) simulating distributions of personality approximate those observed in human populations, and (3) capturing personality-related behaviors aligned with established findings. 
Our findings demonstrate that the PSI approach enhances human-like variability of simulated personality data and effectively captures the patterns of personality-relevant behavior.

In summary, our paper makes three key contributions: (1) a theory-informed LLM-based simulation method, PSI, for personality research, along with a framework for developing such interview protocols; (2) the release of a dataset comprising 357 structured interview transcripts; and (3) an evaluation framework for LLM-simulated psychometric data, grounded in measurement theory.
We further discuss the potential of the PSI method and the dataset to advance AI and psychology research (see Appendix~\ref{sec:appendix_psychometric}). 
The interview question development framework underlying PSI can be generalized to capture information relevant to other constructs (e.g., value), thereby facilitating the development of more theory-aligned data simulations and enhancing the alignment of LLMs with humans.

\section{Related Works}
\vspace{-1mm}
\paragraph{Personality Definition and Construct}
\citet{mackinnon1944structure} proposed two complementary definitions of personality.
One emphasizes internal factors like temperament and interpersonal strategies that drive consistent behavior across time, situations, and cultures. 
The other focuses on interpersonal characteristics as perceived by others, linking personality to reputation. 
The former highlights internal drives, while the latter centers on external behaviors and social perception. 
Together, these perspectives underscore personality's role in shaping thought patterns and behaviors in social interactions (e.g., \citealp{hogan1996personality}).

Personality encodes rich and complex information in language and text (e.g., \citealp{goldberg1990alternative, saucier2001lexical}). 
The Five Factor Model (FFM) of personality is extensively researched (e.g.,~\citealp{costa2008revised,john1999big,mccrae1997personality}); it is based on the \textit{lexical hypothesis}, which posits that individual differences that are important in human interactions (e.g., have survival value across cultures) tend to become encoded in language.
The five factors are Openness, Conscientiousness, Extraversion, Agreeableness, and Neuroticism (OCEAN).
This theoretical foundation has evolved over decades, from \citet{galton1884measurement}'s early work on trait descriptors, through \citet{allport1936trait}'s lexical studies, and into more systematic factor-analytic approaches by scholars such as \citet{norman1963toward} and \citet{goldberg1990alternative}. 
These developments have shaped our modern understanding of personality structure (see Appendix~\ref{sec:appendix personality_structure} for further details).
Given the linguistic basis of personality, LLMs trained on vast and diverse text corpora are well-equipped to model personality data from natural language use.

\vspace{-1mm}
\paragraph{Generative Personality Simulation}

Current approaches to simulating personality data with LLMs exhibit several notable limitations. 
The Persona-Chat dataset, introduced by \citet{zhang2018personalizing}, was originally designed to enhance chit-chat models through increased personalization. 
As such, its primary goal is to foster conversational engagement rather than to accurately model underlying personality constructs.
Another approach, the adjective-based categorization method (Shape method) proposed by \citet{serapio2023personality}, enables the simulation of profile-specific patterns by targeting high or low expressions of individual personality dimensions. 
However, this method struggles to realistically capture personality distributions, as it constrains each case to a single trait dimension—oversimplifying the inherently multidimensional and hierarchical nature of personality structure (e.g.,~\citealp{kachur2020assessing}).
To address these limitations, we explore the use of PSI as a means to generate more nuanced, ecologically valid, and representative simulations of personality data.
\section{Personality Structured Interview for LLM Simulation}
\label{personality_interview_question}

The PSI differs from earlier interview datasets, such as the Life Interview transcripts~\citep{park2024generative}, which ask broad, general questions about a person's life. 
In contrast, the PSI is purposefully designed to elicit linguistic cues associated with personality constructs.
Based on psychometric scale development methods, we developed a structured framework to guide the formulation of interview questions (see Appendix~\ref{sec:appendix_psychometric}).

Simply rephrasing scale items into open-ended questions often fails to elicit meaningful elaboration. 
Converting an item \texttt{``Values artistic, aesthetic experiences''} into \texttt{``Do you value artistic, aesthetic experiences?''} typically yields a brief \texttt{``yes''} or \texttt{``no''} answer, offering little insight~\citep{trull1998structured}. 
Such questions also diverge from the behavioral prompts commonly used in structured interviews, which are more effective for eliciting rich, narrative data (e.g., \citealp{campion1997review}). 
To explore aesthetic values, asking \texttt{``Describe a moment when you felt inspired by an artistic or aesthetic experience''} invites deeper, more contextual responses.

\subsection{Personality Structured Interview Development Process}
\label{question_design}

The overall process of developing a structured interview designed to assess the target construct is detailed in Appendix~\ref{sec:appendix_psychometric}. 
Here, let $\mathcal{C}$ denote the personality construct of interest. 
We present a complete example, formulated as a functional chain, to illustrate how the final set of structured interview questions, $\mathcal{Q}_{\text{final}}$, is produced: 
$\displaystyle \mathcal{C} \xrightarrow{f_1} \mathcal{I} \xrightarrow{f_2} \mathcal{B} \xrightarrow{f_3} \mathcal{Q}_1 \xrightarrow{f_4} \mathcal{Q}_2^{(0)} 
\xrightarrow{f_5~(\text{iterative})} (\mathcal{R}, \mathcal{V}) \xrightarrow{f_6} \mathcal{Q}_{\text{final}}$.


In the original steps, we first need to identify the behavior/perception indicators $\mathcal{I}$ demonstrated by $\mathcal{C}$ ($\displaystyle \mathcal{C} \xrightarrow{f_1} \mathcal{I}$), and based on $\mathcal{I}$ to build the initial blueprint $\mathcal{B}$ for the structured interview questions ($\displaystyle \mathcal{I} \xrightarrow{f_2} \mathcal{B}$). 
Then, based on $\mathcal{B}$, we can generate the initial question pool $\mathcal{Q}_1$ ($\displaystyle \mathcal{B} \xrightarrow{f_3} \mathcal{Q}_1$).
However, given the substantial body of research on personality assessment and constructs—such as narrative identity theory~\citep{mcadams1995we, mcadams1996personality, mcadams2001psychology} and the Structured Interview for the Five-Factor Model (SIFFM;~\citealp{trull1998structured})\footnote{McAdams’ three levels of personality (traits, personal concerns, and narrative identity) as well as other established personality structured interviews, highlight the importance of thoughtfully guiding individuals toward deeper self-expression when designing such interviews. 
Instead of focusing on routine events or superficial details, we should prompt reflection on pivotal moments, meaningful relationships, and future aspirations. 
For example, questions like \texttt{``Can you describe an event that changed the trajectory of your life?''} or \texttt{``Tell me about a moment you are most proud of.''} can elicit rich, narrative-driven insights into personality.}—four subject matter experts ($\displaystyle\text{SMEs} = \{\text{Ph.D.}_{student1}, \text{Ph.D.}_{student2}, \text{Postdoc}, \text{Professor}\}$), all specializing in Personality Psychology, adapted and modified these existing theory and interview questions accordingly to generate the initial question pool: $\mathcal{Q}_1$ ($\displaystyle \mathcal{Q}_1 = \text{adapt}(\mathcal{Q}_\text{prior}, \text{SMEs})$).

Following this, we conducted pilot testing with six undergraduate research assistants affiliated with a personality research laboratory ($\displaystyle\mathcal{D}_{\text{pilot}}$). 
Based on their feedback, the interview questions were further revised by SMEs ($\displaystyle\mathcal{Q}_2 = f_4(\mathcal{Q}_1, \text{Stat}(\mathcal{D}_{\text{pilot}}))$). 
Then we conducted field testing with a convenience sample of university participants ($\displaystyle \mathcal{D}_{\text{field}}$). 
During field testing, the other three doctoral students in Industrial/Organizational and Personality Psychology served as independent observers: they reviewed the interview transcripts and provided observer ratings of participants' personalities to validate the reliability ($\displaystyle \mathcal{R}$) and validity ($\displaystyle \mathcal{V}$) of the structured personality interview questions ($\displaystyle \mathcal{R}^{(t)}, \mathcal{V}^{(t)} = f_5(\mathcal{Q}_2^{(t)}, \mathcal{D}_{\text{field}}) $).
This is an iterative process—if the evaluated metrics do not meet the predefined thresholds $\tau_{\text{rel}}$ and $\tau_{\text{val}}$, the questions are further revised ($\displaystyle \mathcal{Q}_2^{(t+1)} = f_{\text{revise}}(\mathcal{Q}_2^{(t)}, \mathcal{R}^{(t)}, \mathcal{V}^{(t)})$); otherwise, the process produces $\mathcal{Q}_{\text{final}}$ ($\displaystyle \mathcal{R}, \mathcal{V} \xrightarrow{f_6} \mathcal{Q}_{\text{final}}$).

Our field testing results have shown that the average correlation between observer ratings and self-report ratings ($\displaystyle \mathcal{V}$) was 0.36, and the average inter-rater reliability (intraclass correlation; $\displaystyle \mathcal{R}$) was 0.76. These values are consistent with theoretical expectations—meta-analytic research suggests an average self–other rating correlation of 0.36 ($\tau_{\text{val}}$) for the Big Five traits (e.g., \citealp{connolly2007convergent}), and the inter-rater reliability obtained exceeds the commonly accepted threshold of 0.70 ($\tau_{\text{rel}}$).
Detailed results from the validation phase are presented in Appendix~\ref{sec:appendix PSID}.
As a result, $\mathcal{Q}_{\text{final}}$, comprising 32 carefully developed questions, was developed, as shown in Table~\ref{tab:structured_interview_questions} in Appendix~\ref{sec:appendix PSID}. 

\vspace{-1mm}
\subsection{Personality Structured Interview Dataset}
\label{dataset_psi}

Data were collected online using the finalized set of interview questions, $\mathcal{Q}_{\text{final}}$ (Institutional Review Board approval was obtained; protocol number and institution are masked for blind review). 
Participants were sampled from both undergraduate and working adult populations. 
Undergraduate participants were recruited from a large public university in the U.S. Midwest and received course credit for their participation. Working adults were recruited via two widely recognized, high-quality crowdsourcing platforms—Prolific\footnote{\url{https://www.prolific.com/}} and CloudResearch Connect\footnote{\url{https://connect.cloudresearch.com/}}—both known for providing access to diverse and demographically representative samples.
All working adult participants were compensated for their time. 
On average, each structured interview session lasted approximately 34 minutes.

In addition to responding to $\mathcal{Q}{\text{final}}$, participants provided demographic information and completed a standardized personality inventory (BFI-2; \citealp{soto2017next}), along with scales measuring personality-relevant behaviors (organizational citizenship behavior \& counterproductive work behavior; \citealp{spector2010counterproductive}) (see Appendix~\ref{sec:appendix dataset} for detailed descriptions and examples). 
Their responses to $\mathcal{Q}{\text{final}}$ were subsequently incorporated into prompt templates used for generating simulated data (see Appendix~\ref{sec:appendix_prompt} for the specific simulation prompts).
After excluding incomplete responses and those that failed attention checks, the final analytic sample included 357 participants. 
The average age of the participants was 33.30 years, with a standard deviation of 13.06 years.
In terms of gender identity, 52.40\% identified as male and 44.80\% as female.
The racial composition of the sample aligned with national demographic trends, with 70.59\% identifying as White, 10.92\% as Black or African American, 9.24\% as Hispanic/Latinx, and 7.84\% as Asian, and 1.40\% as Other.

\section{Experiments and Results}
\label{experiment}

\subsection{General Settings}
\label{general_settings}

\paragraph{Personality Scale}
We used the Big Five Inventory–2 (BFI-2;~\citealp{soto2017next}), a widely used and well-validated psychological measure specifically designed to assess three distinct facets within each of the five major personality domains of the FFM. 
Each facet is measured by four items, resulting in a total of 60 items. 
Both human respondents and LLMs were explicitly instructed to rate the extent to which they agreed with each item using a 5-point Likert scale (1 = ``Strongly disagree'', 2 = ``Somewhat disagree'', 3 = ``Neither agree nor disagree'', 4 =``Somewhat agree'', 5 = ``Strongly agree''). 
Detailed item content and scoring criteria are presented in Table~\ref{fig:bfi-2}-\ref{fig:bfi-2_facet} in Appendix~\ref{sec:appendix BFI-2}.

\vspace{-1mm}
\paragraph{LLMs}
We evaluated seven widely used LLMs: Mistral-7B~\citep{jiang2023mistral}, Gemma-2-9B and Gemma-2-27B~\citep{gemma_2024}, Llama3-8B and Llama3-70B~\citep{llama3modelcard}, GPT-4o-mini~\citep{gpt4omini}, and GPT-4o~\citep{gpt4o}.
To ensure reproducibility, all models were consistently tested at a temperature of zero.
Detailed prompt descriptions are in Appendix~\ref{sec:appendix_prompt}.

\vspace{-1mm}
\paragraph{Metrics} 
The evaluation metrics vary across experiments and are introduced in detail within their respective sections (§\ref{response_similarity}, §\ref{method_comparison}, and §\ref{related_behavior}).
\subsection{Response Similarity}
\label{response_similarity}
The main purpose of this experiment is to assess the similarity between LLM-simulated responses based on the PSI method and human self-reports. 

\vspace{-1.5mm}
\paragraph{Metrics}
To evaluate the degree of similarity between the outputs of LLMs and human respondents, we report the Mean Absolute Error (MAE) and the Pearson correlation coefficient (\( r \)). 
These two metrics are widely used in personality modeling and psychological assessment due to their ability to capture both absolute differences and linear relationships. 
Specifically, lower MAE values and higher \( r \) values indicate a stronger alignment between the two data sources.
The formula used for calculating \( r \) is provided in Appendix~\ref{sec:appendix additional_setting}.

\begin{figure}[h] 
  \centering
  \includegraphics[width=1\linewidth]{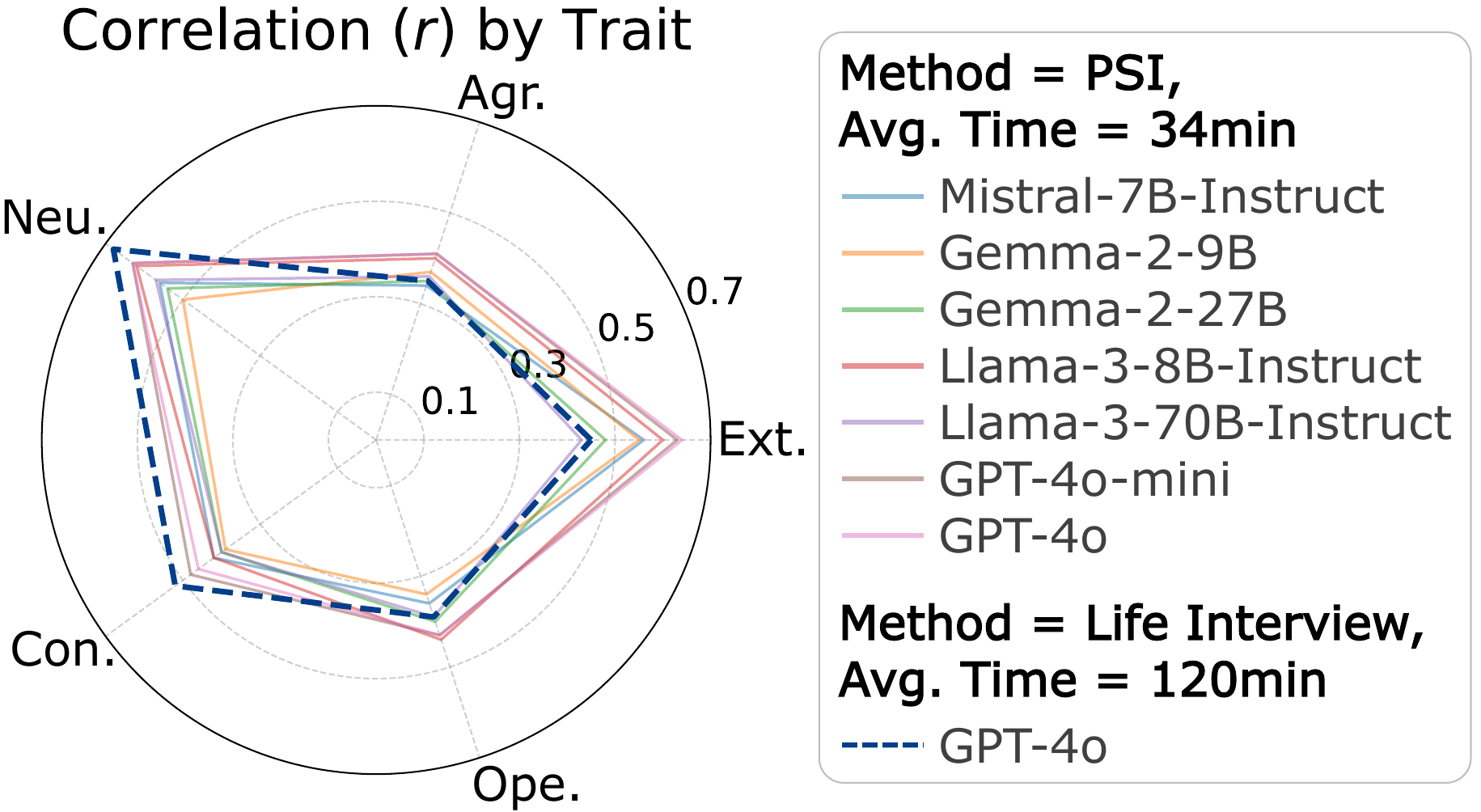} 
  \caption{Correlation of human vs. LLM-simulated data across different LLMs and two methods. PSI achieves comparable correlations to Life Interview using only \textbf{one-quarter} of the time, demonstrating higher efficiency. For specific \( r \) see Table~\ref{tab:exp1} in Appendix~\ref{sec:appendix results.01}.}
  \label{fig:exp1}
\end{figure}

\vspace{-1.5mm}
\paragraph{Results} 
Figure~\ref{fig:exp1} illustrates the \( r \) between PSI method LLM-simulated data and human self-reported data. 
Across models, average \( r \) range from 0.43 to 0.52, indicating a statistically significant and moderately strong positive association between simulated and human responses.

\vspace{-2mm}

\vspace{-3mm}

We further compared our results with prior work by \citet{park2024generative}, who employed a Life Interview simulation—widely regarded as one of the most information-rich methods for simulating personality data.
Their work is notable for offering one of the few direct, one-to-one performance comparisons between LLMs and human participants.
Compared to the Life Interview approach, the PSI method demonstrates better performance across nearly all personality domains. 
When evaluated using the same backbone model, GPT-4o, PSI yields a lower MAE in four domains and a higher \( r \) in three. 
Even where PSI shows slightly higher MAE or lower \( r \), the differences are minor (e.g., MAE for Openness: 0.80 vs. 0.62; \( r \) for Neuroticism: 0.63 vs. 0.68; and Conscientiousness: 0.46 vs. 0.52).

Notably, the Life Interview method described by \citet{park2024generative} requires approximately two hours of interview time. 
In contrast, the PSI approach achieves comparable outcomes with an average duration of just 34 minutes, about one quarter of the time. 
This significant reduction in data collection time not only lessens participant burden but also conserves computational resources needed for simulation. 
These results highlight the efficiency of the PSI method and its effectiveness in leveraging LLMs to simulate personality data.

\subsection{Human Personality Distribution Simulation}
\label{method_comparison}

The primary goal of this experiment is to further examine differences among prompt-based methods for simulating human personality distributions. Specifically, we compared the PSI method, the Persona method (which utilizes dialogue-based information; \citealp{zhang2018personalizing}), and the Shape method (which employs adjective-based dimensional categorization; \citealp{serapio2023personality}), to assess how each simulates a representative human sample.

\vspace{-2mm}

\paragraph{Human Sample Criterion}
The human samples used for this experiment were collected as part of a broader project related to personality assessment through Prolific (Institutional Review Board approval was obtained). 
Participants were instructed to complete a set of demographic questions, the BFI-2, and a set of criterion measures.
In total, 1,559 respondents provided valid responses. 
The average age of the participants was 42.29 years, with a standard deviation of 11.79 years, 50.80\% identifying as men, 49.20\% identifying as women.

\begin{figure*}
    \centering
    \includegraphics[width=1\linewidth]{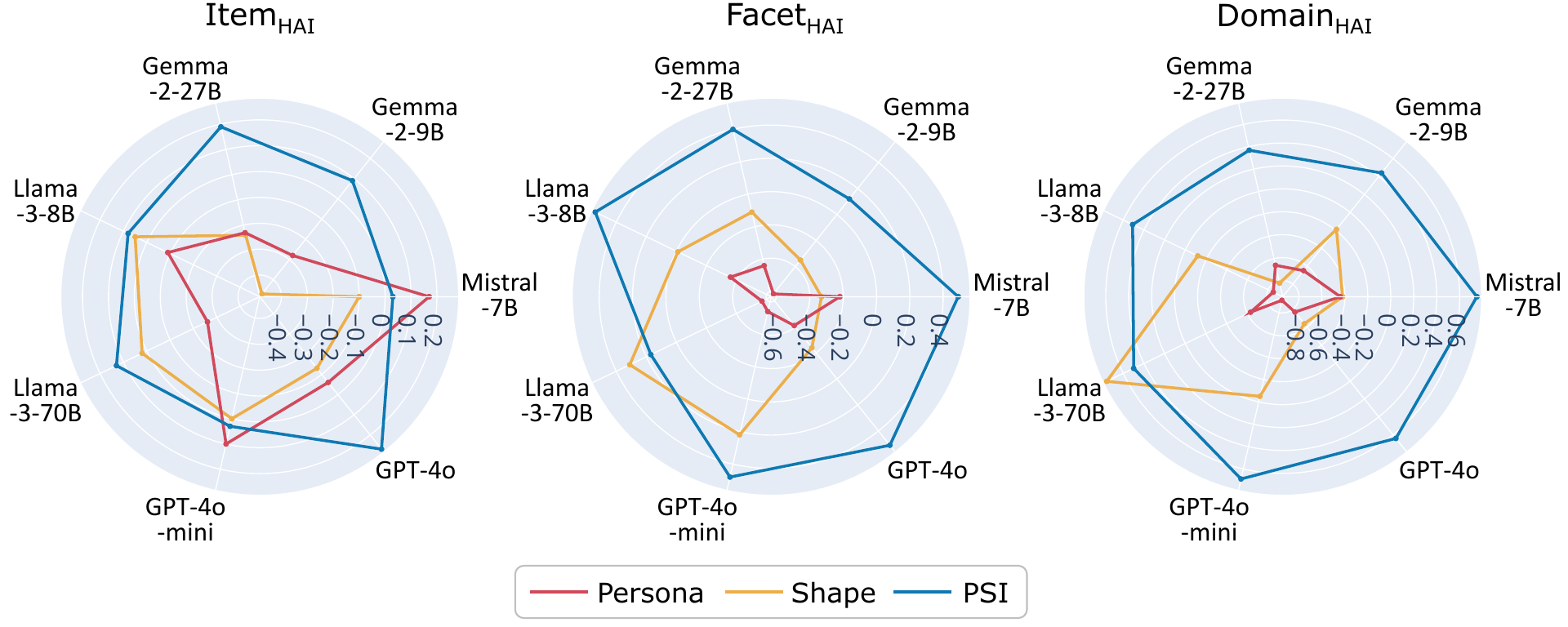}
    \caption{HAI results for human vs. LLM-simulated sample across seven backbone LLMs and three methods. Each subplot shows results at a different granularity level. Higher HAIs (larger area) indicate closer alignment with the heterogeneity distribution in the human sample. PSI method performance far exceeds that of Persona and Shape.}
    \label{fig:hai_results}
\vspace{-5mm}
\end{figure*}

\vspace{-2mm}

\paragraph{Metrics}
To evaluate the similarity in distribution between the human sample and the LLM-simulated sample, we leveraged multiple metrics. 
We computed the sample means ($\mu$) and standard deviations ($\sigma$) at the domain, facet, and item levels. 
For the domain and facet levels, we also calculated Cronbach’s alpha coefficients to assess internal consistency, as well as inter-scale correlations to examine the relationships among constructs.
We further computed the Heterogeneity Alignment Index (HAI\footnote{The HAI can be understood as a profile correlation between different observations or subgroups of $\sigma$. It has been widely applied across various fields to evaluate the similarity between samples or individuals based on multidimensional features (e.g., \citealp{humbad2013quantifying,mccrae2008note}).}) by correlating the $\sigma$ profiles at each level, providing a measure of similarity in the variability structure of personality data between the human and LLM samples.
\vspace{-2mm}
\begin{align}
& \text{HAI} = \notag \\
& \frac{
    \sum_{i=1}^{n} (\sigma_{\text{H},i}\!-\!\bar{\sigma}_{\text{H}})(\sigma_{\text{M},i}\!-\!\bar{\sigma}_\text{M})
}{
    \sqrt{\sum_{i=1}^{n} (\sigma_{\text{H},i}\!-\!\bar{\sigma}_{\text{H}})^2}\!\cdot\!
    \sqrt{\sum_{i=1}^{n} (\sigma_{M,i}\!-\!\bar{\sigma}_{M})^2}
}, \notag
\end{align}

where $\sigma_{\text{H},i}$ and $\sigma_{\text{M},i}$ denote the $\sigma$ of the $i$-th human and LLM sample, respectively. 
Additionally, we conducted a three-factor confirmatory factor analysis (CFA; \citealp{joreskog1969general}) for both the human and LLM-simulated samples, separately within each BFI-2 domain.
In these models, referred to as the Three-Factor Models (TFMs), facets were treated as latent factors.
Beyond modeling the facet structure within each domain, we also employed the facet-level scores as observed indicators in higher-order CFA models to evaluate the broader FFM structure.
We compared model fit indices, factor loadings, and inter-factor correlations across human and LLM-simulated samples.
To quantify the personality construct structural similarity between the two sample types, we used Tucker’s congruence coefficient (TCC; \citealp{tucker1951}) and the MAE of the factor loadings.
Detailed descriptions of the CFA model specifications, fit indices, and the TCC formula are provided in Appendix~\ref{sec:appendix additional_setting}. 
This evaluation framework is adaptable and can be applied to assess the fidelity of other types of simulated psychometric data (see Appendix~\ref{sec:appendix evaluation_framework}).

\vspace{-1mm}

\paragraph{Descriptive Statistics Results}
Figure~\ref{fig:hai_results} presents the HAI results comparing human sample vs. LLM-simulated samples using seven different LLMs and three simulation methods (Persona, Shape, and PSI). 
The three radar plots represent HAI scores at varying levels of granularity (item, facet, and domain). 
Each axis corresponds to a different LLM, and higher HAI values (indicated by a larger area within the plot) reflect a closer alignment with the heterogeneity patterns observed in human responses. 
Across all three levels of granularity, the PSI method consistently outperforms both Persona and Shape, demonstrating substantially greater alignment with human-like heterogeneity.

Table~\ref{tab:mae_results} in Appendix~\ref{sec:appendix results.1} reports the MAE for both $\mu$ and $\sigma$ at the item, facet, and domain levels. 
The $\mu$\textsubscript{MAE} captures the average discrepancy in central tendency between the LLM-simulated and human samples, while the $\sigma$\textsubscript{MAE} reflects differences in variability.
For $\mu$\textsubscript{MAE}, the three methods showed comparable performance. 
However, for $\sigma$\textsubscript{MAE}, both the PSI and Shape methods yielded substantially lower values than the Persona method.

Factor structure analysis of personality relies on variance–covariance matrices at each level, inherently capturing the heterogeneity of multivariate data. 
A high HAI value is thus important for accurately simulating personality data. 
Although the Shape method enhances response variability (evidenced by a low $\sigma$\textsubscript{MAE}), it still falls short in replicating the true variance pattern observed in human samples. 
In contrast, the PSI method partially overcomes this limitation by producing $\sigma$ profiles (HAI) that more closely align with those found in actual human data, indicating a better approximation of real-world psychometric characteristics.

The above results reflect analyses of $\mu$ and $\sigma$ for both human and LLM-simulated samples across different levels of granularity.
Detailed $\mu$ and $\sigma$ values are reported in Tables~\ref{tab:Mistral_d}–\ref{tab:GPT-4o} in Appendix~\ref{sec:appendix results.1}.

\begin{figure}[h] 
  \centering
  \includegraphics[width=1\linewidth]{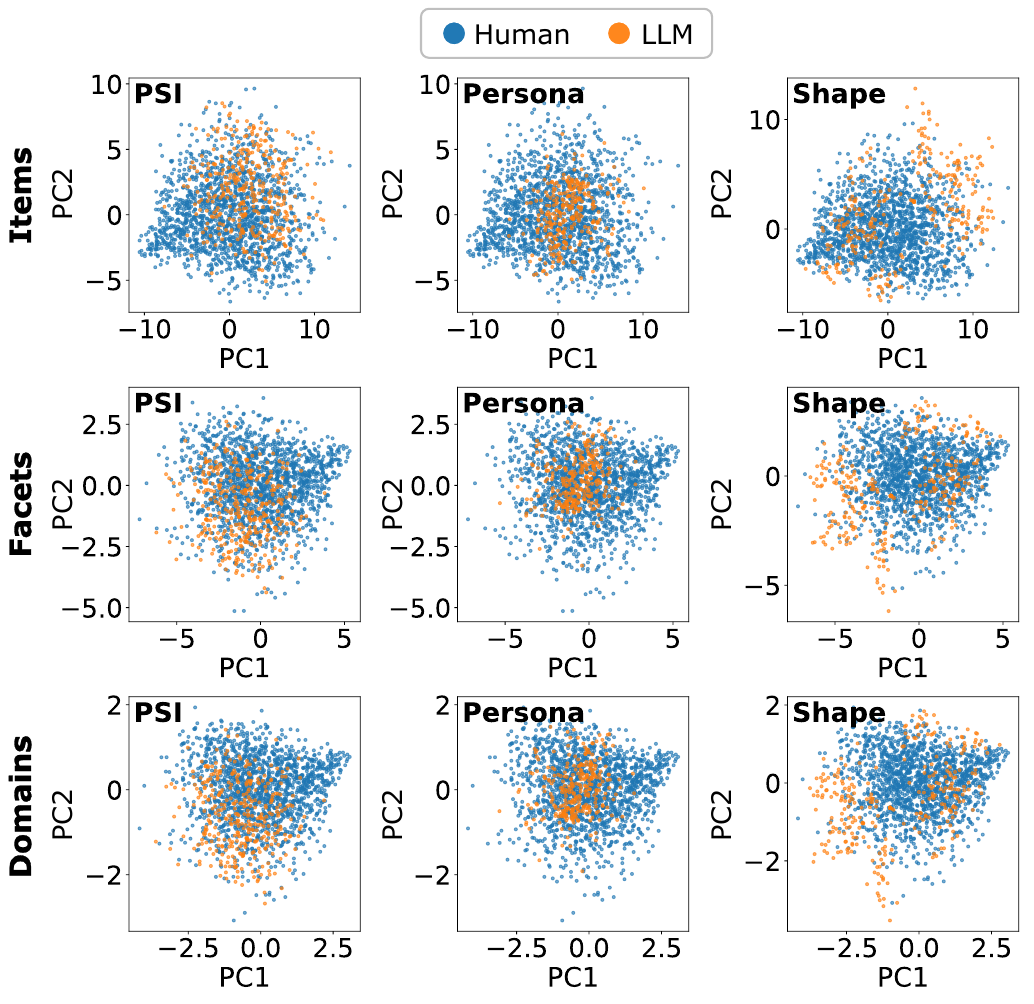} 
  \caption{PCA comparison of human and GPT-4o simulated data across different levels and methods. PSI distributes closest to human; Persona centers with reduced variance; Shape disperses at the extremes.}
  \label{fig:pca}
\end{figure}

\vspace{-1mm}
\paragraph{Personality Data Distribution Visualization}
As shown in Figure~\ref{fig:pca}, we visualized and compared personality data from the human sample and three LLM-based simulation methods across levels by projecting them into a two-dimensional space using principal component analysis (PCA).
We fitted the two-dimensional PCA model using only the human sample, which served as the reference for subsequent projections of the simulated data.

\vspace{-3mm}

\vspace{-3mm}

Among the three methods, PSI yielded the closest alignment with the human distribution, suggesting a stronger capacity to replicate the multivariate structure of real personality data. 
In contrast, the Persona method exhibited a centrally clustered distribution, indicating limited heterogeneity. 
The Shape method showed a more dispersed distribution, reflecting greater variability, as well as a larger divergence from the human pattern.

\begingroup
\sisetup{detect-weight,detect-family,mode=text,reset-text-series=false,reset-text-family=false}

\begin{table*}[t]
\centering
\setlength{\tabcolsep}{3pt} 
\renewcommand{\arraystretch}{0.8}
\begin{tabular}{l l S S S S S S}
\toprule
\textbf{Model} & \textbf{Method} 
& {\textbf{SOC\textsubscript{TCC}}~$\uparrow$} 
& {\textbf{ASS\textsubscript{TCC}}~$\uparrow$} 
& {\textbf{ENE\textsubscript{TCC}}~$\uparrow$} 
& {\textbf{SOC\textsubscript{MAE}}~$\downarrow$}
& {\textbf{ASS\textsubscript{MAE}}~$\downarrow$}
& {\textbf{ENE\textsubscript{MAE}}~$\downarrow$} \\
\midrule
\multirow{3}{*}{Mistral-7B} 
 & Persona & 0.91 & 0.92 & 0.98 & 0.09 & 0.08 & 0.04 \\
 & Shape   & 0.99 & 0.94 & 0.95 & 0.03 & 0.07 & 0.07 \\
 & PSI (ours)     & \textbf{\tablenum{1.00}} & \textbf{\tablenum{0.96}} & \textbf{\tablenum{0.99}} & \textbf{\tablenum{0.02}} & \textbf{\tablenum{0.07}} & \textbf{\tablenum{0.02}} \\
\midrule
\multirow{3}{*}{Gemma-2-9B} 
 & Persona       & \tablenum{1.00} & \tablenum{0.97} & \textbf{\tablenum{1.00}} & \tablenum{0.01} & \tablenum{0.05} & \textbf{\tablenum{0.02}} \\
 & Shape         & 0.97 & 0.86 & 0.97 & 0.04 & 0.10 & 0.07 \\
 & PSI (ours)      & \textbf{\tablenum{1.00}} & \textbf{\tablenum{0.97}} & 0.86 & \textbf{\tablenum{0.01}} & \textbf{\tablenum{0.05}} & 0.09 \\
\midrule
\multirow{3}{*}{Gemma-2-27B} 
 & Persona      & \tablenum{1.00} & \tablenum{0.97} & \tablenum{0.98} & \tablenum{0.02} & \tablenum{0.06} & \tablenum{0.05} \\
 & Shape        & \tablenum{1.00} & \textbf{\tablenum{0.97}} & 0.97 & 0.02 & \textbf{\tablenum{0.06}} & 0.08 \\
 & PSI (ours)       & \textbf{\tablenum{1.00}} & 0.92 & \textbf{\tablenum{0.98}} & \textbf{\tablenum{0.02}} & 0.09 & \textbf{\tablenum{0.04}} \\
\midrule
\multirow{3}{*}{Llama3-8B} 
 & Persona  & -0.78 & 0.55 & 0.71 & 0.41 & 0.16 & 0.14 \\
 & Shape    & -0.81 & 0.79 & -0.80 & 0.40 & 0.11 & 0.33 \\
 & PSI (ours)     & \textbf{\tablenum{1.00}} & \textbf{\tablenum{0.98}} & \textbf{\tablenum{0.98}} & \textbf{\tablenum{0.02}} & \textbf{\tablenum{0.04}} & \textbf{\tablenum{0.04}} \\
\midrule
\multirow{3}{*}{Llama3-70B} 
 & Persona & 0.99 & \textbf{\tablenum{0.98}} & 0.98 & 0.03 & \textbf{\tablenum{0.05}} & 0.05 \\
 & Shape   & 0.99 & 0.97 & 0.96 & 0.05 & 0.05 & 0.06 \\
 & PSI (ours)    & \textbf{\tablenum{1.00}} & 0.95 & \textbf{\tablenum{0.99}} & \textbf{\tablenum{0.03}} & 0.09 & \textbf{\tablenum{0.03}} \\
\midrule
\multirow{3}{*}{GPT-4o-mini} 
 & Persona       & \tablenum{1.00} & \tablenum{0.99} & \tablenum{0.99} & \textbf{\tablenum{0.01}} & 0.04 & 0.05 \\
 & Shape         & 0.91 & 0.95 & \textbf{\tablenum{1.00}} & 0.09 & 0.07 & \textbf{\tablenum{0.03}} \\
 & PSI (ours)     & \textbf{\tablenum{1.00}} & \textbf{\tablenum{0.99}} & 0.98 & 0.02 & \textbf{\tablenum{0.03}} & 0.05 \\
\midrule
\multirow{3}{*}{GPT-4o} 
 & Persona            & \tablenum{1.00} & \tablenum{0.97} & \tablenum{0.98} & \textbf{\tablenum{0.01}} & 0.06 & \textbf{\tablenum{0.05}} \\
 & Shape              & \tablenum{1.00} & \textbf{\tablenum{0.99}} & 0.97 & 0.02 & 0.05 & 0.06 \\
 & PSI (ours)        & \textbf{\tablenum{1.00}} & 0.98 & \textbf{\tablenum{0.98}} & 0.02 & \textbf{\tablenum{0.04}} & 0.06 \\
\bottomrule
\end{tabular}
\caption{TCC and MAE for TFM of Extraversion for human vs. LLM-simulated data across different Backbone Model + Method combinations. Best TCC and MAE are in \textbf{bold}; Higher TCC ($\uparrow$) and lower MAE ($\downarrow$) are better.}
\label{tab:TCC_MAE_EXT}
\vspace{-5mm}
\end{table*}

\endgroup

\vspace{-1mm}

\paragraph{Factor Analysis Results}
Factor analysis encompasses multiple components; here, we present model fit and structural validity results (e.g., fit indices and factor loadings). 
Additional aspects, such as scale reliability and discriminant validity, are reported in Appendix~\ref{sec:appendix results.1}.

\textbf{Model Fit:}
Model fit information for the TFMs and the FFM is presented in Table~\ref{tab:modelfit} in Appendix~\ref{sec:appendix results.1}. 
For both TFMs and FFM, the model fit indices for the Persona and PSI methods are relatively comparable to those observed in the human sample, whereas the Shape method exhibits notably poorer fit.
Detailed explanations of model fit indices are provided in Appendix~\ref{sec:appendix additional_setting}.

\textbf{Structural Validity:}
TCC and MAE are used to assess the overall similarity and discrepancy in factor loadings. 
Table~\ref{tab:TCC_MAE_EXT} illustrates the TFM results for the Extraversion domain as an example, where SOC, ASS, and ENE represent the three facet-level factors. 
Following \citet{lorenzo2006tucker}, a TCC above 0.95 indicates good similarity, while values between 0.85 and 0.94 suggest moderate similarity.
TCC values were generally high, indicating good alignment in factor loadings between the LLM-simulated and human samples across most methods. 
However, notable exceptions were observed for the Persona and Shape methods when using Llama3-8B as the backbone model, where TCC values were negative, suggesting a failure to replicate the human factor structure. 
In terms of MAE, the PSI method generally yielded lower errors, demonstrating better performance in capturing the true factor loading patterns.
Detailed values of TCC and MAE for all TFMs and FFM are provided in Tables~\ref{tcc_tfm_e}–\ref{tab:mae_ffm} in Appendix~\ref{sec:appendix results.1}.
Standardized factor loadings for all TFMs and the FFM are shown in Tables~\ref{tab:fc_sociability}–\ref{tab:fc_creative_imagination} and Tables~\ref{tab:ffm_extraversion}–\ref{tab:ffm_openness}, respectively.
These patterns are consistent with the example presented in Table~\ref{tab:TCC_MAE_EXT}.

Tables~\ref{tab:inter_extraversion}–\ref{tab:inter-factor-ffm} in Appendix~\ref{sec:appendix results.1} report the inter-factor correlation. 
Consistent with our previous findings, the PSI method generally outperforms the other two methods. 
For example, in the TFM results, the Persona and Shape methods exhibited anomalies, such as inter-factor correlations exceeding one and accompanied by warning messages during model fitting. 
In contrast, the PSI method produced more stable and plausible results.

Other results like scale reliability and discriminant validity are in Appendix~\ref{sec:appendix results.1}.
Overall, the PSI method consistently yields better performance compared to the Persona and Shape methods.

\subsection{Personality-Related Behavioral Performance}
\label{related_behavior}

This experiment aims to explore whether LLMs, when assigned specific personality settings, exhibit behaviors that align with theoretical expectations.

\paragraph{Personality-Related Behavior}
We examined well-studied workplace behaviors: organizational citizenship behavior (OCB) and counterproductive work behavior (CWB), both of which are closely linked to personality (e.g.,~\citealp{organ1995meta,berry2007interpersonal}).
Human responses were collected using the scale developed by~\citet{spector2010counterproductive}, and the same items were administered to the LLMs to generate simulated data.
For details on the measurement, see Appendix~\ref{sec:appendix BFI-2}; for prompt construction, refer to Appendix~\ref{sec:appendix_prompt}.

\paragraph{Metrics}
Our primary focus is on the \( r \) between personality domains from different data sources and two behavioral outcomes: OCB and CWB.
We expect that the \( r \) between personality and OCB/CWB in the LLM-simulated data will closely mirror those observed in the human sample.

\paragraph{Results}
Table~\ref{tab:ocb} and \ref{tab:cwb} in Appendix~\ref{sec:appendix results.2} reports the \( r \) between personality domains and OCB/CWB in both human data and data simulated using the PSI method.
The results demonstrate that correlations \( r \) generated by PSI method largely reflect the patterns found in human data across most personality domains, with the exception of Openness. 
Generally, when human data exhibit positive, negative, or negligible \( r \), the LLM-simulated results tend to align in the same direction.
However, these simulated \( r \) are often stronger in magnitude.
This amplification may stem from the models’ reliance on ``typical'' or ``idealized'' knowledge structures internalized during training, which can reinforce and magnify common associations. 
In contrast, human responses are shaped by individual variability and random noise, leading to more diffuse patterns.

\begin{figure*}
    \centering
    \includegraphics[width=0.95\linewidth]{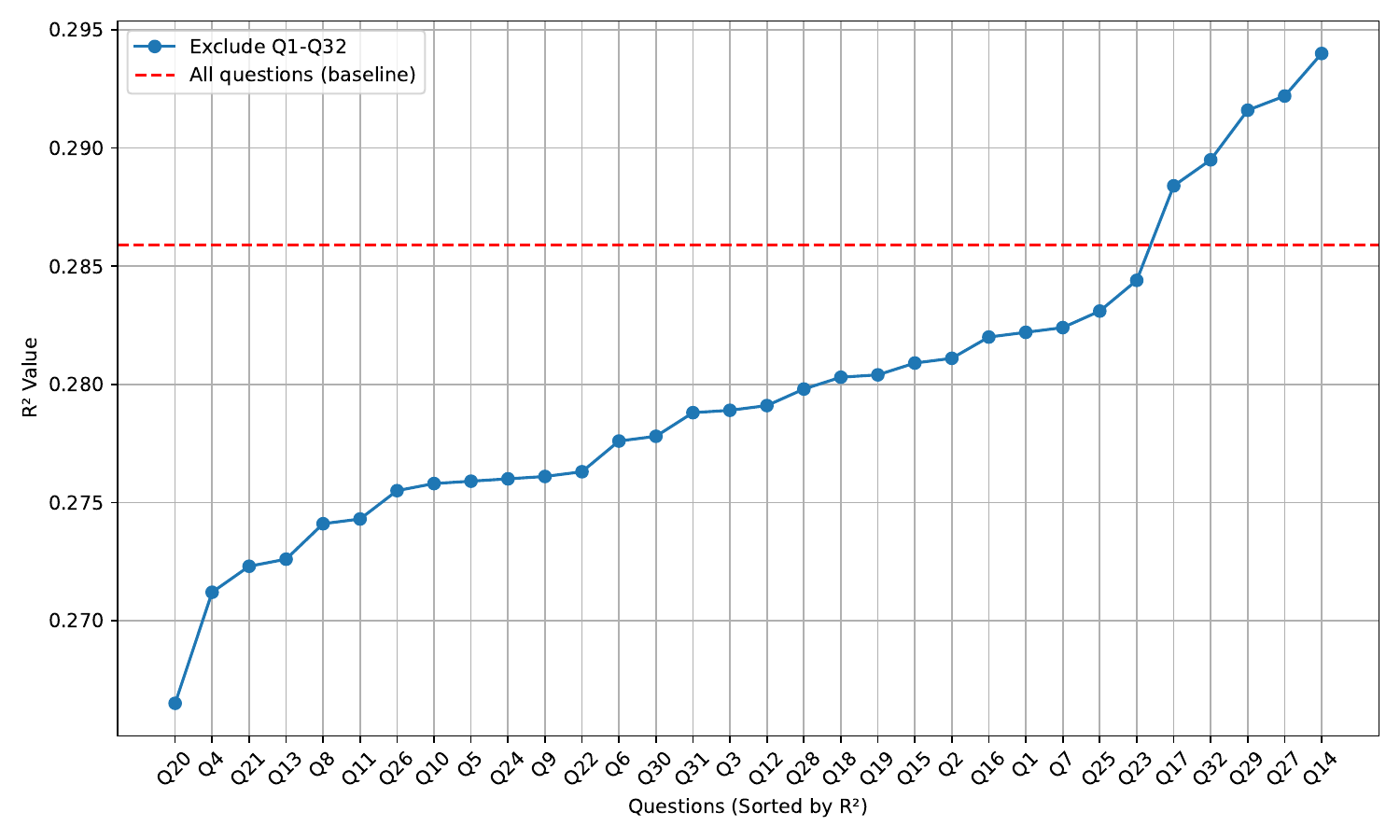}
    \caption{Ablation test results for predicting Conscientiousness. The x-axis represents each question removed on its own, and the y-axis shows the $R^2$ value. The red dashed line indicates the baseline $R^2$ when all questions are included.}
    \label{fig:c_ablation}
\vspace{-5mm}
\end{figure*}

\subsection{Ablation Study on PSI Questions}
We further investigated the individual contributions of each PSI question to personality prediction through an ablation study. 
The PSI was designed as a highly integrated framework, with its questions intentionally co-developed and structurally interlinked to capture diverse facets of personality-related information. 
As a result, isolating individual questions is inherently challenging, as doing so could undermine the integrity of the overall design.

Moreover, each question tends to tap into different aspects of personality, rendering them functionally complementary. 
This also implies that certain questions may be more effective than others in capturing information relevant to specific domains; some items may be more closely aligned with one domain than others. 
Given this design, we expected the differences in individual question effectiveness to be relatively minor overall. 
However, we also anticipated that the relative importance of each question would vary depending on the specific personality domain being predicted.

We conducted an ablation study by leaving one question out each time and simulated personality traits based on the remaining 31 question–answer pairs (see Table~\ref{tab:structured_interview_questions} in Appendix~\ref{sec:appendix PSID} for all questions).
We then calculated the coefficient of determination ($R^2$) using human self-reported personality scores as the reference, in order to assess the impact of each question on model performance. 
If $R^2$ increases after a particular question is removed, this suggests that the question may have negatively affected prediction—potentially introducing redundancy, noise, or irrelevant information that weakened overall performance.
Conversely, if $R^2$ decreases after removal, it indicates that the question made a substantial contribution to prediction and was a key factor supporting model performance.

As shown in the Figure~\ref{fig:c_ablation}, the removal of most questions leads to a slight decrease in the $R^2$, as one would expect, although the changes are minor and generally remain within a ±0.01 range. 
Notably, the relative importance of individual questions varies across different prediction domains (see Figure~\ref{fig:o_ablation}-\ref{fig:n_ablation} in the Appendix~\ref{sec:appendix results.3}).
This empirical finding supports our earlier hypothesis: different types of questions capture distinct facets of personality, offering complementary rather than redundant information. 
Furthermore, the relative importance of questions varies across target domains, indicating that different questions make distinct contributions to domain-specific predictions. 
Together, these findings underscore the meaningful contributions of each PSI question.

\section{Conclusion}

This study introduces a novel method, PSI, for simulating human personality data with LLMs and provides a detailed account of its development and the associated dataset.
Across three experiments, we evaluated the effectiveness of PSI. 
The results show that PSI performs well in simulating personality, yielding moderately strong and statistically significant correlations with human self-report data.
Moreover, when modeling the distribution of personality data at the population level, PSI outperforms existing methods in both human-like heterogeneity and broader factor analysis results. 

Our experiments also assessed PSI’s ability to simulate personality-related behavior. Although LLMs approximate human-like patterns, they still exhibit idealized responses that deviate from the natural variability found in human behavior.
In summary, theory-informed structured interviews such as PSI offer a more realistic and psychometrically grounded approach to simulating human-like data. 
We further discuss the utility of such data in simulation-based research and its potential to improve future psychometric studies in Appendix~\ref{sec:appendix_psychometric}.

\section*{Limitations}

This study is not without limitations.
First, our initial evidence suggests that theory-driven prompts can successfully elicit personality-relevant information embedded in human narratives, enabling LLMs to simulate corresponding psychometric data. 
However, further research is necessary to examine the robustness and generalizability of this approach across diverse contexts and constructs.

Second, the personality assessment in this study is based on self-report measures originally designed for humans.
These tools are intended to capture the stable, internal personality that individuals can access and report, which LLMs, by design, do not inherently possess (see Appendix~\ref{sec:appendix_psychometric} for a detailed discussion on psychometric considerations).
Nevertheless, LLMs can simulate human-like responses that reflect particular personality patterns when guided by theory-informed prompts.
Because the goal of this study is not to assert that LLMs have personality, but rather that they can simulate personality-driven behavior, it is reasonable to apply human-based assessment tools to evaluate the quality and fidelity of that simulation.

\section*{Ethical Statement}

This paper presents a comparison between human data and LLM-simulated data by three different methods. 
All human data were collected in strict accordance with relevant ethical guidelines and were approved by the Institutional Review Board. 
Participants received either reasonable monetary compensation or course credit (in the case of student participants) to ensure fair treatment and appropriate recognition of their contributions.

We placed a strong emphasis on transparency and ethical integrity throughout the research process. 
All participants provided informed consent prior to participation. 
Moreover, to protect privacy and maintain ethical standards, the publicly shared dataset was carefully screened to remove any personally identifiable information.

\section*{Acknowledgments}

This research project is partially supported by the Science of Trustworthy AI Award from Schmidt Sciences, the AI2AI Award from Amazon, the Microsoft Accelerate Foundation Models Research (AFMR) grant program, and by the National Science Foundation (NSF) under Award Number 2522411.
The content is solely the responsibility of the authors and does not necessarily represent the official views of the NSF.

Pengda Wang: conceptualization (equal, lead), methodology (equal, lead), software (lead), formal analysis (lead), investigation (lead), data curation (lead), writing - original draft (lead), writing - review \& editing (equal, lead), visualization (lead); Huiqi Zou: methodology (support), validation (support), formal analysis (support), investigation (support), data curation (support), writing - original draft (support), writing - review \& editing (equal, support), visualization (support); Han Jiang: methodology (support), validation (support), investigation (support), data curation (support), writing - original draft (support), writing - review \& editing (equal, support), visualization (support); 
Hanjie Chen: writing - review \& editing (equal, support); Tianjun Sun: conceptualization (equal, lead), methodology (equal, lead), validation (support), formal analysis (support), investigation (support), resources (equal, lead), data curation (support), writing - original draft (support), writing - review \& editing (equal, lead), supervision (equal, lead), funding acquisition (equal, lead); Xiaoyuan Yi: review \& editing (equal, support), supervision (equal, lead); Ziang Xiao: conceptualization (equal, lead), methodology (equal, lead), validation (support), formal analysis (support), investigation (support), resources (equal, lead), data curation (support), writing - original draft (support), writing - review \& editing (equal, lead), supervision (equal, lead), funding acquisition (equal, lead); Frederick L. Oswald: methodology (support), writing - review \& editing (equal, lead).

\bibliography{custom}
\clearpage


\appendix
\addcontentsline{toc}{section}{Appendices}
\renewcommand{\thesection}{\Alph{section}}

\begin{center}
\LARGE{\textbf{Content of Appendix}}
\end{center}

\begingroup
\hypersetup{linkcolor=black}
\setcounter{tocdepth}{3} 
\startcontents[sections]
\printcontents[sections]{l}{1}{}
\endgroup
\clearpage

\section{Appendix: BFI-2 Scale and Behavioral Measures Scales}
\label{sec:appendix BFI-2}

\subsection{BFI-2 Scale}
\textbf{Instructions:} Here are a number of characteristics that may or may not apply to you. For example, do you agree that you are someone who likes to spend time with others? Please write a number next to each statement to indicate the extent to which you agree or disagree with that statement.

\textbf{Scales:}

\begin{table}[h]
\centering
\scriptsize 
\begin{tabular}{ll} 
\toprule
\# & \textbf{Statement} \\ \midrule
1 & I am someone who is outgoing, sociable. \\
2 & I am someone who is compassionate, has a soft heart. \\
3 & I am someone who tends to be disorganized. \\
4 & I am someone who is relaxed, handles stress well. \\
5 & I am someone who has few artistic interests. \\
6 & I am someone who has an assertive personality. \\
7 & I am someone who is respectful, treats others with respect. \\
8 & I am someone who tends to be lazy. \\
9 & I am someone who stays optimistic after experiencing a setback. \\
10 & I am someone who is curious about many different things. \\
11 & I am someone who rarely feels excited or eager. \\
12 & I am someone who tends to find fault with others. \\
13 & I am someone who is dependable, steady. \\
14 & I am someone who is moody, has up and down mood swings. \\
15 & I am someone who is inventive, finds clever ways to do things. \\
16 & I am someone who tends to be quiet. \\
17 & I am someone who feels little sympathy for others. \\
18 & I am someone who is systematic, likes to keep things in order. \\
19 & I am someone who can be tense. \\
20 & I am someone who is fascinated by art, music, or literature. \\
21 & I am someone who is dominant, acts as a leader. \\
22 & I am someone who starts arguments with others. \\
23 & I am someone who has difficulty getting started on tasks. \\
24 & I am someone who feels secure, comfortable with self. \\
25 & I am someone who avoids intellectual, philosophical discussions. \\
26 & I am someone who is less active than other people. \\
27 & I am someone who has a forgiving nature. \\
28 & I am someone who can be somewhat careless. \\
29 & I am someone who is emotionally stable, not easily upset. \\
30 & I am someone who has little creativity. \\
31 & I am someone who is sometimes shy, introverted. \\
32 & I am someone who is helpful and unselfish with others. \\
33 & I am someone who keeps things neat and tidy. \\
34 & I am someone who worries a lot. \\
35 & I am someone who values art and beauty. \\
36 & I am someone who finds it hard to influence people. \\
37 & I am someone who is sometimes rude to others. \\
38 & I am someone who is efficient, gets things done. \\
39 & I am someone who often feels sad. \\
40 & I am someone who is complex, a deep thinker. \\
41 & I am someone who is full of energy. \\
42 & I am someone who is suspicious of others’ intentions. \\
43 & I am someone who is reliable, can always be counted on. \\
44 & I am someone who keeps their emotions under control. \\
45 & I am someone who has difficulty imagining things. \\
46 & I am someone who is talkative. \\
47 & I am someone who can be cold and uncaring. \\
48 & I am someone who leaves a mess, doesn’t clean up. \\
49 & I am someone who rarely feels anxious or afraid. \\
50 & I am someone who thinks poetry and plays are boring. \\
51 & I am someone who prefers to have others take charge. \\
52 & I am someone who is polite, courteous to others. \\
53 & I am someone who is persistent, works until the task is finished. \\
54 & I am someone who tends to feel depressed, blue. \\
55 & I am someone who has little interest in abstract ideas. \\
56 & I am someone who shows a lot of enthusiasm. \\
57 & I am someone who assumes the best about people. \\
58 & I am someone who sometimes behaves irresponsibly. \\
59 & I am someone who is temperamental, gets emotional easily. \\
60 & I am someone who is original, comes up with new ideas. \\
\bottomrule
\end{tabular}
\caption{BFI-2 scale.}
\label{fig:bfi-2}
\end{table}

\textbf{Scoring:} Reverse-keyed items appear as ``R.''

Reverse coding is a widely used technique in psychological measurement and scale development.
Its primary purpose is to align the scoring direction of all items, ensuring consistency across the scale. 
This process enhances the reliability of the overall measurement and supports a more accurate interpretation of the results.

A balanced inclusion of positively and negatively worded items is a well-established strategy to reduce response biases—such as acquiescence bias, consistency effects, or patterned responding (e.g., selecting the same scale point across items). 
These biases can obscure respondents’ true attitudes or behavioral tendencies, compromising measurement validity.

Although it is not guaranteed, reverse coding can help reduce the influence of social desirability bias. 
When all items are presented in the same direction, participants may easily guess the purpose of the test and provide responses that align with perceived expectations. 
By mixing positive and negative statements and applying reverse coding, the scale can disrupt this pattern, ideally making participants reflect more carefully on their inner states or attitudes when providing responses.

\setlength\tabcolsep{3.5pt}
\begin{table}[h]
\centering
\scriptsize 
\begin{tabular}{ll} 
\toprule
\textbf{Domain Level} & \textbf{Item Numbers} \\ 
\midrule
Extraversion & 1, 6, 11R, 16R, 21, 26R, 31R, 36R, 41, 46, 51R, 56 \\
Agreeableness & 2, 7, 12R, 17R, 22R, 27, 32, 37R, 42R, 47R, 52, 57 \\
Conscientiousness & 3R, 8R, 13, 18, 23R, 28R, 33, 38, 43, 48R, 53, 58R \\
Neuroticism & 4R, 9R, 14, 19, 24R, 29R, 34, 39, 44R, 49R, 54, 59 \\
Openness & 5R, 10, 15, 20, 25R, 30R, 35, 40, 45R, 50R, 55R, 60 \\
\bottomrule
\end{tabular}
\caption{BFI-2 domain level with item numbers.}
\label{fig:bfi-2_domain}
\end{table}

\begin{table}[h]
\centering
\scriptsize 
\begin{tabular}{ll} 
\toprule
\textbf{Facet Level} & \textbf{Item Numbers} \\ 
\midrule
Sociability & 1, 16R, 31R, 46 \\
Assertiveness & 6, 21, 36R, 51R \\
Energy Level & 11R, 26R, 41, 56 \\
Compassion & 2, 17R, 32, 47R \\
Respectfulness & 7, 22R, 37R, 52 \\
Trust & 12R, 27, 42R, 57 \\
Organization & 3R, 18, 33, 48R \\
Productiveness & 8R, 23R, 38, 53 \\
Responsibility & 13, 28R, 43, 58R \\
Anxiety & 4R, 19, 34, 49R \\
Depression & 9R, 24R, 39, 54 \\
Emotional Volatility & 14, 29R, 44R, 59 \\
Intellectual Curiosity & 10, 25R, 40, 55R \\
Aesthetic Sensitivity & 5R, 20, 35, 50R \\
Creative Imagination & 15, 30R, 45R, 60 \\
\bottomrule
\end{tabular}
\caption{BFI-2 facet level with item numbers.}
\label{fig:bfi-2_facet}
\end{table}

\subsection{Behavioral Measures Scales} 
\textbf{Organizational Citizenship Behavior (OCB):} OCB was measured using ten items from~\citet{spector2010counterproductive} to assess extra-role behaviors. 
Items were rated on a frequency scale ranging from 1 (never) to 5 (every day). 
Example item: \texttt{``In the past year, how often have you helped new employees get oriented to the job?''}. 
Internal consistency was Cronbach’s alpha = 0.83.

\textbf{Counterproductive Work Behavior (CWB):} CWB was measured using ten items from~\citet{spector2010counterproductive}, designed to assess harmful workplace behaviors. 
Items were rated on a frequency scale ranging from 1 (never) to 5 (every day). 
Example item: \texttt{``In the past year, how often have you ignored someone at work?''.} 
Internal consistency was Cronbach’s alpha = 0.86.

The detailed scale information for both can be found in Table~\ref{fig:OCB} and Table~\ref{fig:CWB} below.

\begin{table}[h]
\centering
\scriptsize 
\begin{tabular}{ll} 
\toprule
\# & \textbf{Statement} \\
\midrule
1 & Took time to advise, coach, or mentor a co-worker. \\
2 & Helped co-worker learn new skills or shared job knowledge. \\
3 & Helped new employees get oriented to the job. \\
4 & Lent a compassionate ear when someone had a work problem. \\
5 & Offered suggestions to improve how work is done. \\
6 & Helped a co-worker who had too much to do. \\
7 & Volunteered for extra work assignments. \\
8 & Worked weekends or other days off to complete a project or task. \\
9 & Volunteered to attend meetings or work on committees on own time. \\
10 & Gave up meal and other breaks to complete work. \\
\bottomrule
\end{tabular}
\caption{OCB Scale}
\label{fig:OCB}
\end{table}

\begin{table}[h]
\centering
\scriptsize 
\begin{tabular}{ll} 
\toprule
\# & \textbf{Statement} \\
\midrule
1 & Purposely wasted your employer's materials/supplies. \\
2 & Complained about insignificant things at work. \\
3 & Told people outside the job what a lousy place you work for. \\
4 & Came to work late without permission. \\
5 & Stayed home from work and said you were sick when you weren't. \\
6 & Insulted someone about their job performance. \\
7 & Made fun of someone's personal life. \\
8 & Ignored someone at work. \\
9 & Started an argument with someone at work. \\
10 & Insulted or made fun of someone at work. \\
\bottomrule
\end{tabular}

\caption{CWB Scale}
\label{fig:CWB}
\end{table}

\textbf{Instructions:} Recall that in the past year, how often have you done the following things (1=Never, 2=Once or twice, 3=Once or twice per month, 4=Once or twice per week, 5=Every day)?

\section{Appendix: Development and Validation of Personality-Structured Interview Questions}
\label{sec:appendix PSID}

\subsection{Questions and Development Process}
Table~\ref{tab:structured_interview_questions} presents the final set of 32 questions that form the basis of our personality structured interview. 
This framework was developed by adapting and modifying McAdams's life history interview and narrative identity approach~\citep{mcadams1995we,mcadams1996personality,mcadams2001psychology}, while also incorporating components from the Structured Interview of the Five-Factor Model (SIFFM;~\citealp{trull1998structured}).

We adapted and modified the life narrative identity approach proposed by \citet{mcadams1995we, mcadams1996personality, mcadams2001psychology}, incorporating elements from the Structured Interview for the Five-Factor Model (SIFFM; \citealp{trull1998structured}). The initial pool of interview questions was drafted and refined through collaborative discussions among SMEs, including two doctoral students, one postdoctoral researcher, and one professor, all specializing in Personality Psychology.

Following this, we conducted pilot testing with six undergraduate research assistants affiliated with a personality research laboratory. Based on their feedback, the interview questions were further revised by SMEs. 
Once the final set of questions was determined, field testing was conducted with a convenience sample of university participants. 
During field testing, the other three doctoral students in Industrial/Organizational and Personality Psychology served as independent observers: they reviewed the interview transcripts and provided observer ratings of participants' personalities to validate the reliability and validity of the structured personality interview questions. 
If the evaluated metrics do not meet the predefined thresholds, the questions are further revised.
This iterative process of development and feedback led to the construction of the personality structured interview. 

The development process is similar to the development of psychological tests or scales.
We have provided more details on psychometrics and the development framework in Appendix~\ref{sec:appendix_psychometric}.

\begin{table}[h]

\centering
\scriptsize 
\begin{tabularx}{\linewidth}{lX} 
\toprule
\# & \textbf{Questions} \\
\midrule
1 & To get us started, where are you from? Where did you grow up and what was the place like? \\
2 & Thinking back, what kind of student were you in school? \\
3 & Did you have a teacher or teachers that were influential? If so, why? What were they like? \\
4 & What was your favorite subject in school, and why? \\
5 & What was your least favorite subject in school, and why? \\
6 & Still thinking back, who were your heroes when you were young and why? \\
7 & When you were little, what did you want to be when you grew up? And why? \\
8 & What were your dreams and plans when you graduated from high school? What made you have those dreams or plans? \\
9 & If you had complete freedom, what would your dream job be, and why? \\
10 & How have your dreams and goals changed throughout your life? \\
11 & Shifting gears to your childhood, how would you describe the personalities of people in the family you grew up in? For example, what were your parents and/or siblings like? \\
12 & How are you similar or different from your parents and/or siblings? \\
13 & How do you think your similarities and/or differences influenced your relationship with them? \\
14 & What was the best part of your childhood? \\
15 & What do you think were the worst parts of your childhood? \\
16 & Switching gears a little bit, what was your first paid job? How old were you then? (If this is not applicable to you, then please put `NA') \\
17 & What other jobs have you had? (If this is not applicable to you, then please put `NA') \\
18 & What do you do now for a living? And why did you choose it? \\
19 & Please describe your typical work day. \\
20 & What is the best and worst part of your current work? \\
21 & Did you serve in the military? Please tell us about that experience, what was the best and worst part of it? \\
22 & Moving on, what are your adult friendships like? \\
23 & How are your adult friendships different from your childhood friendships? \\
24 & What are your strongest qualities as a friend? In other words, what makes you a great friend to have? \\
25 & What about your weakest qualities in friendships? In other words, what do you struggle with when you are trying to be a friend to someone? \\
26 & Moving onto more general questions, when thinking about your life in general, what are you most proud of? \\
27 & What hobbies or other interests do you have? \\
28 & What things frighten you now? \\
29 & What were some things that frightened you most as a child? \\
30 & What are the three biggest news events that have occurred in your lifetime? \\
31 & If you had the power to solve one and only one problem in the world, what would it be, and why? \\
32 & Tell me about a time when you did not know if you would make it. How did you overcome that challenge? \\
\bottomrule
\end{tabularx}
\caption{Personality Structured Interview (PSI) questions.}
\label{tab:structured_interview_questions}
\end{table}

\subsection{Validation Results}

As previously described, three doctoral students specializing in Industrial/Organizational and Personality Psychology served as observers. 
They read the interview transcripts and rated participants' personalities using an other-report format of the 15-item Extra Short Form of the Big Five Inventory–2 (BFI-2-XS; ~\citealp{soto2017short}). 
To adapt the BFI-2-XS from self-report to observer-report, raters evaluated the extent to which each item (e.g., \texttt{``The participant worries a lot''}) appeared characteristic of the participant based on the transcript. 
The BFI-2-XS was selected due to its brevity and efficiency, which reduced the time burden on raters.
The BFI-2-XS has demonstrated strong convergent validity with the full BFI-2 in both self- and observer-report formats, with trait-level correlations exceeding \textit{r} = 0.85~\citep{soto2017short}. These findings support its use as a brief yet psychometrically sound instrument for personality assessment.

Observer ratings were averaged across the three raters, and interrater reliability was assessed. 
All five traits demonstrated adequate agreement (see Table~\ref{tab:self_obs_corr}), with more behaviorally salient traits such as conscientiousness and neuroticism showing higher intraclass correlations (ICCs), and less observable traits such as openness showing lower ICCs, consistent with prior findings~\citep{allik2010generalizability}. The average interrater reliability (ICC) for observer ratings was 0.76, exceeding the commonly accepted threshold of 0.70.

As shown in Table~\ref{tab:self_obs_corr}, all convergent correlations between self- and observer-rated personality traits were positive and statistically significant, averaging 0.36 across the Big Five traits. These findings align with meta-analytic estimates of self–observer convergence~\citep{connolly2007convergent}. Discriminant correlations among the five traits averaged 0.25 for self-reports and 0.18 for observer-reports, suggesting appropriate trait differentiation.

\begin{table*}[h]
\renewcommand{\arraystretch}{1.3}
\centering
\definecolor{lightyellow}{RGB}{255, 255, 153}
\definecolor{orange}{RGB}{255, 204, 102}
\definecolor{red}{RGB}{255, 102, 102}
\definecolor{graytext}{gray}{0.5}

\begin{tabular}{llcc|ccccc|ccccc}
\toprule
 & &  &  & \multicolumn{5}{c|}{Self-report} & \multicolumn{5}{c}{Observer-rated} \\
\cmidrule(lr){5-9} \cmidrule(lr){10-14}
 & & $\mu$& $\sigma$& Ext & Agr & Con & Neu & Ope & Ext & Agr & Con & Neu & Ope \\
\midrule
\multirow{5}{*}{\rotatebox{90}{Self-report}} 
& Ext      & 3.22 & 0.69 & \textit{0.84} & & & & & & & & & \\
& Agr   & 3.74 & 0.54 & \cellcolor{lightyellow}0.15 & \textit{0.77} & & & & & & & & \\
& Con & 3.51 & 0.67 & \cellcolor{lightyellow}0.30 & \cellcolor{lightyellow}0.37 & \textit{0.85} & & & & & & & \\
& Neu     & 3.07 & 0.78 & \cellcolor{lightyellow}-0.39 & \cellcolor{lightyellow}-0.24 & \cellcolor{lightyellow}-0.41 & \textit{0.89} & & & & & & \\
& Ope        & 3.76 & 0.61 & \cellcolor{lightyellow}0.22 & \cellcolor{lightyellow}0.20 & \cellcolor{lightyellow}0.25 & \cellcolor{lightyellow}\textcolor{graytext}{0.00} & \textit{0.84} & & & & & \\
\midrule
\multirow{5}{*}{\rotatebox{90}{Observer-rated}} 
& Ext      & 3.10 & 0.78 & \cellcolor{red}0.46 & \textcolor{graytext}{0.00} & \textcolor{graytext}{0.05} & -0.19 & \textcolor{graytext}{0.07} & \textit{0.75} & & & & \\
& Agr    & 3.57 & 0.69 & \textcolor{graytext}{-0.07} & \cellcolor{red}0.29 & 0.13 & \textcolor{graytext}{0.00} & \textcolor{graytext}{0.03} & \cellcolor{orange}\textcolor{graytext}{-0.01} & \textit{0.76} & & & \\
& Con & 3.73 & 0.67 & 0.17 & 0.15 & \cellcolor{red}0.39 & -0.20 & 0.12 & \cellcolor{orange}0.10 & \cellcolor{orange}0.32 & \textit{0.79} & & \\
& Neu       & 2.76 & 0.76 & -0.17 & \textcolor{graytext}{0.00} & -0.10 & \cellcolor{red}0.39 & \textcolor{graytext}{0.07} & \cellcolor{orange}-0.27 & \cellcolor{orange}-0.16 & \cellcolor{orange}-0.29 & \textit{0.78} & \\
& Ope       & 3.43 & 0.69 & \textcolor{graytext}{-0.07} & \textcolor{graytext}{0.00} & \textcolor{graytext}{0.00} & 0.11 & \cellcolor{red}0.29 & \cellcolor{orange}\textcolor{graytext}{0.01} & \cellcolor{orange}0.22 & \cellcolor{orange}\textcolor{graytext}{0.07} & \cellcolor{orange}\textcolor{graytext}{-0.02} & \textit{0.72} \\
\bottomrule
\end{tabular}
\caption{Correlation matrix of self- and observer-scored big five personality. Ext = extroversion. Agr = agreeableness. Con = conscientiousness. Neu = neuroticism. Ope = openness.
Gray values indicate non-significant correlations.
The diagonal reports reliabilities in italics, using Cronbach’s alpha for self-ratings and intraclass correlations for observer ratings.
Red highlights indicate convergent correlations between self- and observer-reported scores.
Light yellow and dark yellow highlights represent discriminant correlations for self-report and observer-rated personality scores, respectively.}
\label{tab:self_obs_corr}
\end{table*}
\section{Appendix: Personality Structured Interview Dataset and Data Collection}
\label{sec:appendix dataset}

The data were collected through an online questionnaire, followed by an online structured interview.
We will share the dataset that we have obtained permission to share, which also excludes any personally identifiable information.
This portion of the data is available upon request and may not be used for any commercial purposes. Academic use is permitted only with prior approval.
Each example is composed of the following characteristics: 

\begin{enumerate}
\item \textbf{Gender:} The gender of the participant.
\item \textbf{Race:} The racial background of the participant.
\item \textbf{English:} Whether English is the participant’s first language.
\item \textbf{Age:} The participant’s age.
\item \textbf{Weight:} The participant’s weight.
\item \textbf{Height}: The participant’s height.
\item \textbf{OCB1–OCB10:} Self-reported Organizational Citizenship Behavior data, measured using the scale from ~\citet{spector2010counterproductive}, and see Table~\ref{fig:OCB} for specific item details.
\item \textbf{CWB1–CWB10:} Self-reported Counterproductive Work Behavior data, measured using the scale from ~\citet{spector2010counterproductive}, and see Table~\ref{fig:CWB} for specific item details.
\item \textbf{Q1–Q32:} Participant responses to each personality structured interview questions, see Table~\ref{tab:structured_interview_questions} for specific question details.
\item \textbf{Item1–Item60:} Responses to each item of the BFI-2 (already reverse coded). Refer to Appendix~\ref{sec:appendix BFI-2} for item descriptions and scoring guidelines.
\end{enumerate}
\section{Appendix: Prompts List}
\label{sec:appendix_prompt}

\subsection{Personality Scale Responses Prompt Format} 
To minimize the interference of the prompt template on model behavior, we use a standardized prompt structure, defined as: $\displaystyle \text{Prompt}(d) = T_{\text{base}} + d$, where $T_{\text{base}}$ is a fixed base template: $\displaystyle T_{\text{base}}$ = \texttt{``For the following task, respond in a way that matches:''}
The variable $d$ represents the personality description content, determined by the method used. Specifically: $\displaystyle d \in \{ d_{\text{Persona}}, d_{\text{Shape}}, d_{\text{PSI}} \}$. 
The construction of $d$ varies across methods. 
For the Persona method, $\displaystyle  d_{\text{Persona}} = \textstyle\sum_{i=1}^{5} s_i$, where each $s_i$ is a short sentence from the Persona-Chat dataset, such as: \texttt{``I wear a lot of leather.''} 
Each $d_{\text{Persona}}$ represents a unique individual profile.
For the Shape method, $\displaystyle d_{\text{Shape}} = \textstyle\sum_{i=1}^{5} q(a_i)$, where $a_i$ is an adjective representing a personality trait (e.g., \textit{friendly, energetic}), and $q(\cdot)$ is a linguistic intensity modifier (e.g., \textit{extremely, very}). 
For the PSI method, $\displaystyle d_{\text{PSI}} = \textstyle\sum_{i=1}^{32} (\texttt{[Q}_i\texttt{]} + \texttt{[A}_i\texttt{]})$, where \(\texttt{Q}_i\) denotes the \(i\)-th interview question and \(\texttt{A}_i\) is the corresponding response.

Table~\ref{tab:personality_prompt} presents the prompts used to generate LLM-simulated responses for the selected personality test (BFI-2), formatted on a Likert scale. 
In each prompt, \texttt{personality\_description} denotes the personality-specific framing, while \texttt{test\_item} refers to an individual item from the BFI-2 scale.
Example prompts for each method are also included in Table~\ref{tab:personality_prompt}.

\textbf{PSI Method:}
In the PSI method, the personality prompt integrates both the interview question and the interviewee’s corresponding response, as collected in the PSI dataset (see Appendix~\ref{sec:appendix dataset}).

\textbf{Persona Method:}
The Persona method is based on the Persona-Chat dataset constructed by~\citet{zhang2018personalizing}. 
The dataset consists of persona descriptions, and each is made up of five short sentences containing demographic information collected through Amazon Mechanical Turk crowdsourcing. 
To avoid sentence similarity or repetition, these persona descriptions were required to be paraphrased (e.g., changing \texttt{``I am very shy''} to \texttt{``I am not a social person''}). 
\citet{zhang2018personalizing} demonstrated through machine learning model validation and human evaluations that such persona descriptions provide an effective method to enhance personalization. 
Currently, incorporating personal profiles into prompts is widely used in research related to LLM agents~\citep{park2023choicemates,wang2023does,xi2023rise}.
In our study, we treated each persona description as an individual entity (i.e., a single subject) and randomly selected 300 persona descriptions from the dataset. 
One example is, \texttt{``I wear a lot of leather. I have boots I always wear. I sleep in late during the day. I listen to metal music. I have black spiky hair.''}

\textbf{Shape Method:}
The Shape method is based on the work of~\citet{serapio2023personality}, who introduced a prompting approach to shape synthetic personality in LLMs along desired dimensions. 
The researchers expanded upon~\citet{goldberg1990alternative}, expanding his list of 70 bipolar adjectives~\citep{goldberg1992development} to include 104 trait adjectives. 
Additionally, they employed linguistic qualifiers commonly used in Likert-type scales~\citep{likert1932technique}, such as \texttt{``a bit''}, \texttt{``very''}, and \texttt{``extremely''} to set target levels for each adjective. 
This resulted in a fine-grained prompting method with nine levels: 1. extremely {low adjective}; 2. very {low adjective}; 3. {low adjective}; 4. a bit {low adjective}; 5. neither {low adjective} nor {high adjective}; 6. a bit {high adjective}; 7. {high adjective}; 8. very {high adjective}; 9. extremely {high adjective}. 

In our study, each prompt involves five randomly selected adjective markers from a specific personality domain. 
These markers are positioned after a consistent linguistic qualifier to set the prompt at one of nine intensity levels. 
For example, one prompt is: \texttt{``You are extremely friendly, extremely energetic, extremely assertive, extremely bold, and extremely active.''}
These are five positively keyed adjectives targeting the factor (trait) of Extraversion. 
In this case, the prompt seeks to create a highly sociable and dynamic personality profile, which might result in responses characterized by narrower traits (facets) of enthusiasm, confidence, and proactivity. 
We also randomly select 300 prompts.


\subsection{Behavioral Question Responses Prompt Format}
The complete prompt format for eliciting LLMs' responses to personality-related behavioral questions is defined in Table \ref{tab:behavior_prompt}. 
The \textit{questions\_and\_responses} refers to the transcript of a structured human personality interview, while \textit{question\_list} comprises statements evaluating OCB and CWB, as detailed in Table~\ref{fig:OCB} and Table~\ref{fig:CWB} in Appendix~\ref{sec:appendix BFI-2}. 
The model is instructed to rate each statement using a frequency scale from one to five.

\begin{table*}
\caption{Prompt format and examples for gathering LLMs' responses to BFI-2 scale.}     \label{tab:personality_prompt}
    \centering
    \includegraphics[width=\linewidth]{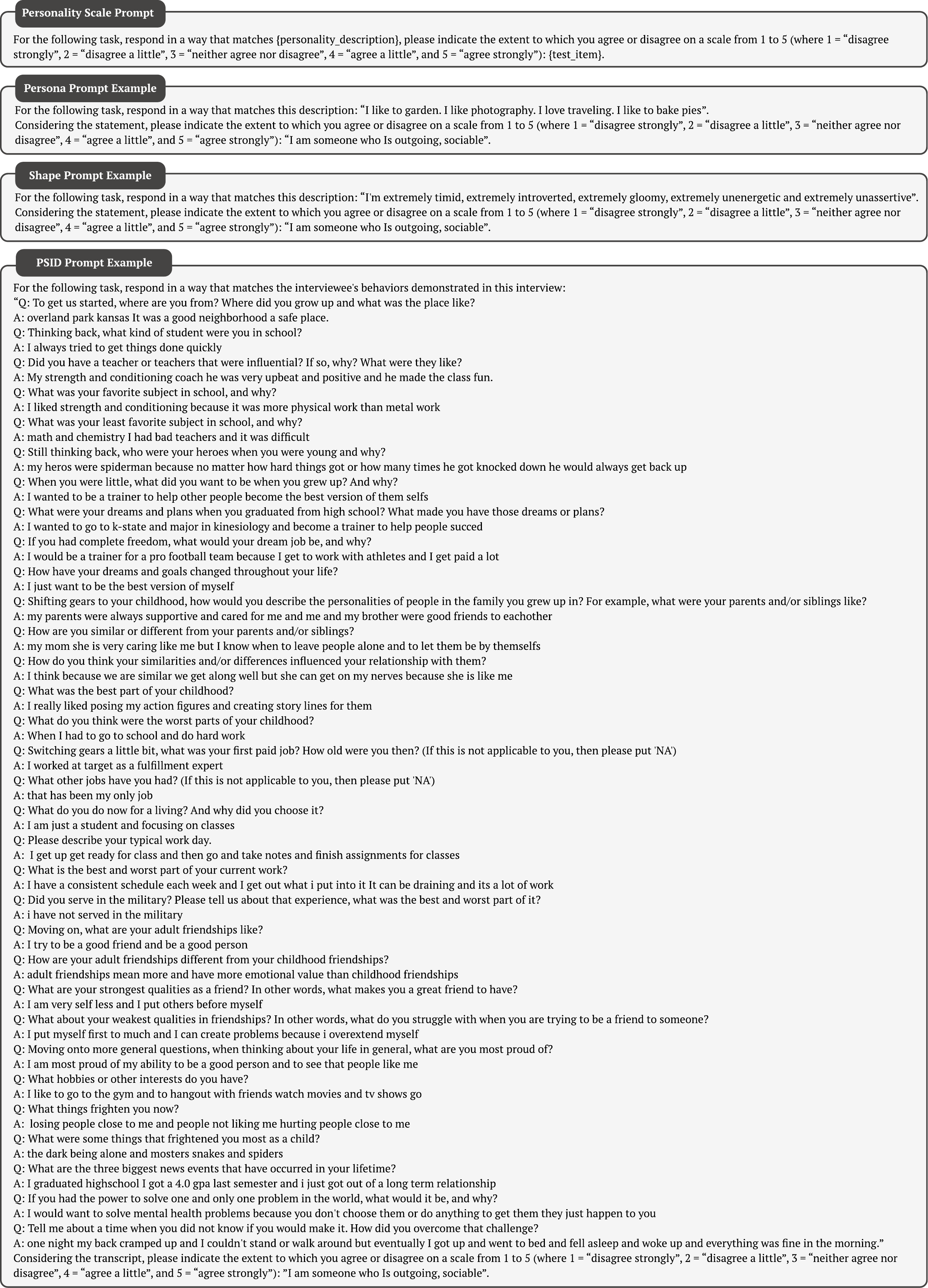}     
\end{table*}

\begin{table*}[h]
\caption{Prompt format for gathering LLMs' responses to personality-related behavioral questions.}
    \label{tab:behavior_prompt}
    \centering
    \includegraphics[width=\linewidth]{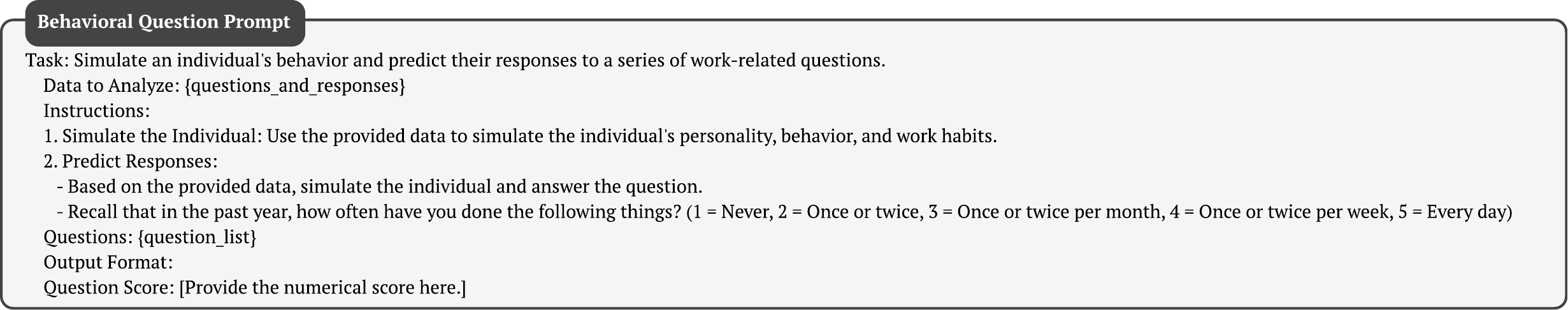}     
\end{table*}

\subsection{Compute Resources}
All LLMs were conducted using the OpenAI and TogetherAI platforms, running a total of seven models: Mistral-7B~\citep{jiang2023mistral}, Gemma-2-9B, Gemma-2-27B~\citep{gemma_2024}, Llama3-8B, Llama3-70B~\citep{llama3modelcard}, GPT-4o-mini~\citep{gpt4omini}, and GPT-4o~\citep{gpt4o}.
Using the GPT-4o tokenizer as an illustrative example, the approximate input token counts for generating BFI-2 LLM responses are as follows: Persona method: 2,555,820 tokens; Shape method: 2,316,240 tokens; and PSI method: 34,510,812 tokens.
The total number of output tokens for the BFI-2 LLM responses is approximately 25,280.
For the PSI method, the input tokens used to generate responses to the OCB and CWB scales were approximately 690,497 and 681,215, respectively, with output tokens totaling around 153,887 and 139,207.
Token counts may vary slightly across models due to differences in tokenizers, formatting, or prompt structures, but they generally fall within a comparable range.
\section{Appendix: Evaluation Metrics and Formulas}
\label{sec:appendix additional_setting}

\subsection{Pearson Correlation Coefficient}

The formula for correlation is as follows: 
\[
r = \frac{\sum (X_i - \overline{X}) (Y_i - \overline{Y})}{\sqrt{\sum (X_i - \overline{X})^2 \sum (Y_i - \overline{Y})^2}}
\]
\( X_i \) and \( Y_i \) represent the data values for LLMs and human. 
The symbols \( \overline{X} \) and \( \overline{Y} \) represent the means of variables \( X \) and \( Y \), respectively. 
The numerator, \( \sum (X_i - \overline{X})(Y_i - \overline{Y}) \), represents the covariance between \( X \) and \( Y \). 
The denominator, \( \sqrt{\sum (X_i - \overline{X})^2 \sum (Y_i - \overline{Y})^2} \), standardizes the result, let the value of \( r \) range between -1 and 1. 

When \( r \) is close to 1, it indicates a strong positive correlation between the two variables; when \( r \) is close to -1, it indicates a strong negative correlation; and when \( r \) is close to 0, it indicates no significant linear relationship between the two variables.

\subsection{CFA Model}
The basic form of the CFA model is:
\[
y = \Lambda \eta + \epsilon
\]
where \( y \) represents the vector of observed variables; \( \Lambda \) is the factor loading matrix (i.e., the loadings of each observed variable on the latent factors); \( \eta \) is the vector of latent factors; \( \epsilon \) is the vector of error terms, with the assumption that the error terms have a mean of zero and are mutually independent~\citep{joreskog1969general}.

Latent factors are variables that are not directly observed but are inferred from other variables that are observed (measured). 
In the context of a CFA model, latent factors represent underlying constructs or traits that are believed to influence the observed variables. 
For example, in psychology, a latent factor might represent a construct like ``intelligence'' or ``anxiety,'' which cannot be measured directly but can be estimated through related observed behaviors or responses on a test.

\subsection{Model Fit Information}

\textbf{Chi-Square Test, \(\mathbf{\chi^2}\):} 
The chi-square test is used to measure the difference between the observed covariance matrix and the factor model's fitted covariance matrix.
\[
\chi^2 = (N - 1) \times F_{\text{ML}}
\]
where \( N \) is the sample size; \( F_{\text{ML}} \) is the value of the fit function under maximum likelihood estimation.
A smaller chi-square value indicates a better model fit.
However, with large samples, the chi-square value tends to be large, so other fit indices are usually the primary reference.

\textbf{Degrees of Freedom, \textit{df}:}
The \textit{df} represents the relationship between model parameters and observed variables:
\[
df = \frac{p(p+1)}{2} - q
\]
where \( p \) is the number of observed variables, and \( q \) is the number of model parameters.

Degrees of freedom reflect the amount of independent information available in a statistical model to estimate its parameters. 
They are calculated as the number of information points provided by the data minus the number of model parameters, representing the extent to which the model can adjust freely. 
Therefore, a higher degree of freedom indicates fewer parameters in the model and fewer constraints on the data. 
With more degrees of freedom, the model has fewer restrictions, though the fitting difficulty may increase. 
Too few degrees of freedom may lead to overfitting, while too many can result in underfitting.

\textbf{Comparative Fit Index, CFI:}
The CFI is used to compare the goodness of fit of a model with a baseline model (usually an independent model).
\[
\text{CFI} = 1 - \frac{\max(\chi^2 - \textit{df}, 0)}{\max(\chi_{\text{null}}^2 - \textit{df}_{\text{null}}, 0)}
\]
where \( \chi^2 \) and \( \text{df} \) are the chi-square value and degrees of freedom of the target model; \( \chi_{\textit{null}}^2 \) and \( \textit{df}_{\text{null}} \) are the chi-square value and degrees of freedom of the baseline (independent) model.

\textbf{Tucker-Lewis Index, TLI:}
TLI, also known as the Non-Normed Fit Index (NNFI), considers model complexity.
\[
\text{TLI} = \frac{\left(\chi^2_{\text{null}} / \textit{df}_{\text{null}}\right) - \left(\chi^2 / \textit{df}\right)}{\left(\chi^2_{\text{null}} / \textit{df}_{\text{null}}\right) - 1}
\]
This value ranges from 0 to 1, with a value typically greater than.90 indicating good model fit.

\textbf{Root Mean Square Error of Approximation, RMSEA:}
The RMSEA quantifies the error per degree of freedom in a model, with smaller values indicating better model fit.
\[
\text{RMSEA} = \frac{\chi^2 - df}{df(N - 1)}
\]
where \(\chi^2\) is the chi-square value of the model; \textit{df} is the degrees of freedom; \textit{N} is the sample size.

\textbf{Standardized Root Mean Square Residual, SRMR:}
The SRMR measures the discrepancy between model-predicted values and actual observed values, calculated as:
\[
\text{SRMR} = \sqrt{\frac{\sum_{i=1}^{p} \sum_{j=1}^{p} \left(s_{ij} - \hat{s}_{ij}\right)^2}{\frac{p(p+1)}{2}}}
\]
where \( s_{ij} \) is an element in the observed covariance matrix; \( \hat{s}_{ij} \) is an element in the model-fitted covariance matrix.

Based on~\citet{hu1999cutoff}, CFI/TLI $\geq$ 0.95, RMSEA $\leq$ 0.06, and SRMR $\leq$ 0.08 are considered good fit thresholds.

\subsection{Tucker's Congruence Coefficient}
Tucker's congruence coefficient, also known as the coefficient of congruence, is typically used to assess the similarity between two-factor structures in factor analysis~\citep{tucker1951}. 
The formula is given by:

\[
\phi = \frac{\sum_{i=1}^n a_i b_i}{\sqrt{\sum_{i=1}^n a_i^2 \cdot \sum_{i=1}^n b_i^2}}
\]
where \( a_i \) and \( b_i \) are the loadings of the \( i \)-th factor for two different factor solutions (or different samples or methods); \( n \) is the total number of factors.

The coefficient ranges from -1 to 1, where values close to 1 indicate high similarity (congruence) between the factor solutions, values close to 0 indicate low similarity, and negative values indicate a dissimilar or inverse relationship.
According to~\citet{lorenzo2006tucker}, a TCC above 0.95 indicates good similarity, while a TCC of 0.85 to 0.94 suggests fair similarity.
However, this is a relatively lenient criterion; specific differences still need to be determined based on the factor loadings.

\section{Appendix: The Structure of Personality}
\label{sec:appendix personality_structure}

A key question in personality-related LLM research pertains to personality structure: What is the nature and breadth of the personality traits we want to simulate the human personality distributions?
In the research literature, personality structures often emerge from applying factor analysis to individuals' responses to a large number of personality-relevant items. 
This approach is what has been used to identify the five personality factors in the FFM. 
Moreover, personality is better understood in terms of one's continuous standing on each of multiple dimensions rather than as static types or profiles. 
Research data clearly supports this view~\citep{wilmot2015contemporary, wilmot2019direct}. 
Dividing individuals into limited categories (e.g., 16 types in MBTI) artificially segments continuous dimensions into discrete units, which may overlook important individual differences~\citep{ones2018Wiernik}. 

Another question worth paying attention to is that, although the FFM occupies a dominant position in personality research, it remains debatable whether focusing solely on the FFM is sufficient in light of the many newly proposed personality variables and alternative structures.
It turns out that there can be conceptual overlap among these models (e.g.,~\citealp{hough2015beyond}). 
For example, a meta-analysis by~\citet{joseph2010emotional} revealed that emotional intelligence (EI) shows statistically and practically significant relationships with neuroticism and extroversion within the FFM. 
In fact, when controlling for personality variables, the unique contribution of EI almost disappears. 
Similarly, \citet{crede2017much} conducted a meta-analysis on grit and found that its core components can largely be explained by conscientiousness, with little added predictive validity beyond that. 
Moreover, although the HEXACO structure introduces an honesty-humility dimension, the remaining five dimensions align closely with the FFM structure~\citep{lee2004psychometric, lee2006further}. 
Research by \citet{cutler2023deep} further indicates that nearly all personality semantic information can be classified within the FFM structure. 

However, this does not mean that the FFM has exhausted the depiction of personality. 
On the contrary, although the FFM performs well in terms of structure and predictive power, it may still miss some more nuanced aspects of personality. 
For instance, measurement tools such as the Facet MAP (\citealp{irwing2024towards}) attempt to capture ``micro-level differences'' in personality by focusing on the lower-level structure of each dimension. 
Therefore, while the FFM is ``good,'' there remains room for further refinement in exploring the complexity and diversity of personality. 
Rather than pursuing a wide range of personality frameworks, this paper will delve deeper into the FFM structure, which is widely accepted among personality psychologists.
\section{Appendix: Psychometrics and Structured Interview Development Framework}
\label{sec:appendix_psychometric}

Psychometrics is a field of psychology dedicated to the theory and practice of psychological measurement. 
It primarily focuses on quantifying psychological traits, behaviors, and abilities through systematic testing and analysis. 
Psychological traits, such as personality dimensions, cognitive abilities, and emotional states, are inherently abstract constructs that cannot be measured directly. 
Therefore, psychometricians rely on tools such as surveys, questionnaires, scales, or interviews to infer these traits through observable indicators or responses.

\subsection{Measuring Psychological Traits}
To measure a psychological trait, psychometricians typically operationalize the trait by identifying observable behaviors or self-identities that correlate with the underlying construct. 
For example, extraversion can be measured by assessing behaviors such as sociability, assertiveness, and enjoyment of social interactions, or by examining identities such as seeing oneself as an outgoing person and believing that one thrives in social situations.

These behaviors are translated into measurable items (for scale; e.g., \texttt{``I enjoy being the center of attention''}) or questions (for interview; e.g., \texttt{``What are your strongest qualities as a friend? In other words, what makes you a great friend to have?''}).
The challenge lies in ensuring that these items/questions accurately and consistently capture the construct across different populations and contexts.

\subsection{Theory-Informed Structured Interview Transcripts for LLM Data Simulation}
Theory-informed structured interviews are the most suitable method for enabling LLMs to simulate psychometric data. 
These interviews are specifically designed to capture the constructs underlying the targeted psychometric measures, ensuring that the simulated data aligns with the intended psychological construct.
By extracting textual information that directly reflects the target construct, theory-informed structured interviews facilitate the representation of heterogeneous data while preserving a high degree of human diversity, thereby enhancing the validity and applicability of the simulated psychometric data.

Moreover, since the information is extracted based on theoretical foundations, it also provides a certain level of interpretability for the LLM’s simulation. 
This not only allows the generated data to be compared with theoretical expectations but also increases its potential for practical applications.

\subsection{The Potential for Advancing Research} 
A theory-informed structured interview transcript-based simulation can generate data more effectively by focusing on the target construct. 
Ideally, the simulated data should reproduce the same constructs reflected in real-world data and simulate behaviors associated with these constructs.

Take personality as an example---if simulated data can accurately replicate real-world personality constructs, it enables research that would be difficult to conduct in reality, such as developing contextualized personality assessment tools and exploring new personality theories through multi-agent simulations.

\textbf{Developing Contextualized Personality Assessment Tools:}
Traditional personality assessments mainly rely on standardized questionnaires or laboratory tasks, which often fail to adequately simulate real-world social contexts. 
By using theory-informed structured interview transcript-based simulations, we can generate more fine-grained and context-sensitive individual response data. 
For instance, we can simulate various occupational scenarios (such as crisis management, teamwork, or remote work) and analyze how different personality traits manifest in these contexts. 
This approach not only aids in developing measurement tools tailored to specific applications but also enhances ecological validity, allowing for more accurate assessments of personality across different situations.

\textbf{Exploring New Personality Theories through Multi-Agent Simulation:}
If simulated data can accurately reflect real-world personality constructs, we can leverage multi-agent interactive systems to simulate individuals' behavioral patterns and observe how different personality traits evolve in group dynamics. 
For example, virtual agents with distinct personality traits can be placed in cooperative tasks, competitive environments, or social interactions, enabling researchers to test whether existing personality theories effectively predict these interaction patterns. 
Additionally, this approach can uncover new personality dynamics, such as whether certain personality trait combinations produce unexpected group effects or whether behavior in specific situations deviates from traditional theoretical predictions.

However, these assumptions are based on an ideal premise—that simulated data can successfully reflect the same constructs as real-world data. 
A theory-informed structured interview undoubtedly offers a promising pathway in this regard, warranting further in-depth exploration.

\subsection{Structured Interview Development Framework}
Developing a structured interview for LLM to simulate data involves a series of carefully designed steps to ensure that the resulting test reliably and validly measures the construct of interest. 
Table~\ref{scale} outlines the framework.

\begin{table*}[h]
\centering
\begin{tabularx}{\linewidth}{lX}
\toprule
\# & \textbf{Steps} \\
\midrule
1 & Identify behaviors/perceptions that represent the construct or define the domain. \\
2 & Prepare a set of structured interview specifications, structured interview blueprint. \\
3 & Build an initial question pool. \\
4 & Have questions reviewed by substantive experts (and revise as necessary). \\
5 & Hold preliminary question tryouts. \\
6 & Determine statistical properties of questions (and eliminate poor questions or revise as necessary). \\
7 & Field-test the structured interview on a large representative sample of the intended examinee population. \\
8 & Design and conduct reliability and validity studies for the final form of the structured interview. \\
9 & Develop guidelines for administration and interpretation of the structured interview. \\
\bottomrule
\end{tabularx}
\caption{Structured interview development framework.}
\label{scale}
\end{table*}

\textbf{Identify behaviors/perceptions that represent the construct or define the domain:} Clearly defining the construct is essential to developing relevant structured interview questions. 
The construct should be operationalized by identifying specific behaviors or attributes that indicate its presence.
In other words, you need to find a theory to guide you on how to measure the target constructs.
This step may involve reviewing literature, conducting expert interviews, or organizing focus groups to understand the various dimensions and observable characteristics of the construct.

\textbf{Prepare a set of structured interview specifications, structured interview blueprint:} A blueprint outlines the structured interview’s structure and content, specifying how questions are distributed across the construct’s dimensions. 
It typically includes the number of questions per domain and the content of the questions.

\textbf{Build an initial question pool:} In this step, an extensive list of questions is created to cover the full range of the construct. 
Question wording should be clear, concise, and relevant to the target population. 
It is common practice to generate more questions than needed to ensure that poorly performing questions can be removed later without compromising the structured interview.

\textbf{Have questions reviewed by substantive experts (and revised as necessary):} SMEs review the question pool for content accuracy, relevance, clarity, and bias. 
Experts assess whether the questions align with the construct’s definition and whether any important aspects are missing.
Feedback from experts helps refine the wording, remove ambiguous questions, and identify questions with potential cultural or gender biases.

\textbf{Hold preliminary question tryouts:} Before large-scale testing, questions are piloted on a small group of individuals representative of the target population. 
This stage helps identify any immediate issues with question comprehension, response format, or instructions. 
It can also include cognitive interviews where participants are asked to explain their thought processes when answering questions.
The feedback from this stage informs further revisions, ensuring that questions are clear and easy to understand.

\textbf{Determine statistical properties of questions (and eliminate poor questions or revise as necessary):} Question performance is assessed through statistical analyses to evaluate difficulty, discrimination, and internal consistency.
Standardized scoring criteria, such as Behaviorally Anchored Rating Scales (BARS), can be used for obtaining question scores.
Then, methods like correlations, factor analysis, and item response theory (IRT) can help determine how effectively each question measures the intended construct.

Questions that demonstrate poor psychometric properties---such as low discrimination or high measurement error---are either revised or removed. 
For example, questions with low correlations with the overall score or incorrect factor loadings may be eliminated.

\textbf{Field-test the structured interview on a large representative sample of the intended examinee population:} The revised item set is administered to a large, representative sample to gather comprehensive data on the scale’s performance. 
This step ensures that the sample reflects the population for which the test is intended, which is critical for generalizability.
Statistical analyses are conducted to refine the test further. 
This process may involve removing redundant items, assessing dimensionality, and ensuring that items work well across demographic subgroups.

\textbf{Design and conduct reliability and validity studies for the final form of the structured interview:} To ensure the structured interview is psychometrically sound, various reliability and validity studies are conducted:
Reliability studies measure the structured interview’s consistency and stability, including internal consistency (e.g., Cronbach's alpha), test-retest reliability, and inter-rater reliability (if applicable).
Validity studies assess whether the structured interview measures what it is intended to measure. 
This includes: Content validity (the extent to which items cover the construct); Construct validity (e.g., convergent and discriminant validity); Criterion-related validity (e.g., predictive or concurrent validity).
These studies provide evidence that the structured interview is both reliable and valid for its intended purpose.

\textbf{Develop guidelines for administration and interpretation of the structured interview:} The final step involves creating a comprehensive, structured interview manual that includes instructions for structured interview administration, scoring procedures, and guidelines for interpreting results. 
This manual ensures consistency in the structured interview's use across different settings and helps minimize errors in administration and scoring.
Guidelines for interpreting scores may include norms, cutoff points, and descriptions of what various score ranges indicate.
\section{Appendix: Psychometric Data Evaluation Framework}
\label{sec:appendix evaluation_framework}

Here, we present the evaluation framework used to assess the fidelity of simulated psychometric data (i.e., how well it aligns with human data).

\begin{figure*}
    \centering
    \includegraphics[width=\linewidth]{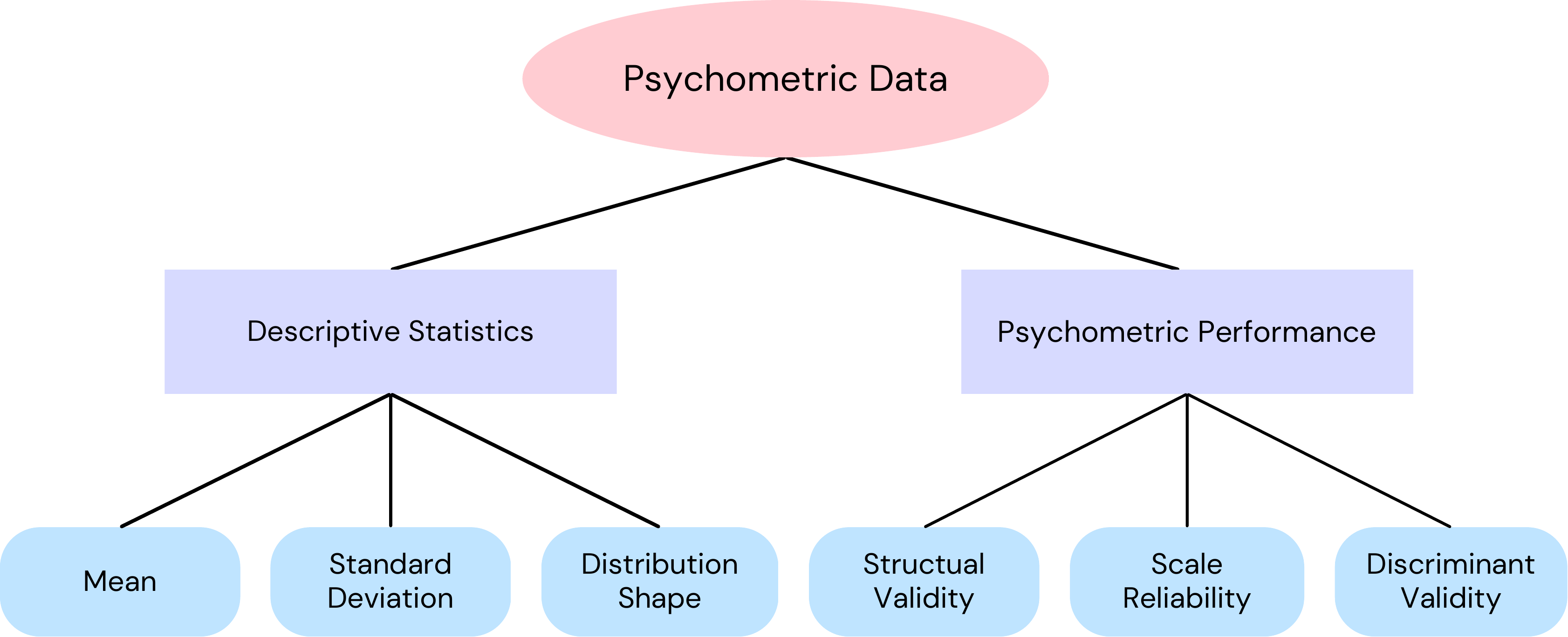}
    \caption{Psychometric data evaluation framework.}
    \label{fig:psychometric_framework}
\end{figure*}

The evaluation is conducted at different levels. 
For example, for personality, it includes Item, Facet, and Domain levels. 
Generally, the structure of psychometric data is hierarchical, where observed responses (Item level) map onto latent traits (Domain level).
Our evaluation framework incorporates both descriptive statistics and psychometric performance metrics to ensure a comprehensive evaluation (see Figure~\ref{fig:psychometric_framework}). 
Below, we outline each component and the rationale for its inclusion.

\subsection{Descriptive Statistics}
Descriptive statistics are primarily used to summarize, outline, and present the basic features of data. 
They help researchers understand the distribution and fundamental trends of the data without interpreting or measuring specific psychological constructs.
Do not forget that the evaluation of psychometric data is hierarchical; we need to evaluate it at both the Item and Domain levels.

\textbf{Mean ($\mu$):}
The mean represents the central tendency of the responses, reflecting the average score across individuals.

\textbf{Standard Deviation ($\sigma$):}
The standard deviation captures the dispersion of responses, indicating how much variation exists within the data.

Both $\mu$ and $\sigma$ can be quantified for similarity to human distribution using MAE and \( r \). However, these metrics provide a more summarized level of comparison; we also need to examine performance on specific items and the domain.

\textbf{Distribution Shape:}
The distribution shape describes the overall pattern of how responses are spread across the scale. It provides insight into whether the data follows a normal distribution or exhibits skewness and kurtosis.

Skewness measures the asymmetry of the distribution. 
A positive skew indicates a longer right tail (more low scores with a few high scores), while a negative skew indicates a longer left tail (more high scores with a few low scores).

Kurtosis captures the ``tailedness'' of the distribution. 
High kurtosis (leptokurtic) suggests heavy tails with more extreme values, while low kurtosis (platykurtic) indicates a flatter distribution with fewer extreme values.

\subsection{Psychometric Performance}
Psychometric performance is primarily used to evaluate the measurement quality of psychological constructs, ensuring that assessments accurately and reliably capture the intended traits. 


\textbf{Structural Validity:}
Structural validity refers to the extent to which the internal structure of a measurement instrument aligns with the theoretical construct it is intended to assess.
It can be assessed through model fit, factor loadings, and inter-factor correlations.

The model fit indices provide an overall assessment of how well the proposed structure aligns with the observed data. 
Ideally, the simulated psychometric data will have a similar model fit compared with the human data.
Common indices include \(\mathbf{\chi^2}\), CFI, TLI, RMSEA, and SRMR.

Factor loadings indicate the extent to which each item represents the intended construct, while inter-factor correlations reveal relationships between latent variables.
The simulated psychometric data should also resemble human data in these two aspects.
We can use TCC as a summarized level of comparison; however, it is typically considered too lenient, so we need to examine factor loadings and inter-factor correlation values more closely.

\textbf{Scale Reliability:}
Scale reliability assesses the internal consistency of items measuring the same construct. 
Cronbach’s alpha is a widely used reliability coefficient,
The simulated psychometric data should also resemble human data in this aspect.

\textbf{Discriminant Validity:}
Discriminant validity ensures that distinct constructs are not excessively correlated. 
It can be examined by calculating the mean absolute correlation.

\section{Appendix: Additional Analyses and Results}
\label{sec:appendix results}

\subsection{Additional Response Similarity Results}
\label{sec:appendix results.01}

\subsection{Additional Human Personality Distribution Simulation Results}
\label{sec:appendix results.1}

\subsubsection{Additional Descriptive Statistics Results}
Here we show the detailed $\mu$ and $\sigma$ for human responses and LLM responses at the Item, Facet, and Domain levels, see Tables~\ref{tab:Mistral_d}-\ref{tab:GPT-4o}.
We also present detail MAE results in Table~\ref{tab:mae_results}.

\subsubsection{Additional Psychometric Performance Results}
Table~\ref{tab:modelfit} presents the model fit information for both the TFMs and FFM. 
TCC results are reported in Tables~\ref{tcc_tfm_e}-\ref{tab:tcc_ffm}, while factor loading MAE results appear in Tables~\ref{tab:mae_extraversion}-\ref{tab:mae_ffm}. 
Specific standardized factor loadings for all TFMs are shown in Tables~\ref{tab:fc_sociability}-\ref{tab:fc_creative_imagination}, while standardized factor loadings for the FFM are shown in Tables~\ref{tab:ffm_extraversion}-\ref{tab:ffm_openness}.
Inter-factor correlations are provided in Tables~\ref{tab:inter_extraversion}-\ref{tab:inter-factor-ffm}.

\textbf{Scale Reliability:}
Facet level and domain level Cronbach’s alpha for different method LLM responses on BFI-2 and human responses are shown in Tables~\ref{tab:alpha_extraversion}-\ref{tab:alpha_domain}.

It can be observed that the PSI method, compared to the Persona and Shape methods, performs closer to the results of the human sample in terms of Cronbach's alpha (with the number of data marked in italics and bold being the smallest). 
This indicates that the LLM personality data generated by the PSI method holds an advantage in consistency and reliability, enabling it to more accurately simulate the statistical characteristics of human samples.

\textbf{Discriminant Validity:}
The results for discriminant validity are shown in Table~\ref{tab:dimension_correlation}.
We can further observe that, compared to the Persona and Shape methods, the PSI method demonstrates the closest performance to human samples in terms of discriminant validity.
Specifically, the PSI method shows the closest mean of absolute values when examining its correlations with human samples.

Both higher and lower levels of external validity reveal the degree of differences between the methods and human samples.
Higher external validity indicates that the Big Five factors in human samples are more distinctly differentiated from one another, while lower external validity suggests the opposite.
Therefore, our focus here is on identifying the approach that most closely aligns with the performance of human samples.

\subsection{Additional Personality-Related Behavioral Performance Results}
\label{sec:appendix results.2}

Table~\ref{tab:ocb} and \ref{tab:cwb} present the correlations between the personality dimensions and OCB/CWB reported by all model+PSI simulations, with a comparison to human self-reported data.

\subsection{Additional Abalation Test Results}
\label{sec:appendix results.3}

\begin{figure*}[h]
    \centering
    \includegraphics[width=0.945\linewidth]{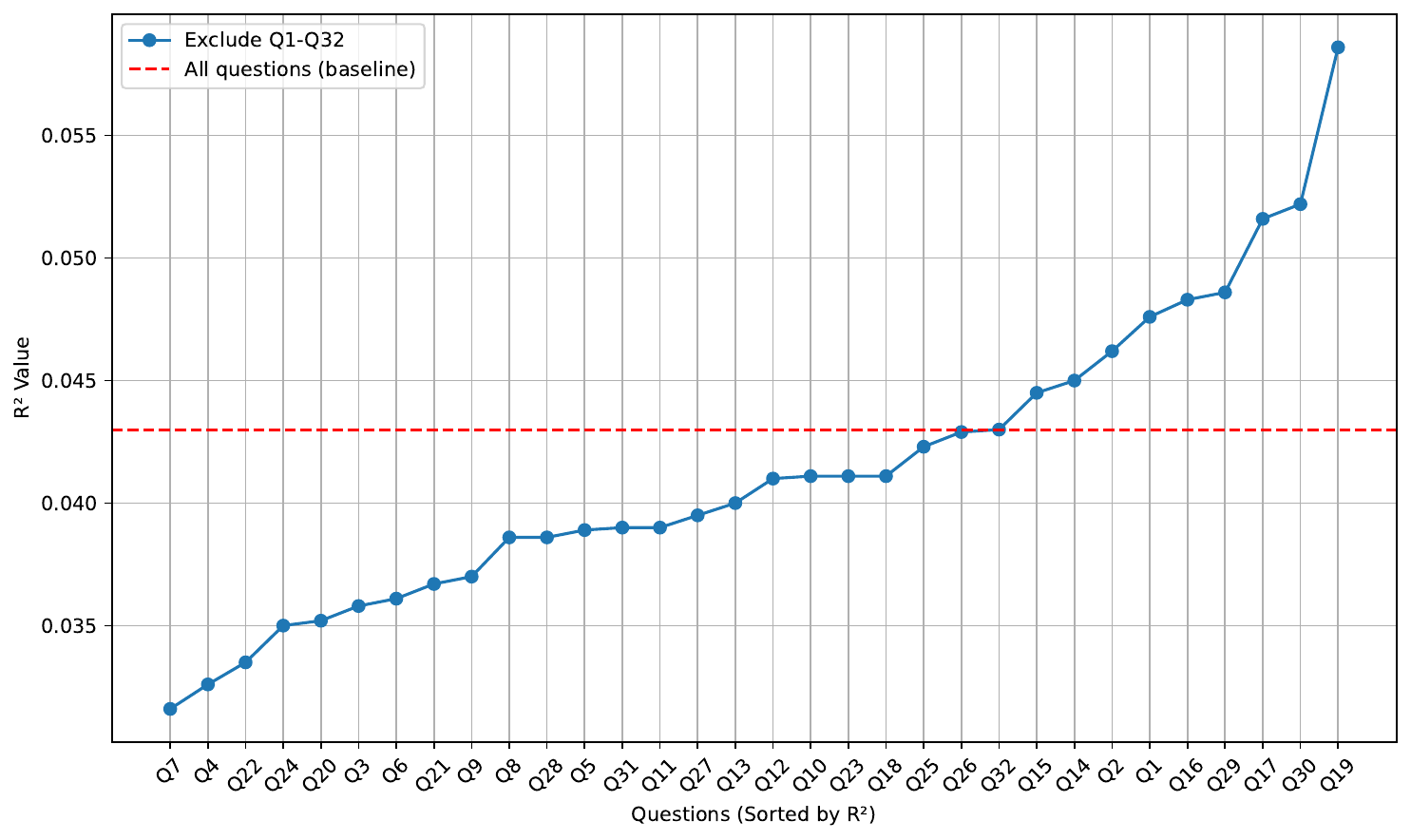}
    \caption{Ablation test results for predicting Openness. The x-axis represents each removed question, and its corresponding response, and the y-axis shows the $R^2$ value. The red dashed line indicates the baseline $R^2$ when all questions are included.}
    \label{fig:o_ablation}
\end{figure*}

\begin{figure*}[h]
    \centering
    \includegraphics[width=0.945\linewidth]{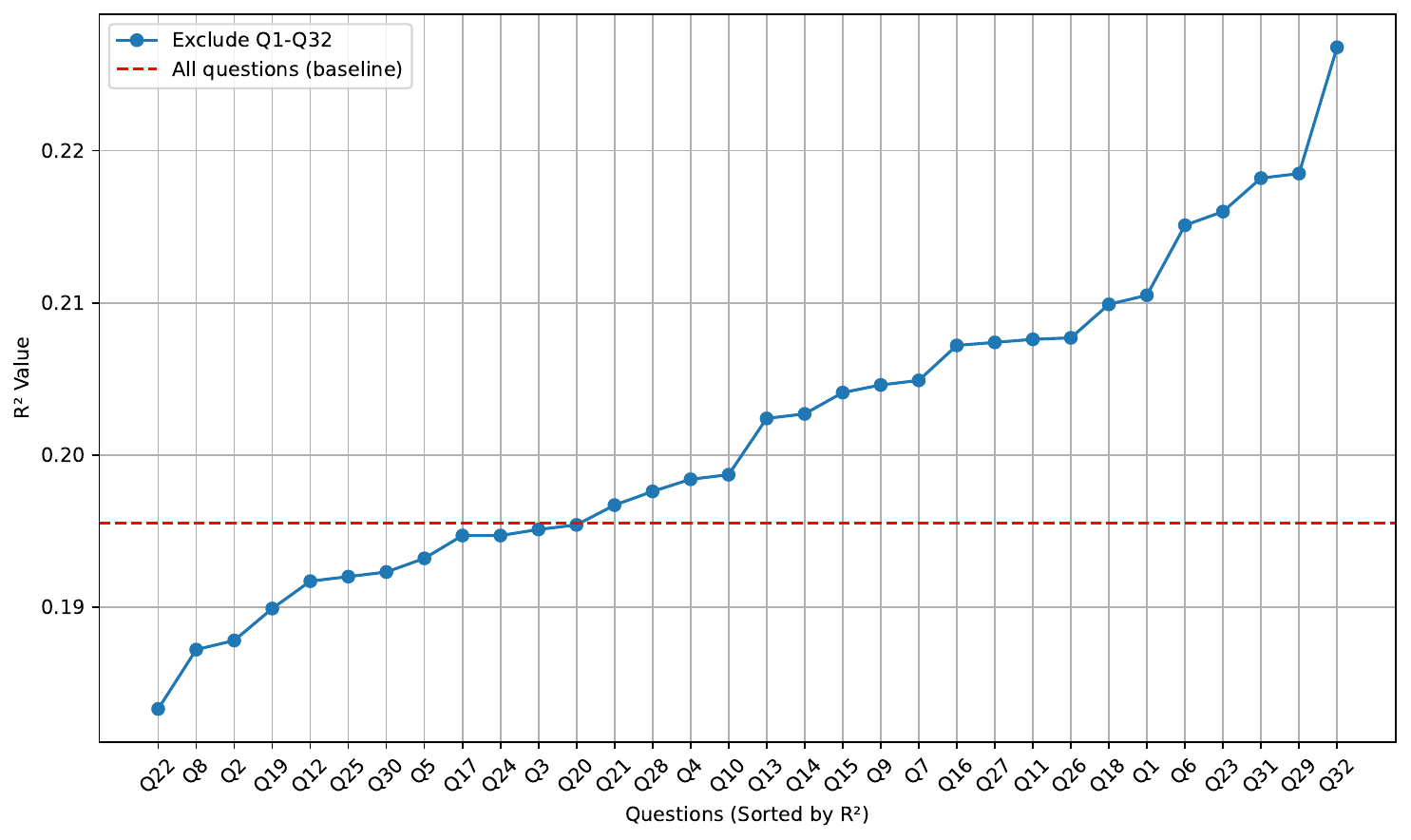}
    \caption{Ablation test results for predicting Extraversion. The x-axis represents each removed question, and its corresponding response, and the y-axis shows the $R^2$ value. The red dashed line indicates the baseline $R^2$ when all questions are included.}
    \label{fig:e_ablation}
\end{figure*}

\begin{figure*}[h]
    \centering
    \includegraphics[width=0.945\linewidth]{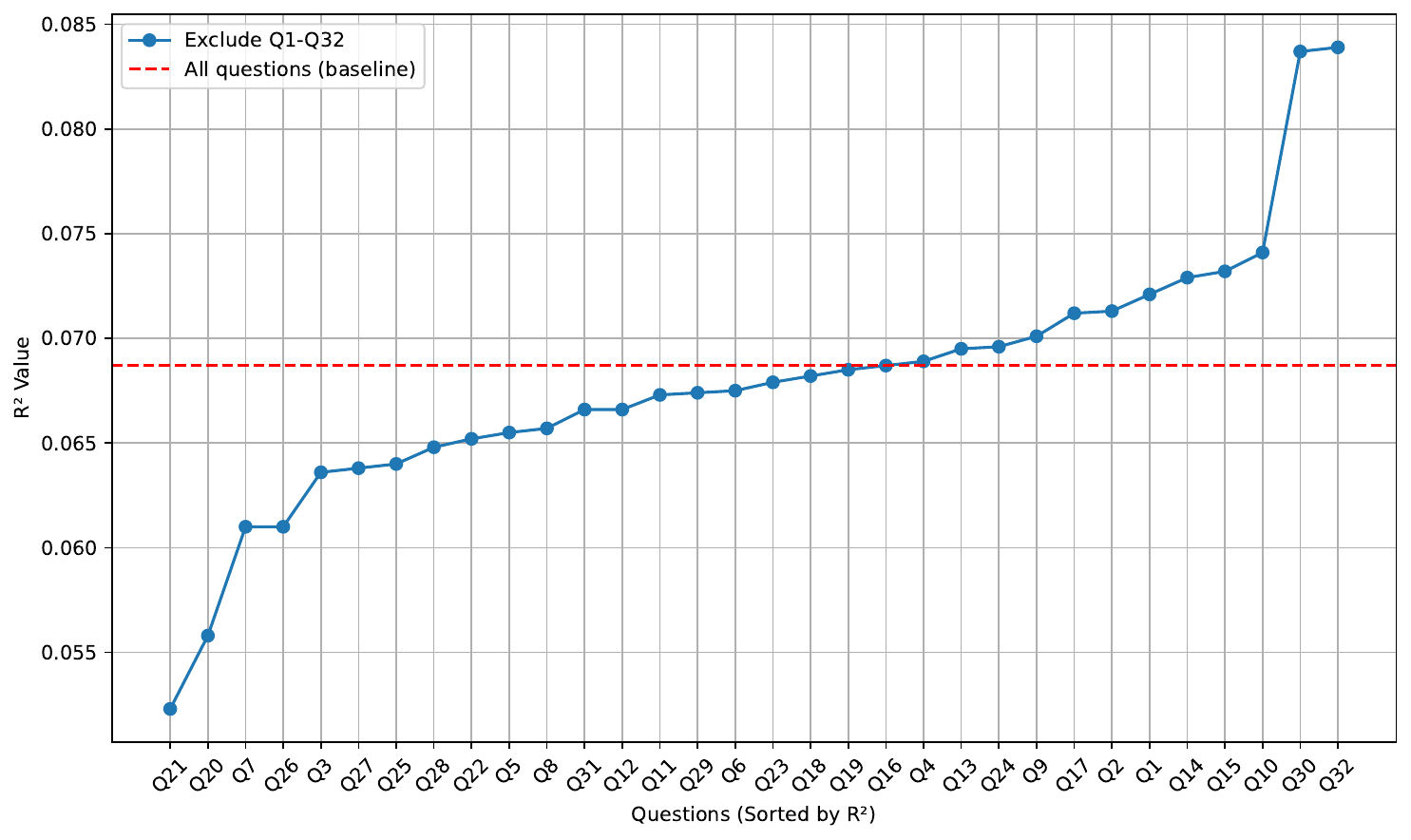}
    \caption{Ablation test results for predicting Agreeableness. The x-axis represents each removed question, and its corresponding response, and the y-axis shows the $R^2$ value. The red dashed line indicates the baseline $R^2$ when all questions are included.}
    \label{fig:a_ablation}
\end{figure*}

\begin{figure*}[h]
    \centering
    \includegraphics[width=0.945\linewidth]{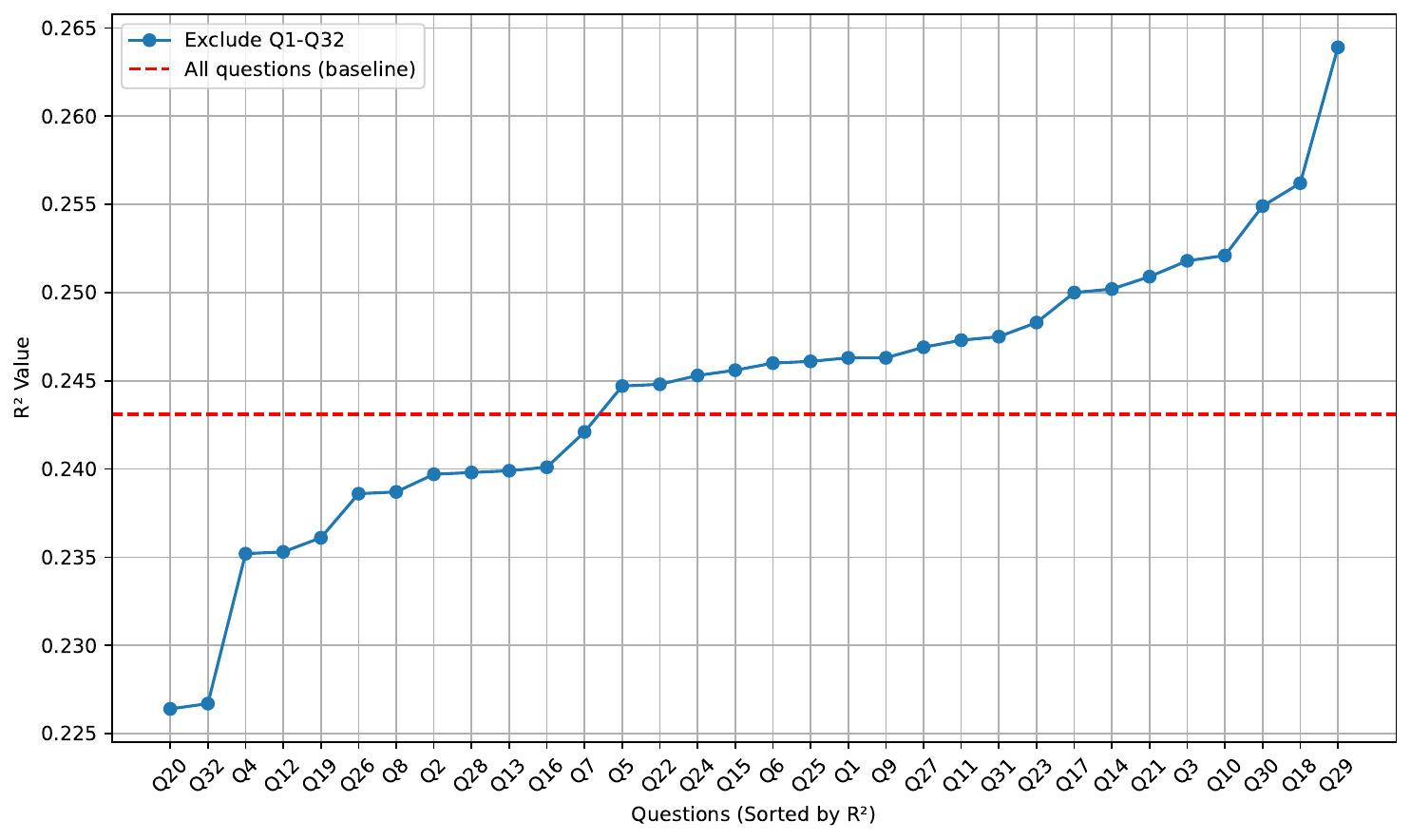}
    \caption{Ablation test results for predicting Neuroticism. The x-axis represents each removed question, and its corresponding response, and the y-axis shows the $R^2$ value. The red dashed line indicates the baseline $R^2$ when all questions are included.}
    \label{fig:n_ablation}
\end{figure*}

Figure~\ref{fig:o_ablation}-\ref{fig:n_ablation} present the relative importance of individual questions for different prediction domains.

\begin{table*}[h]
\resizebox{\linewidth}{!}{
\begin{tabular}{l *{5}{c} *{5}{c}}
\toprule
& \multicolumn{5}{c}{\textbf{MAE}} & \multicolumn{5}{c}{\textbf{$r$}} \\
\cmidrule(lr){2-6} \cmidrule(lr){7-11}
\textbf{Model} & Ext & Agr & Neu & Con & Ope & Ext & Agr & Neu & Con & Ope \\
\midrule
Mistral-7B+PSI      & 0.57 & 0.66 & 0.60 & 0.57 & 0.58 & 0.56 & 0.34 & 0.56 & 0.42 & 0.36 \\
Gemma-2-9B+PSI             & 0.62 & 0.65 & 0.79 & 0.73 & 0.95 & 0.55 & 0.37 & 0.50 & 0.39 & 0.34 \\
Gemma-2-27B+PSI            & 0.76 & 0.50 & 0.71 & 0.65 & 1.09 & 0.48 & 0.35 & 0.54 & 0.40 & 0.40 \\
Llama3-8B+PSI        & 0.62 & 0.51 & 0.61 & 0.72 & 0.93 & 0.60 & 0.40 & 0.62 & 0.42 & 0.44 \\
Llama3-70B+PSI       & 0.75 & 0.56 & 0.59 & 0.66 & 1.27 & 0.43 & 0.36 & 0.57 & 0.40 & 0.39 \\
GPT-4o-mini+PSI                & 0.60 & 0.55 & 0.63 & 0.56 & 0.81 & 0.63 & 0.41 & 0.63 & 0.48 & 0.43 \\
GPT-4o+PSI                     & 0.58 & 0.53 & 0.63 & 0.59 & 0.80 & 0.64 & 0.41 & 0.63 & 0.46 & 0.43 \\
GPT-4o+Life Interview~\citep{park2024generative} & 0.72 & 0.60 & 0.75 & 0.63 & 0.62 & 0.45 & 0.35 & 0.68 & 0.52 & 0.39 \\
\bottomrule
\end{tabular}}
\caption{MAE and \( r \) of human and LLM-simulated data across seven Models+PSI and GPT-4o+Life Interview (see~\citealp{park2024generative}, page 35 Table 3) for BFI-2 each Personality domain.}
\label{tab:exp1}
\end{table*}

\onecolumn
\begin{longtable}{c p{4.15cm} *{8}{S}}
\caption{Mistral-7B descriptive statistics results.}
\label{tab:Mistral_d} \\

\toprule
 & & \multicolumn{4}{c}{\textbf{$\mu$}} & \multicolumn{4}{c}{\textbf{$\sigma$}} \\
\cmidrule(lr){3-6}\cmidrule(lr){7-10}
{\textbf{No.}} & {\textbf{Content}} & {Human} & {Persona} & {Shape} & {PSI} & {Human} & {Persona} & {Shape} & {PSI} \\
\midrule
\endfirsthead 

\multicolumn{10}{@{}l}{\tablename\ \thetable{} -- continued from previous page} \\
\toprule
 & & \multicolumn{4}{c}{\textbf{$\mu$}} & \multicolumn{4}{c}{\textbf{$\sigma$}} \\
\cmidrule(lr){3-6}\cmidrule(lr){7-10}
{\textbf{No.}} & {\textbf{Content}} & {Human} & {Persona} & {Shape} & {PSI} & {Human} & {Persona} & {Shape} & {PSI} \\
\midrule
\endhead 

\midrule
\multicolumn{10}{r@{}}{{Continued on next page}} \\ 
\endfoot

\bottomrule 
\endlastfoot 

\multicolumn{10}{c}{\textbf{Item Level}} \\
\midrule
1  & I am someone who is outgoing, sociable & 3.02 & 3.23 & 3.30 & 2.69 & 1.40 & 1.09 & 1.30 & 1.10 \\
2  & I am someone who is compassionate, has a soft heart & 4.34 & 3.86 & 3.74 & 4.17 & 0.90 & 1.04 & 1.29 & 1.03 \\
3  & I am someone who tends to be disorganized & 3.62 & 2.89 & 3.13 & 3.90 & 1.36 & 0.96 & 1.22 & 0.68 \\
4  & I am someone who is relaxed, handles stress well & 2.56 & 2.28 & 2.38 & 2.73 & 1.30 & 0.69 & 0.85 & 0.93 \\
5  & I am someone who has few artistic interests & 3.68 & 3.73 & 2.96 & 4.07 & 1.33 & 0.54 & 0.78 & 0.26 \\
6  & I am someone who has an assertive personality & 2.96 & 3.61 & 3.26 & 3.34 & 1.38 & 0.71 & 0.89 & 1.02 \\
7  & I am someone who is respectful, treats others with respect & 4.62 & 4.65 & 4.64 & 4.50 & 0.61 & 0.77 & 0.89 & 0.80 \\
8  & I am someone who tends to be lazy & 3.62 & 3.96 & 3.28 & 4.38 & 1.31 & 0.59 & 1.52 & 0.77 \\
9  & I am someone who stays optimistic after experiencing a setback & 2.43 & 2.17 & 2.12 & 1.99 & 1.24 & 0.52 & 0.87 & 0.78 \\
10 & I am someone who is curious about many different things & 4.43 & 4.28 & 4.37 & 3.61 & 0.81 & 0.65 & 0.96 & 0.94 \\
11 & I am someone who rarely feels excited or eager & 3.54 & 3.67 & 3.01 & 3.91 & 1.26 & 0.65 & 1.07 & 0.44 \\
12 & I am someone who tends to find fault with others & 3.34 & 3.63 & 3.28 & 4.45 & 1.32 & 0.60 & 1.47 & 0.58 \\
13 & I am someone who is dependable, steady & 4.33 & 3.85 & 3.77 & 4.04 & 0.86 & 0.90 & 1.09 & 0.92 \\
14 & I am someone who is moody, has up and down mood swings & 2.56 & 3.19 & 2.82 & 2.69 & 1.36 & 0.60 & 1.22 & 0.95 \\
15 & I am someone who is inventive, finds clever ways to do things & 3.90 & 3.68 & 3.48 & 2.82 & 1.03 & 0.83 & 1.09 & 0.98 \\
16 & I am someone who tends to be quiet & 2.35 & 2.83 & 2.47 & 3.21 & 1.34 & 0.68 & 1.15 & 1.06 \\
17 & I am someone who feels little sympathy for others & 3.71 & 3.58 & 3.62 & 4.54 & 1.45 & 0.94 & 1.51 & 0.50 \\
18 & I am someone who is systematic, likes to keep things in order & 3.96 & 2.84 & 3.34 & 2.49 & 1.06 & 0.91 & 0.82 & 0.97 \\
19 & I am someone who can be tense & 3.29 & 3.04 & 3.00 & 3.63 & 1.27 & 0.87 & 0.95 & 0.78 \\
20 & I am someone who is fascinated by art, music, or literature & 4.06 & 3.44 & 4.34 & 2.03 & 1.16 & 1.02 & 1.16 & 1.04 \\
21 & I am someone who is dominant, acts as a leader & 2.90 & 2.43 & 2.46 & 1.99 & 1.35 & 0.80 & 1.05 & 0.68 \\
22 & I am someone who starts arguments with others & 4.23 & 4.54 & 3.72 & 4.96 & 1.05 & 0.87 & 1.75 & 0.23 \\
23 & I am someone who has difficulty getting started on tasks & 3.31 & 3.43 & 2.78 & 3.74 & 1.39 & 0.77 & 1.22 & 0.70 \\
24 & I am someone who feels secure, comfortable with self & 2.26 & 1.90 & 1.97 & 2.06 & 1.27 & 0.70 & 0.97 & 0.90 \\
25 & I am someone who avoids intellectual, philosophical discussions & 3.94 & 3.11 & 2.72 & 3.96 & 1.21 & 0.74 & 1.14 & 0.42 \\
26 & I am someone who is less active than other people & 3.35 & 3.37 & 2.79 & 3.47 & 1.31 & 0.97 & 1.16 & 0.98 \\
27 & I am someone who has a forgiving nature & 3.79 & 3.36 & 3.51 & 3.72 & 1.20 & 0.49 & 0.94 & 0.75 \\
28 & I am someone who can be somewhat careless & 3.56 & 2.85 & 3.12 & 3.40 & 1.25 & 0.92 & 0.98 & 0.95 \\
29 & I am someone who is emotionally stable, not easily upset & 2.44 & 2.62 & 2.27 & 2.93 & 1.27 & 0.56 & 0.98 & 0.85 \\
30 & I am someone who has little creativity & 3.95 & 3.73 & 3.12 & 4.05 & 1.20 & 0.45 & 0.82 & 0.22 \\
31 & I am someone who is sometimes shy, introverted & 2.35 & 2.15 & 2.56 & 3.04 & 1.36 & 0.45 & 0.66 & 0.94 \\
32 & I am someone who is helpful and unselfish with others & 4.18 & 3.69 & 4.15 & 4.03 & 0.88 & 0.76 & 1.14 & 0.92 \\
33 & I am someone who keeps things neat and tidy & 3.71 & 3.20 & 3.50 & 2.32 & 1.24 & 0.84 & 1.05 & 0.60 \\
34 & I am someone who worries a lot & 3.30 & 3.68 & 3.30 & 3.60 & 1.47 & 0.65 & 1.07 & 1.00 \\
35 & I am someone who values art and beauty & 4.16 & 3.94 & 4.45 & 3.11 & 1.07 & 0.93 & 0.89 & 1.17 \\
36 & I am someone who finds it hard to influence people & 3.09 & 3.38 & 2.79 & 3.94 & 1.20 & 0.55 & 0.85 & 0.32 \\
37 & I am someone who is sometimes rude to others & 3.79 & 3.65 & 3.52 & 4.27 & 1.25 & 0.78 & 1.48 & 0.66 \\
38 & I am someone who is efficient, gets things done & 4.23 & 3.52 & 3.70 & 3.40 & 0.88 & 0.91 & 1.08 & 1.10 \\
39 & I am someone who often feels sad & 2.71 & 3.46 & 3.19 & 3.07 & 1.43 & 0.83 & 1.14 & 1.08 \\
40 & I am someone who is complex, a deep thinker & 4.02 & 4.34 & 3.88 & 4.17 & 1.08 & 0.70 & 0.93 & 0.73 \\
41 & I am someone who is full of energy & 3.14 & 3.75 & 3.50 & 3.18 & 1.30 & 0.97 & 1.36 & 1.20 \\
42 & I am someone who is suspicious of others’ intentions & 2.75 & 3.19 & 3.28 & 4.25 & 1.30 & 0.49 & 1.21 & 0.71 \\
43 & I am someone who is reliable, can always be counted on & 4.32 & 3.79 & 3.84 & 4.13 & 0.88 & 0.94 & 1.15 & 0.95 \\
44 & I am someone who keeps their emotions under control & 2.22 & 2.88 & 2.35 & 3.31 & 1.14 & 0.42 & 0.90 & 0.79 \\
45 & I am someone who has difficulty imagining things & 4.17 & 3.09 & 2.86 & 3.89 & 1.06 & 0.37 & 0.72 & 0.37 \\
46 & I am someone who is talkative & 2.87 & 3.04 & 2.94 & 3.20 & 1.38 & 0.96 & 1.01 & 1.19 \\
47 & I am someone who can be cold and uncaring & 3.84 & 3.66 & 3.47 & 4.26 & 1.23 & 0.63 & 1.27 & 0.72 \\
48 & I am someone who leaves a mess, doesn’t clean up & 4.15 & 3.44 & 3.07 & 4.79 & 1.12 & 1.24 & 1.90 & 0.45 \\
49 & I am someone who rarely feels anxious or afraid & 3.44 & 3.00 & 2.27 & 3.34 & 1.37 & 0.29 & 1.17 & 0.73 \\
50 & I am someone who thinks poetry and plays are boring & 3.66 & 2.65 & 2.66 & 3.46 & 1.37 & 1.34 & 1.39 & 1.68 \\
51 & I am someone who prefers to have others take charge & 2.96 & 3.26 & 2.86 & 4.11 & 1.28 & 0.75 & 1.49 & 0.55 \\
52 & I am someone who is polite, courteous to others & 4.50 & 4.21 & 4.55 & 4.30 & 0.70 & 0.89 & 0.88 & 0.87 \\
53 & I am someone who is persistent, works until the task is finished & 4.28 & 4.36 & 4.09 & 4.55 & 0.91 & 0.68 & 1.03 & 0.66 \\
54 & I am someone who tends to feel depressed, blue & 2.65 & 3.37 & 2.85 & 2.85 & 1.46 & 0.84 & 1.36 & 1.14 \\
55 & I am someone who has little interest in abstract ideas & 3.83 & 3.32 & 3.03 & 3.97 & 1.22 & 0.55 & 0.61 & 0.25 \\
56 & I am someone who shows a lot of enthusiasm & 3.35 & 4.39 & 3.70 & 3.71 & 1.22 & 0.75 & 1.30 & 1.18 \\
57 & I am someone who assumes the best about people & 3.37 & 3.39 & 3.67 & 3.44 & 1.28 & 0.53 & 0.87 & 0.89 \\
58 & I am someone who sometimes behaves irresponsibly & 3.64 & 3.05 & 3.24 & 3.81 & 1.30 & 0.99 & 0.99 & 0.76 \\
59 & I am someone who is temperamental, gets emotional easily & 2.53 & 3.17 & 2.79 & 2.15 & 1.33 & 0.96 & 1.42 & 0.92 \\
60 & I am someone who is original, comes up with new ideas & 3.84 & 3.67 & 3.36 & 2.96 & 1.07 & 0.91 & 1.17 & 1.11 \\
\midrule
\multicolumn{10}{c}{\textbf{Facet Level}} \\
\midrule
1  & Sociability & 2.65 & 2.81 & 2.82 & 3.04 & 1.17 & 0.59 & 0.85 & 0.89 \\
2  & Assertiveness & 2.98 & 3.17 & 2.84 & 3.34 & 1.04 & 0.46 & 0.86 & 0.47 \\
3  & Energy Level & 3.35 & 3.79 & 3.25 & 3.57 & 0.95 & 0.66 & 1.05 & 0.73 \\
4  & Compassion & 4.02 & 3.70 & 3.74 & 4.25 & 0.81 & 0.56 & 1.03 & 0.65 \\
5  & Respectfulness & 4.29 & 4.26 & 4.11 & 4.51 & 0.70 & 0.64 & 1.01 & 0.55 \\
6  & Trust & 3.31 & 3.39 & 3.44 & 3.97 & 1.00 & 0.36 & 0.93 & 0.57 \\
7  & Organization & 3.86 & 3.09 & 3.26 & 3.37 & 1.01 & 0.67 & 0.94 & 0.50 \\
8  & Productiveness & 3.86 & 3.82 & 3.46 & 4.02 & 0.91 & 0.57 & 1.05 & 0.64 \\
9  & Responsibility & 3.96 & 3.39 & 3.49 & 3.85 & 0.84 & 0.76 & 0.87 & 0.75 \\
10 & Anxiety & 3.15 & 3.00 & 2.74 & 3.32 & 1.11 & 0.43 & 0.70 & 0.66 \\
11 & Depression & 2.51 & 2.72 & 2.53 & 2.49 & 1.14 & 0.51 & 0.91 & 0.81 \\
12 & Emotional Volatility & 2.44 & 2.96 & 2.56 & 2.77 & 1.10 & 0.45 & 0.90 & 0.72 \\
13 & Intellectual Curiosity & 4.05 & 3.76 & 3.50 & 3.93 & 0.83 & 0.48 & 0.65 & 0.41 \\
14 & Aesthetic Sensitivity & 3.89 & 3.44 & 3.60 & 3.17 & 1.01 & 0.67 & 0.86 & 0.73 \\
15 & Creative Imagination & 3.96 & 3.54 & 3.20 & 3.43 & 0.88 & 0.47 & 0.77 & 0.53 \\
\midrule
\multicolumn{10}{c}{\textbf{Domain Level}} \\
\midrule
1 & Extraversion & 2.99 & 3.26 & 2.97 & 3.32 & 0.85 & 0.47 & 0.84 & 0.62 \\
2 & Agreeableness & 3.87 & 3.78 & 3.76 & 4.24 & 0.69 & 0.45 & 0.92 & 0.54 \\
3 & Conscientiousness & 3.89 & 3.43 & 3.41 & 3.75 & 0.80 & 0.59 & 0.90 & 0.56 \\
4 & Neuroticism & 2.70 & 2.90 & 2.61 & 2.86 & 1.02 & 0.38 & 0.75 & 0.65 \\
5 & Openness & 3.97 & 3.58 & 3.43 & 3.51 & 0.76 & 0.44 & 0.70 & 0.45 \\
\end{longtable}

\begin{longtable}{c p{4.15cm} *{8}{S}}
\caption{Gemma-2-9B descriptive statistics results.}
\label{tab:Gemma-2-9B_d} \\

\toprule
 & & \multicolumn{4}{c}{\textbf{$\mu$}} & \multicolumn{4}{c}{\textbf{$\sigma$}} \\
\cmidrule(lr){3-6}\cmidrule(lr){7-10}
{\textbf{No.}} & {\textbf{Content}} & {Human} & {Persona} & {Shape} & {PSI} & {Human} & {Persona} & {Shape} & {PSI} \\
\midrule
\endfirsthead 

\multicolumn{10}{@{}l}{\tablename\ \thetable{} -- continued from previous page} \\
\toprule
 & & \multicolumn{4}{c}{\textbf{$\mu$}} & \multicolumn{4}{c}{\textbf{$\sigma$}} \\
\cmidrule(lr){3-6}\cmidrule(lr){7-10}
{\textbf{No.}} & {\textbf{Content}} & {Human} & {Persona} & {Shape} & {PSI} & {Human} & {Persona} & {Shape} & {PSI} \\
\midrule
\endhead 

\midrule
\multicolumn{10}{r@{}}{{Continued on next page}} \\ 
\endfoot

\bottomrule 
\endlastfoot 

\multicolumn{10}{c}{\textbf{Item Level}} \\
\midrule
1  & I am someone who is outgoing, sociable                         & 3.02 & 2.98 & 2.46 & 2.26 & 1.40 & 0.79 & 1.22 & 0.80 \\
2  & I am someone who is compassionate, has a soft heart            & 4.34 & 3.50 & 3.09 & 3.44 & 0.90 & 0.82 & 1.31 & 0.93 \\
3  & I am someone who tends to be disorganized                      & 3.62 & 2.97 & 3.05 & 2.95 & 1.36 & 0.39 & 1.03 & 0.37 \\
4  & I am someone who is relaxed, handles stress well               & 2.56 & 3.09 & 3.33 & 3.59 & 1.30 & 0.62 & 1.09 & 0.65 \\
5  & I am someone who has few artistic interests                   & 3.68 & 3.15 & 3.11 & 2.76 & 1.33 & 0.72 & 1.01 & 0.62 \\
6  & I am someone who has an assertive personality                 & 2.96 & 3.10 & 2.93 & 2.38 & 1.38 & 0.62 & 1.15 & 0.72 \\
7  & I am someone who is respectful, treats others with respect    & 4.62 & 3.99 & 3.71 & 3.51 & 0.61 & 1.07 & 1.42 & 0.72 \\
8  & I am someone who tends to be lazy                             & 3.62 & 3.38 & 3.37 & 3.36 & 1.31 & 0.67 & 1.27 & 0.80 \\
9  & I am someone who stays optimistic after experiencing a setback & 2.43 & 2.67 & 2.72 & 2.82 & 1.24 & 0.74 & 1.48 & 0.94 \\
10 & I am someone who is curious about many different things       & 4.43 & 3.65 & 3.27 & 3.01 & 0.81 & 0.69 & 1.40 & 0.62 \\
11 & I am someone who rarely feels excited or eager                & 3.54 & 3.57 & 3.02 & 2.85 & 1.26 & 0.72 & 1.57 & 0.96 \\
12 & I am someone who tends to find fault with others               & 3.34 & 3.05 & 3.31 & 3.15 & 1.32 & 0.32 & 1.29 & 0.52 \\
13 & I am someone who is dependable, steady                         & 4.33 & 3.58 & 3.10 & 3.55 & 0.86 & 0.78 & 1.40 & 0.73 \\
14 & I am someone who is moody, has up and down mood swings        & 2.56 & 3.16 & 2.85 & 3.10 & 1.36 & 0.49 & 1.31 & 0.71 \\
15 & I am someone who is inventive, finds clever ways to do things & 3.90 & 3.29 & 2.89 & 2.95 & 1.03 & 0.52 & 1.13 & 0.50 \\
16 & I am someone who tends to be quiet                             & 2.35 & 2.99 & 2.97 & 2.56 & 1.34 & 0.52 & 1.08 & 0.98 \\
17 & I am someone who feels little sympathy for others             & 3.71 & 3.64 & 3.64 & 3.41 & 1.45 & 0.90 & 1.45 & 1.00 \\
18 & I am someone who is systematic, likes to keep things in order & 3.96 & 3.27 & 2.98 & 2.82 & 1.06 & 0.66 & 1.14 & 0.69 \\
19 & I am someone who can be tense                                 & 3.29 & 3.37 & 3.36 & 3.42 & 1.27 & 0.61 & 1.18 & 0.68 \\
20 & I am someone who is fascinated by art, music, or literature   & 4.06 & 3.49 & 2.95 & 2.94 & 1.16 & 0.82 & 1.16 & 0.83 \\
21 & I am someone who is dominant, acts as a leader               & 2.90 & 3.03 & 3.08 & 2.46 & 1.35 & 0.57 & 1.13 & 0.73 \\
22 & I am someone who starts arguments with others                & 4.23 & 3.61 & 3.74 & 3.52 & 1.05 & 0.86 & 1.40 & 0.84 \\
23 & I am someone who has difficulty getting started on tasks     & 3.31 & 3.00 & 3.11 & 2.81 & 1.39 & 0.26 & 1.21 & 0.47 \\
24 & I am someone who feels secure, comfortable with self         & 2.26 & 2.73 & 3.00 & 3.45 & 1.27 & 0.82 & 1.40 & 0.74 \\
25 & I am someone who avoids intellectual, philosophical discussions & 3.94 & 3.02 & 2.93 & 2.85 & 1.21 & 0.28 & 1.19 & 0.63 \\
26 & I am someone who is less active than other people            & 3.35 & 3.21 & 3.20 & 2.74 & 1.31 & 0.70 & 1.12 & 0.52 \\
27 & I am someone who has a forgiving nature                     & 3.79 & 3.03 & 3.20 & 3.03 & 1.20 & 0.36 & 1.16 & 0.59 \\
28 & I am someone who can be somewhat careless                   & 3.56 & 2.90 & 2.66 & 3.10 & 1.25 & 0.42 & 1.00 & 0.64 \\
29 & I am someone who is emotionally stable, not easily upset    & 2.44 & 3.11 & 2.87 & 3.31 & 1.27 & 0.53 & 1.28 & 0.78 \\
30 & I am someone who has little creativity                      & 3.95 & 3.01 & 3.28 & 3.01 & 1.20 & 0.49 & 1.13 & 0.49 \\
31 & I am someone who is sometimes shy, introverted              & 2.35 & 2.88 & 3.04 & 2.89 & 1.36 & 0.57 & 1.06 & 0.87 \\
32 & I am someone who is helpful and unselfish with others       & 4.18 & 3.24 & 3.28 & 3.51 & 0.88 & 0.76 & 1.50 & 0.76 \\
33 & I am someone who keeps things neat and tidy                 & 3.71 & 2.97 & 2.89 & 3.03 & 1.24 & 0.44 & 1.02 & 0.36 \\
34 & I am someone who worries a lot                              & 3.30 & 3.20 & 2.80 & 3.52 & 1.47 & 0.61 & 1.31 & 1.06 \\
35 & I am someone who values art and beauty                     & 4.16 & 3.56 & 3.16 & 3.22 & 1.07 & 0.81 & 1.18 & 0.64 \\
36 & I am someone who finds it hard to influence people         & 3.09 & 3.01 & 3.04 & 2.89 & 1.20 & 0.32 & 1.06 & 0.47 \\
37 & I am someone who is sometimes rude to others               & 3.79 & 3.09 & 3.49 & 3.29 & 1.25 & 0.50 & 1.32 & 0.79 \\
38 & I am someone who is efficient, gets things done            & 4.23 & 3.39 & 3.25 & 3.33 & 0.88 & 0.69 & 1.24 & 0.66 \\
39 & I am someone who often feels sad                          & 2.71 & 3.09 & 2.65 & 3.38 & 1.43 & 0.52 & 1.21 & 0.72 \\
40 & I am someone who is complex, a deep thinker                & 4.02 & 3.40 & 3.00 & 2.72 & 1.08 & 0.67 & 1.04 & 0.87 \\
41 & I am someone who is full of energy                         & 3.14 & 3.47 & 2.79 & 2.61 & 1.30 & 0.85 & 1.14 & 0.87 \\
42 & I am someone who is suspicious of others’ intentions       & 2.75 & 2.93 & 3.15 & 2.90 & 1.30 & 0.39 & 1.28 & 0.52 \\
43 & I am someone who is reliable, can always be counted on     & 4.32 & 3.49 & 3.37 & 3.47 & 0.88 & 0.86 & 1.49 & 0.75 \\
44 & I am someone who keeps their emotions under control       & 2.22 & 3.12 & 2.61 & 3.26 & 1.14 & 0.47 & 1.37 & 0.66 \\
45 & I am someone who has difficulty imagining things          & 4.17 & 3.10 & 3.45 & 3.06 & 1.06 & 0.38 & 1.09 & 0.42 \\
46 & I am someone who is talkative                             & 2.87 & 3.02 & 3.03 & 2.50 & 1.38 & 0.38 & 0.89 & 0.88 \\
47 & I am someone who can be cold and uncaring                 & 3.84 & 3.42 & 3.26 & 3.45 & 1.23 & 0.73 & 1.48 & 1.10 \\
48 & I am someone who leaves a mess, doesn’t clean up         & 4.15 & 3.27 & 3.06 & 3.16 & 1.12 & 0.62 & 1.44 & 0.51 \\
49 & I am someone who rarely feels anxious or afraid          & 3.44 & 3.18 & 2.77 & 3.64 & 1.37 & 0.48 & 1.26 & 0.83 \\
50 & I am someone who thinks poetry and plays are boring      & 3.66 & 3.53 & 3.34 & 3.12 & 1.37 & 0.57 & 1.45 & 0.43 \\
51 & I am someone who prefers to have others take charge      & 2.96 & 3.29 & 2.84 & 3.05 & 1.28 & 0.62 & 1.46 & 0.78 \\
52 & I am someone who is polite, courteous to others          & 4.50 & 3.55 & 3.65 & 3.43 & 0.70 & 0.93 & 1.40 & 0.75 \\
53 & I am someone who is persistent, works until finished    & 4.28 & 3.65 & 3.49 & 3.63 & 0.91 & 0.71 & 1.22 & 0.65 \\
54 & I am someone who tends to feel depressed, blue          & 2.65 & 3.09 & 2.63 & 3.40 & 1.46 & 0.46 & 1.30 & 0.72 \\
55 & I am someone who has little interest in abstract ideas  & 3.83 & 3.04 & 2.87 & 2.92 & 1.22 & 0.40 & 1.14 & 0.63 \\
56 & I am someone who shows a lot of enthusiasm              & 3.35 & 3.67 & 2.71 & 2.93 & 1.22 & 0.84 & 1.34 & 0.94 \\
57 & I am someone who assumes the best about people          & 3.37 & 3.21 & 3.39 & 2.93 & 1.28 & 0.64 & 1.40 & 0.75 \\
58 & I am someone who sometimes behaves irresponsibly        & 3.64 & 2.88 & 2.81 & 2.82 & 1.30 & 0.40 & 1.07 & 0.62 \\
59 & I am someone who is temperamental, gets emotional easily & 2.53 & 2.90 & 2.72 & 2.61 & 1.33 & 0.52 & 1.34 & 0.76 \\
60 & I am someone who is original, comes up with new ideas  & 3.84 & 3.10 & 2.77 & 2.72 & 1.07 & 0.52 & 1.23 & 0.55 \\
\midrule
\multicolumn{10}{c}{\textbf{Facet Level}} \\
\midrule
1  & Sociability           & 2.65 & 2.97 & 2.88 & 2.55 & 1.17 & 0.47 & 0.84 & 0.74 \\
2  & Assertiveness         & 2.98 & 3.11 & 2.97 & 2.69 & 1.04 & 0.40 & 0.90 & 0.49 \\
3  & Energy Level          & 3.35 & 3.48 & 2.93 & 2.78 & 0.95 & 0.61 & 1.17 & 0.52 \\
4  & Compassion            & 4.02 & 3.45 & 3.32 & 3.46 & 0.81 & 0.68 & 1.35 & 0.79 \\
5  & Respectfulness        & 4.29 & 3.56 & 3.65 & 3.44 & 0.70 & 0.67 & 1.30 & 0.55 \\
6  & Trust                 & 3.31 & 3.06 & 3.26 & 3.00 & 1.00 & 0.32 & 1.21 & 0.41 \\
7  & Organization          & 3.86 & 3.12 & 3.00 & 2.99 & 1.01 & 0.41 & 0.92 & 0.30 \\
8  & Productiveness        & 3.86 & 3.36 & 3.30 & 3.28 & 0.91 & 0.48 & 1.14 & 0.48 \\
9  & Responsibility        & 3.96 & 3.21 & 2.98 & 3.23 & 0.84 & 0.52 & 1.06 & 0.55 \\
10 & Anxiety               & 3.15 & 3.21 & 3.07 & 3.54 & 1.11 & 0.49 & 0.89 & 0.65 \\
11 & Depression            & 2.51 & 2.89 & 2.75 & 3.26 & 1.14 & 0.52 & 1.17 & 0.65 \\
12 & Emotional Volatility  & 2.44 & 3.07 & 2.76 & 3.07 & 1.10 & 0.40 & 1.03 & 0.55 \\
13 & Intellectual Curiosity & 4.05 & 3.28 & 3.02 & 2.87 & 0.83 & 0.36 & 1.05 & 0.51 \\
14 & Aesthetic Sensitivity  & 3.89 & 3.43 & 3.14 & 3.01 & 1.01 & 0.53 & 1.09 & 0.33 \\
15 & Creative Imagination   & 3.96 & 3.13 & 3.10 & 2.93 & 0.88 & 0.32 & 1.02 & 0.30 \\
\midrule
\multicolumn{10}{c}{\textbf{Domain Level}} \\
\midrule
1 & Extraversion     & 2.99 & 3.19 & 2.93 & 2.68 & 0.85 & 0.42 & 0.89 & 0.50 \\
2 & Agreeableness    & 3.87 & 3.35 & 3.41 & 3.30 & 0.69 & 0.51 & 1.25 & 0.52 \\
3 & Conscientiousness & 3.89 & 3.23 & 3.09 & 3.17 & 0.80 & 0.41 & 0.96 & 0.39 \\
4 & Neuroticism      & 2.70 & 3.06 & 2.86 & 3.29 & 1.02 & 0.42 & 0.95 & 0.55 \\
5 & Openness         & 3.97 & 3.28 & 3.08 & 2.94 & 0.76 & 0.33 & 1.00 & 0.30 \\
\end{longtable}

\begin{longtable}{c p{4.15cm} *{8}{S}}
\caption{Gemma-2-27B descriptive statistics results.}

\label{tab:Gemma2-27B-IT} \\

\toprule
 & & \multicolumn{4}{c}{\textbf{$\mu$}} & \multicolumn{4}{c}{\textbf{$\sigma$}} \\
\cmidrule(lr){3-6}\cmidrule(lr){7-10}
{\textbf{No.}} & {\textbf{Content}} & {Human} & {Persona} & {Shape} & {PSI} & {Human} & {Persona} & {Shape} & {PSI} \\
\midrule
\endfirsthead

\multicolumn{10}{@{}l}{\tablename\ \thetable{} -- continued from previous page} \\
\toprule
 & & \multicolumn{4}{c}{\textbf{$\mu$}} & \multicolumn{4}{c}{\textbf{$\sigma$}} \\
\cmidrule(lr){3-6}\cmidrule(lr){7-10}
{\textbf{No.}} & {\textbf{Content}} & {Human} & {Persona} & {Shape} & {PSI} & {Human} & {Persona} & {Shape} & {PSI} \\
\midrule
\endhead

\midrule
\multicolumn{10}{r@{}}{{Continued on next page}} \\
\endfoot

\bottomrule
\endlastfoot

\multicolumn{10}{c}{\textbf{Item Level}} \\
\midrule
1  & I am someone who Is outgoing, sociable                          & 3.02 & 2.92 & 2.50 & 2.31 & 1.40 & 0.73 & 1.15 & 0.70 \\
2  & I am someone who Is compassionate, has a soft heart             & 4.34 & 3.62 & 3.16 & 3.71 & 0.90 & 0.75 & 1.30 & 0.82 \\
3  & I am someone who Tends to be disorganized                       & 3.62 & 3.03 & 3.23 & 3.38 & 1.36 & 0.42 & 1.02 & 0.82 \\
4  & I am someone who Is relaxed, handles stress well                & 2.56 & 3.27 & 3.40 & 3.67 & 1.30 & 0.61 & 1.09 & 0.55 \\
5  & I am someone who Has few artistic interests                    & 3.68 & 3.40 & 2.93 & 2.70 & 1.33 & 0.70 & 0.93 & 0.98 \\
6  & I am someone who Has an assertive personality                  & 2.96 & 3.06 & 2.80 & 2.31 & 1.38 & 0.55 & 1.11 & 0.60 \\
7  & I am someone who Is respectful, treats others with respect     & 4.62 & 4.77 & 4.00 & 4.30 & 0.61 & 0.58 & 1.29 & 0.68 \\
8  & I am someone who Tends to be lazy                              & 3.62 & 3.56 & 3.27 & 3.60 & 1.31 & 0.64 & 1.24 & 0.79 \\
9  & I am someone who Stays optimistic after experiencing a setback & 2.43 & 2.33 & 2.71 & 2.31 & 1.24 & 0.64 & 1.34 & 0.65 \\
10 & I am someone who Is curious about many different things        & 4.43 & 4.08 & 3.58 & 3.53 & 0.81 & 0.54 & 1.21 & 0.64 \\
11 & I am someone who Rarely feels excited or eager                 & 3.54 & 3.62 & 2.91 & 2.63 & 1.26 & 0.72 & 1.27 & 0.92 \\
12 & I am someone who Tends to find fault with others               & 3.34 & 3.67 & 3.15 & 3.63 & 1.32 & 0.54 & 1.25 & 0.71 \\
13 & I am someone who Is dependable, steady                         & 4.33 & 3.60 & 3.33 & 3.79 & 0.86 & 0.70 & 1.30 & 0.72 \\
14 & I am someone who Is moody, has up and down mood swings         & 2.56 & 3.13 & 2.73 & 2.98 & 1.36 & 0.47 & 1.18 & 0.84 \\
15 & I am someone who Is inventive, finds clever ways to do things  & 3.90 & 3.28 & 3.21 & 2.41 & 1.03 & 0.51 & 1.09 & 0.63 \\
16 & I am someone who Tends to be quiet                             & 2.35 & 3.03 & 3.04 & 2.29 & 1.34 & 0.58 & 0.99 & 0.96 \\
17 & I am someone who Feels little sympathy for others              & 3.71 & 3.97 & 3.55 & 3.76 & 1.45 & 0.49 & 1.42 & 0.61 \\
18 & I am someone who Is systematic, likes to keep things in order  & 3.96 & 3.05 & 2.92 & 2.65 & 1.06 & 0.74 & 1.06 & 0.75 \\
19 & I am someone who Can be tense                                 & 3.29 & 3.30 & 3.18 & 3.53 & 1.27 & 0.52 & 1.01 & 0.58 \\
20 & I am someone who Is fascinated by art, music, or literature    & 4.06 & 3.39 & 3.21 & 2.74 & 1.16 & 0.88 & 1.13 & 0.97 \\
21 & I am someone who Is dominant, acts as a leader                 & 2.90 & 2.83 & 2.66 & 2.14 & 1.35 & 0.57 & 1.09 & 0.43 \\
22 & I am someone who Starts arguments with others                 & 4.23 & 3.96 & 3.73 & 3.94 & 1.05 & 0.38 & 1.39 & 0.37 \\
23 & I am someone who Has difficulty getting started on tasks       & 3.31 & 2.95 & 3.00 & 2.58 & 1.39 & 0.48 & 1.24 & 0.66 \\
24 & I am someone who Feels secure, comfortable with self          & 2.26 & 2.87 & 2.73 & 3.43 & 1.27 & 0.72 & 1.35 & 0.71 \\
25 & I am someone who Avoids intellectual, philosophical discussions& 3.94 & 3.18 & 2.92 & 2.55 & 1.21 & 0.82 & 1.22 & 0.85 \\
26 & I am someone who Is less active than other people             & 3.35 & 3.26 & 2.98 & 2.32 & 1.31 & 0.78 & 1.01 & 0.76 \\
27 & I am someone who Has a forgiving nature                      & 3.79 & 3.52 & 3.48 & 3.71 & 1.20 & 0.56 & 1.23 & 0.58 \\
28 & I am someone who Can be somewhat careless                     & 3.56 & 2.91 & 2.99 & 3.04 & 1.25 & 0.41 & 0.93 & 0.59 \\
29 & I am someone who Is emotionally stable, not easily upset      & 2.44 & 3.18 & 3.08 & 3.37 & 1.27 & 0.58 & 1.15 & 0.76 \\
30 & I am someone who Has little creativity                       & 3.95 & 3.59 & 3.14 & 3.16 & 1.20 & 0.59 & 1.16 & 0.93 \\
31 & I am someone who Is sometimes shy, introverted                & 2.35 & 2.75 & 2.96 & 2.36 & 1.36 & 0.56 & 0.83 & 0.87 \\
32 & I am someone who Is helpful and unselfish with others         & 4.18 & 3.73 & 3.51 & 3.94 & 0.88 & 0.66 & 1.43 & 0.62 \\
33 & I am someone who Keeps things neat and tidy                   & 3.71 & 2.90 & 2.86 & 2.87 & 1.24 & 0.62 & 1.07 & 0.64 \\
34 & I am someone who Worries a lot                                & 3.30 & 3.12 & 2.89 & 3.70 & 1.47 & 0.60 & 1.17 & 0.76 \\
35 & I am someone who Values art and beauty                       & 4.16 & 3.71 & 3.37 & 2.97 & 1.07 & 0.78 & 1.09 & 1.03 \\
36 & I am someone who Finds it hard to influence people           & 3.09 & 2.97 & 2.86 & 2.29 & 1.20 & 0.48 & 1.12 & 0.59 \\
37 & I am someone who Is sometimes rude to others                 & 3.79 & 3.57 & 3.52 & 3.74 & 1.25 & 0.54 & 1.23 & 0.58 \\
38 & I am someone who Is efficient, gets things done              & 4.23 & 3.56 & 3.49 & 3.48 & 0.88 & 0.66 & 1.16 & 0.68 \\
39 & I am someone who Often feels sad                             & 2.71 & 3.00 & 2.69 & 2.91 & 1.43 & 0.51 & 1.20 & 1.00 \\
40 & I am someone who Is complex, a deep thinker                  & 4.02 & 3.55 & 3.20 & 2.92 & 1.08 & 0.63 & 1.00 & 0.93 \\
41 & I am someone who Is full of energy                           & 3.14 & 3.36 & 2.98 & 2.45 & 1.30 & 0.74 & 1.20 & 0.69 \\
42 & I am someone who Is suspicious of others’ intentions         & 2.75 & 3.16 & 3.03 & 3.33 & 1.30 & 0.63 & 1.23 & 0.85 \\
43 & I am someone who Is reliable, can always be counted on        & 4.32 & 3.63 & 3.48 & 3.86 & 0.88 & 0.74 & 1.37 & 0.78 \\
44 & I am someone who Keeps their emotions under control          & 2.22 & 3.06 & 2.68 & 2.62 & 1.14 & 0.56 & 1.16 & 0.78 \\
45 & I am someone who Has difficulty imagining things             & 4.17 & 3.41 & 3.19 & 2.78 & 1.06 & 0.54 & 1.02 & 0.61 \\
46 & I am someone who Is talkative                                & 2.87 & 2.96 & 2.83 & 2.62 & 1.38 & 0.46 & 1.04 & 0.82 \\
47 & I am someone who Can be cold and uncaring                    & 3.84 & 3.73 & 3.20 & 3.58 & 1.23 & 0.59 & 1.33 & 0.79 \\
48 & I am someone who Leaves a mess, doesn’t clean up             & 4.15 & 3.55 & 3.18 & 3.90 & 1.12 & 0.60 & 1.33 & 0.34 \\
49 & I am someone who Rarely feels anxious or afraid             & 3.44 & 3.78 & 2.85 & 3.72 & 1.37 & 0.44 & 1.24 & 0.73 \\
50 & I am someone who Thinks poetry and plays are boring          & 3.66 & 3.65 & 3.30 & 2.97 & 1.37 & 0.72 & 1.37 & 0.75 \\
51 & I am someone who Prefers to have others take charge          & 2.96 & 3.69 & 2.97 & 3.55 & 1.28 & 0.61 & 1.18 & 0.83 \\
52 & I am someone who Is polite, courteous to others              & 4.50 & 4.48 & 3.87 & 4.34 & 0.70 & 0.72 & 1.31 & 0.64 \\
53 & I am someone who Is persistent, works until the task is finished & 4.28 & 3.78 & 3.60 & 3.76 & 0.91 & 0.62 & 1.16 & 0.67 \\
54 & I am someone who Tends to feel depressed, blue               & 2.65 & 2.93 & 2.66 & 3.03 & 1.46 & 0.54 & 1.27 & 0.98 \\
55 & I am someone who Has little interest in abstract Ideas       & 3.83 & 2.84 & 2.74 & 2.12 & 1.22 & 0.74 & 1.01 & 0.46 \\
56 & I am someone who Shows a lot of Enthusiasm                   & 3.35 & 3.69 & 3.04 & 2.30 & 1.22 & 0.76 & 1.37 & 0.61 \\
57 & I am someone who Assumes the best about people               & 3.37 & 3.60 & 3.42 & 3.37 & 1.28 & 0.68 & 1.29 & 0.77 \\
58 & I am someone who Sometimes behaves irresponsibly             & 3.64 & 2.86 & 2.88 & 2.84 & 1.30 & 0.44 & 0.91 & 0.65 \\
59 & I am someone who Is temperamental, gets emotional easily     & 2.53 & 2.89 & 2.69 & 2.44 & 1.33 & 0.56 & 1.15 & 0.81 \\
60 & I am someone who Is original, comes up with new Ideas        & 3.84 & 3.24 & 3.23 & 2.46 & 1.07 & 0.49 & 1.04 & 0.63 \\
\midrule
\multicolumn{10}{c}{\textbf{Facet Level}} \\
\midrule
1  & Sociability          & 2.65 & 2.91 & 2.83 & 2.40 & 1.17 & 0.48 & 0.87 & 0.72 \\
2  & Assertiveness        & 2.98 & 3.14 & 2.82 & 2.57 & 1.04 & 0.43 & 0.95 & 0.43 \\
3  & Energy Level         & 3.35 & 3.48 & 2.98 & 2.43 & 0.95 & 0.64 & 1.11 & 0.61 \\
4  & Compassion           & 4.02 & 3.76 & 3.36 & 3.75 & 0.81 & 0.51 & 1.28 & 0.59 \\
5  & Respectfulness       & 4.29 & 4.19 & 3.78 & 4.08 & 0.70 & 0.45 & 1.22 & 0.47 \\
6  & Trust                & 3.31 & 3.49 & 3.27 & 3.51 & 1.00 & 0.49 & 1.16 & 0.62 \\
7  & Organization         & 3.86 & 3.13 & 3.05 & 3.20 & 1.01 & 0.47 & 0.93 & 0.48 \\
8  & Productiveness       & 3.86 & 3.46 & 3.34 & 3.35 & 0.91 & 0.50 & 1.10 & 0.52 \\
9  & Responsibility       & 3.96 & 3.25 & 3.17 & 3.38 & 0.84 & 0.50 & 1.02 & 0.57 \\
10 & Anxiety              & 3.15 & 3.37 & 3.08 & 3.66 & 1.11 & 0.41 & 0.77 & 0.49 \\
11 & Depression           & 2.51 & 2.78 & 2.70 & 2.92 & 1.14 & 0.49 & 1.22 & 0.69 \\
12 & Emotional Volatility & 2.44 & 3.06 & 2.80 & 2.85 & 1.10 & 0.44 & 1.02 & 0.66 \\
13 & Intellectual Curiosity& 4.05 & 3.41 & 3.11 & 2.78 & 0.83 & 0.52 & 0.92 & 0.57 \\
14 & Aesthetic Sensitivity & 3.89 & 3.54 & 3.20 & 2.84 & 1.01 & 0.63 & 1.03 & 0.79 \\
15 & Creative Imagination  & 3.96 & 3.38 & 3.19 & 2.71 & 0.88 & 0.42 & 0.99 & 0.54 \\
\midrule
\multicolumn{10}{c}{\textbf{Domain Level}} \\
\midrule
1 & Extraversion      & 2.99 & 3.18 & 2.88 & 2.47 & 0.85 & 0.46 & 0.91 & 0.53 \\
2 & Agreeableness     & 3.87 & 3.81 & 3.47 & 3.78 & 0.69 & 0.45 & 1.17 & 0.51 \\
3 & Conscientiousness & 3.89 & 3.28 & 3.18 & 3.31 & 0.80 & 0.44 & 0.95 & 0.47 \\
4 & Neuroticism       & 2.70 & 3.07 & 2.86 & 3.14 & 1.02 & 0.39 & 0.92 & 0.54 \\
5 & Openness          & 3.97 & 3.44 & 3.17 & 2.78 & 0.76 & 0.45 & 0.94 & 0.54 \\

\end{longtable}

\begin{longtable}{c p{4.15cm} *{8}{S}}
\caption{Llama3-8B descriptive statistics results.}
\label{tab:Llama3-8B} \\

\toprule
 & & \multicolumn{4}{c}{\textbf{$\mu$}} & \multicolumn{4}{c}{\textbf{$\sigma$}} \\
\cmidrule(lr){3-6}\cmidrule(lr){7-10}
{\textbf{No.}} & {\textbf{Content}} & {Human} & {Persona} & {Shape} & {PSI} & {Human} & {Persona} & {Shape} & {PSI} \\
\midrule
\endfirsthead

\multicolumn{10}{@{}l}{\tablename\ \thetable{} -- continued from previous page} \\
\toprule
 & & \multicolumn{4}{c}{\textbf{$\mu$}} & \multicolumn{4}{c}{\textbf{$\sigma$}} \\
\cmidrule(lr){3-6}\cmidrule(lr){7-10}
{\textbf{No.}} & {\textbf{Content}} & {Human} & {Persona} & {Shape} & {PSI} & {Human} & {Persona} & {Shape} & {PSI} \\
\midrule
\endhead

\midrule
\multicolumn{10}{r@{}}{{Continued on next page}} \\
\endfoot

\bottomrule
\endlastfoot

\multicolumn{10}{c}{\textbf{Item Level}} \\
\midrule
1  & I am someone who is outgoing, sociable                          & 3.02 & 2.31 & 2.62 & 2.80 & 1.40 & 0.70 & 1.32 & 0.92 \\
2  & I am someone who is compassionate, has a soft heart             & 4.34 & 2.86 & 2.56 & 4.08 & 0.90 & 0.89 & 1.16 & 0.87 \\
3  & I am someone who tends to be disorganized                       & 3.62 & 3.46 & 2.87 & 2.84 & 1.36 & 0.66 & 1.12 & 0.61 \\
4  & I am someone who is relaxed, handles stress well                & 2.56 & 3.29 & 3.34 & 2.32 & 1.30 & 0.88 & 1.24 & 0.58 \\
5  & I am someone who has few artistic interests                    & 3.68 & 2.24 & 2.11 & 2.17 & 1.33 & 0.62 & 0.94 & 0.84 \\
6  & I am someone who has an assertive personality                  & 2.96 & 3.25 & 3.12 & 3.11 & 1.38 & 0.82 & 1.16 & 0.74 \\
7  & I am someone who is respectful, treats others with respect     & 4.62 & 3.58 & 3.17 & 4.25 & 0.61 & 0.79 & 1.23 & 0.61 \\
8  & I am someone who tends to be lazy                              & 3.62 & 3.60 & 2.88 & 2.69 & 1.31 & 0.68 & 1.26 & 0.62 \\
9  & I am someone who stays optimistic after experiencing a setback & 2.43 & 2.63 & 2.69 & 2.06 & 1.24 & 0.85 & 1.17 & 0.77 \\
10 & I am someone who is curious about many different things        & 4.43 & 3.63 & 3.04 & 3.88 & 0.81 & 0.73 & 1.17 & 0.83 \\
11 & I am someone who rarely feels excited or eager                 & 3.54 & 2.19 & 2.10 & 2.28 & 1.26 & 0.58 & 0.75 & 0.58 \\
12 & I am someone who tends to find fault with others               & 3.34 & 3.64 & 2.97 & 3.23 & 1.32 & 0.69 & 1.02 & 0.67 \\
13 & I am someone who is dependable, steady                         & 4.33 & 2.90 & 2.76 & 4.13 & 0.86 & 0.88 & 1.20 & 0.72 \\
14 & I am someone who is moody, has up and down mood swings         & 2.56 & 2.99 & 2.95 & 3.60 & 1.36 & 0.91 & 1.11 & 0.61 \\
15 & I am someone who is inventive, finds clever ways to do things  & 3.90 & 3.03 & 2.68 & 3.22 & 1.03 & 0.69 & 1.01 & 0.58 \\
16 & I am someone who tends to be quiet                             & 2.35 & 3.71 & 3.17 & 2.20 & 1.34 & 0.63 & 1.10 & 0.88 \\
17 & I am someone who feels little sympathy for others              & 3.71 & 3.15 & 2.80 & 2.93 & 1.45 & 0.98 & 1.11 & 0.32 \\
18 & I am someone who is systematic, likes to keep things in order  & 3.96 & 2.35 & 2.59 & 3.34 & 1.06 & 0.59 & 1.05 & 0.77 \\
19 & I am someone who can be tense                                 & 3.29 & 3.45 & 3.03 & 3.99 & 1.27 & 0.76 & 1.12 & 0.64 \\
20 & I am someone who is fascinated by art, music, or literature    & 4.06 & 2.79 & 2.77 & 3.40 & 1.16 & 0.90 & 1.12 & 0.94 \\
21 & I am someone who is dominant, acts as a leader                 & 2.90 & 2.74 & 2.65 & 3.05 & 1.35 & 0.75 & 1.02 & 0.61 \\
22 & I am someone who starts arguments with others                 & 4.23 & 3.94 & 3.50 & 3.79 & 1.05 & 0.37 & 1.12 & 0.47 \\
23 & I am someone who has difficulty getting started on tasks       & 3.31 & 2.90 & 2.36 & 2.26 & 1.39 & 0.81 & 0.94 & 0.45 \\
24 & I am someone who feels secure, comfortable with self          & 2.26 & 2.83 & 3.32 & 2.26 & 1.27 & 0.92 & 1.20 & 0.71 \\
25 & I am someone who avoids intellectual, philosophical discussions& 3.94 & 2.11 & 1.98 & 2.30 & 1.21 & 0.44 & 0.63 & 0.60 \\
26 & I am someone who is less active than other people             & 3.35 & 2.98 & 2.22 & 2.15 & 1.31 & 0.94 & 0.96 & 0.58 \\
27 & I am someone who has a forgiving nature                      & 3.79 & 2.91 & 2.65 & 3.74 & 1.20 & 0.90 & 1.08 & 0.73 \\
28 & I am someone who can be somewhat careless                     & 3.56 & 2.98 & 3.23 & 2.27 & 1.25 & 0.92 & 1.06 & 0.48 \\
29 & I am someone who is emotionally stable, not easily upset      & 2.44 & 2.94 & 2.55 & 2.22 & 1.27 & 0.93 & 1.16 & 0.63 \\
30 & I am someone who has little creativity                       & 3.95 & 2.25 & 2.23 & 2.50 & 1.20 & 0.57 & 0.99 & 0.50 \\
31 & I am someone who is sometimes shy, introverted                & 2.35 & 3.63 & 3.13 & 1.97 & 1.36 & 0.64 & 1.07 & 0.85 \\
32 & I am someone who is helpful and unselfish with others         & 4.18 & 3.35 & 3.23 & 4.20 & 0.88 & 0.84 & 1.19 & 0.58 \\
33 & I am someone who keeps things neat and tidy                   & 3.71 & 2.35 & 2.81 & 3.34 & 1.24 & 0.60 & 1.09 & 0.75 \\
34 & I am someone who worries a lot                                & 3.30 & 2.47 & 3.06 & 4.01 & 1.47 & 0.81 & 1.23 & 0.79 \\
35 & I am someone who values art and beauty                       & 4.16 & 3.24 & 3.16 & 3.48 & 1.07 & 0.86 & 1.10 & 0.91 \\
36 & I am someone who finds it hard to influence people           & 3.09 & 2.20 & 2.01 & 2.06 & 1.20 & 0.58 & 0.60 & 0.36 \\
37 & I am someone who is sometimes rude to others                 & 3.79 & 3.49 & 2.87 & 2.74 & 1.25 & 0.82 & 1.17 & 0.64 \\
38 & I am someone who is efficient, gets things done              & 4.23 & 2.48 & 2.76 & 3.89 & 0.88 & 0.74 & 1.19 & 0.63 \\
39 & I am someone who often feels sad                             & 2.71 & 2.68 & 2.71 & 3.68 & 1.43 & 0.86 & 1.20 & 0.77 \\
40 & I am someone who is complex, a deep thinker                  & 4.02 & 3.15 & 2.82 & 3.75 & 1.08 & 0.87 & 1.02 & 0.67 \\
41 & I am someone who is full of energy                           & 3.14 & 2.78 & 2.91 & 3.15 & 1.30 & 0.89 & 1.28 & 0.86 \\
42 & I am someone who is suspicious of others’ intentions         & 2.75 & 3.43 & 2.96 & 2.74 & 1.30 & 0.68 & 1.04 & 0.65 \\
43 & I am someone who is reliable, can always be counted on        & 4.32 & 2.82 & 2.84 & 4.17 & 0.88 & 0.83 & 1.21 & 0.66 \\
44 & I am someone who keeps their emotions under control          & 2.22 & 2.92 & 2.46 & 2.08 & 1.14 & 0.92 & 1.04 & 0.42 \\
45 & I am someone who has difficulty imagining things             & 4.17 & 2.14 & 2.18 & 2.42 & 1.06 & 0.46 & 0.81 & 0.52 \\
46 & I am someone who is talkative                                & 2.87 & 2.73 & 2.58 & 3.43 & 1.38 & 0.93 & 1.16 & 1.00 \\
47 & I am someone who can be cold and uncaring                    & 3.84 & 3.64 & 2.93 & 3.01 & 1.23 & 0.70 & 1.19 & 0.61 \\
48 & I am someone who leaves a mess, doesn’t clean up             & 4.15 & 3.91 & 3.46 & 2.59 & 1.12 & 0.44 & 1.14 & 0.53 \\
49 & I am someone who rarely feels anxious or afraid             & 3.44 & 2.69 & 2.16 & 2.67 & 1.37 & 0.94 & 0.91 & 0.61 \\
50 & I am someone who thinks poetry and plays are boring          & 3.66 & 3.31 & 2.63 & 2.44 & 1.37 & 0.89 & 1.17 & 0.61 \\
51 & I am someone who prefers to have others take charge          & 2.96 & 3.17 & 2.57 & 2.58 & 1.28 & 0.94 & 1.03 & 0.51 \\
52 & I am someone who is polite, courteous to others              & 4.50 & 2.98 & 2.88 & 4.11 & 0.70 & 0.97 & 1.17 & 0.59 \\
53 & I am someone who is persistent, works until the task is finished & 4.28 & 3.58 & 3.10 & 4.20 & 0.91 & 0.76 & 1.27 & 0.67 \\
54 & I am someone who tends to feel depressed, blue               & 2.65 & 2.41 & 2.39 & 3.52 & 1.46 & 0.74 & 1.27 & 0.89 \\
55 & I am someone who has little interest in abstract ideas      & 3.83 & 2.09 & 2.05 & 2.15 & 1.22 & 0.38 & 0.56 & 0.68 \\
56 & I am someone who shows a lot of enthusiasm                   & 3.35 & 3.58 & 2.96 & 2.99 & 1.22 & 0.75 & 1.31 & 0.83 \\
57 & I am someone who assumes the best about people               & 3.37 & 3.01 & 2.89 & 3.48 & 1.28 & 0.95 & 1.09 & 0.85 \\
58 & I am someone who sometimes behaves irresponsibly             & 3.64 & 2.63 & 2.81 & 2.33 & 1.30 & 0.78 & 1.13 & 0.49 \\
59 & I am someone who is temperamental, gets emotional easily     & 2.53 & 2.65 & 2.86 & 3.34 & 1.33 & 0.88 & 1.19 & 0.77 \\
60 & I am someone who is original, comes up with new ideas        & 3.84 & 2.93 & 2.91 & 3.62 & 1.07 & 0.83 & 1.08 & 0.60 \\
\midrule
\multicolumn{10}{c}{\textbf{Facet Level}} \\
\midrule
1  & Sociability          & 2.65 & 3.10 & 2.88 & 2.60 & 1.17 & 0.45 & 0.77 & 0.76 \\
2  & Assertiveness        & 2.98 & 2.84 & 2.59 & 2.70 & 1.04 & 0.39 & 0.61 & 0.42 \\
3  & Energy Level         & 3.35 & 2.88 & 2.55 & 2.64 & 0.95 & 0.42 & 0.63 & 0.59 \\
4  & Compassion           & 4.02 & 3.25 & 2.88 & 3.55 & 0.81 & 0.56 & 0.83 & 0.47 \\
5  & Respectfulness       & 4.29 & 3.50 & 3.10 & 3.72 & 0.70 & 0.47 & 0.88 & 0.43 \\
6  & Trust                & 3.31 & 3.25 & 2.87 & 3.30 & 1.00 & 0.48 & 0.76 & 0.55 \\
7  & Organization         & 3.86 & 3.02 & 2.93 & 3.03 & 1.01 & 0.29 & 0.75 & 0.47 \\
8  & Productiveness       & 3.86 & 3.14 & 2.78 & 3.26 & 0.91 & 0.38 & 0.79 & 0.47 \\
9  & Responsibility       & 3.96 & 2.83 & 2.91 & 3.22 & 0.84 & 0.49 & 0.82 & 0.47 \\
10 & Anxiety              & 3.15 & 2.98 & 2.90 & 3.25 & 1.11 & 0.40 & 0.47 & 0.53 \\
11 & Depression           & 2.51 & 2.64 & 2.78 & 2.88 & 1.14 & 0.51 & 0.80 & 0.64 \\
12 & Emotional Volatility & 2.44 & 2.88 & 2.70 & 2.81 & 1.10 & 0.49 & 0.55 & 0.48 \\
13 & Intellectual Curiosity& 4.05 & 2.74 & 2.47 & 3.02 & 0.83 & 0.34 & 0.57 & 0.57 \\
14 & Aesthetic Sensitivity & 3.89 & 2.90 & 2.67 & 2.87 & 1.01 & 0.46 & 0.76 & 0.69 \\
15 & Creative Imagination  & 3.96 & 2.59 & 2.50 & 2.94 & 0.88 & 0.35 & 0.62 & 0.43 \\
\midrule
\multicolumn{10}{c}{\textbf{Domain Level}} \\
\midrule
1 & Extraversion      & 2.99 & 2.94 & 2.67 & 2.65 & 0.85 & 0.33 & 0.57 & 0.52 \\
2 & Agreeableness     & 3.87 & 3.33 & 2.95 & 3.53 & 0.69 & 0.44 & 0.75 & 0.43 \\
3 & Conscientiousness & 3.89 & 3.00 & 2.87 & 3.17 & 0.80 & 0.30 & 0.70 & 0.41 \\
4 & Neuroticism       & 2.70 & 2.83 & 2.79 & 2.98 & 1.02 & 0.37 & 0.49 & 0.49 \\
5 & Openness          & 3.97 & 2.74 & 2.55 & 2.94 & 0.76 & 0.29 & 0.59 & 0.48 \\

\end{longtable}

\begin{longtable}{c p{4.15cm} *{8}{S}}
\caption{Llama3-70B descriptive statistics results.}
\label{tab:Llama3-70B} \\

\toprule
 & & \multicolumn{4}{c}{\textbf{$\mu$}} & \multicolumn{4}{c}{\textbf{$\sigma$}} \\
\cmidrule(lr){3-6}\cmidrule(lr){7-10}
{\textbf{No.}} & {\textbf{Content}} & {Human} & {Persona} & {Shape} & {PSI} & {Human} & {Persona} & {Shape} & {PSI} \\
\midrule
\endfirsthead

\multicolumn{10}{@{}l}{\tablename\ \thetable{} -- continued from previous page} \\
\toprule
 & & \multicolumn{4}{c}{\textbf{$\mu$}} & \multicolumn{4}{c}{\textbf{$\sigma$}} \\
\cmidrule(lr){3-6}\cmidrule(lr){7-10}
{\textbf{No.}} & {\textbf{Content}} & {Human} & {Persona} & {Shape} & {PSI} & {Human} & {Persona} & {Shape} & {PSI} \\
\midrule
\endhead

\midrule
\multicolumn{10}{r@{}}{{Continued on next page}} \\
\endfoot

\bottomrule
\endlastfoot

\multicolumn{10}{c}{\textbf{Item Level}} \\
\midrule
1  & I am someone who is outgoing, sociable                          & 3.02 & 2.77 & 2.43 & 2.15 & 1.40 & 1.31 & 1.37 & 0.78 \\
2  & I am someone who is compassionate, has a soft heart             & 4.34 & 3.87 & 3.15 & 3.93 & 0.90 & 1.12 & 1.44 & 1.01 \\
3  & I am someone who tends to be disorganized                       & 3.62 & 3.77 & 3.29 & 3.27 & 1.36 & 0.95 & 1.47 & 0.96 \\
4  & I am someone who is relaxed, handles stress well                & 2.56 & 2.94 & 3.34 & 2.97 & 1.30 & 1.16 & 1.39 & 0.95 \\
5  & I am someone who has few artistic interests                    & 3.68 & 3.80 & 2.91 & 2.71 & 1.33 & 1.31 & 1.32 & 1.12 \\
6  & I am someone who has an assertive personality                  & 2.96 & 3.07 & 2.79 & 2.24 & 1.38 & 1.18 & 1.37 & 0.70 \\
7  & I am someone who is respectful, treats others with respect     & 4.62 & 4.43 & 3.60 & 4.23 & 0.61 & 0.92 & 1.43 & 0.78 \\
8  & I am someone who tends to be lazy                              & 3.62 & 4.23 & 3.44 & 3.77 & 1.31 & 1.04 & 1.49 & 0.96 \\
9  & I am someone who stays optimistic after experiencing a setback & 2.43 & 2.29 & 2.94 & 2.28 & 1.24 & 0.98 & 1.54 & 0.92 \\
10 & I am someone who is curious about many different things        & 4.43 & 3.71 & 3.14 & 3.06 & 0.81 & 1.12 & 1.55 & 0.92 \\
11 & I am someone who rarely feels excited or eager                 & 3.54 & 4.18 & 2.98 & 2.40 & 1.26 & 0.75 & 1.42 & 0.81 \\
12 & I am someone who tends to find fault with others               & 3.34 & 4.31 & 3.44 & 3.72 & 1.32 & 0.69 & 1.48 & 0.68 \\
13 & I am someone who is dependable, steady                         & 4.33 & 3.70 & 3.09 & 3.98 & 0.86 & 1.12 & 1.59 & 0.75 \\
14 & I am someone who is moody, has up and down mood swings         & 2.56 & 2.52 & 2.62 & 2.31 & 1.36 & 1.09 & 1.42 & 0.93 \\
15 & I am someone who is inventive, finds clever ways to do things  & 3.90 & 3.24 & 2.90 & 2.35 & 1.03 & 1.14 & 1.44 & 0.71 \\
16 & I am someone who tends to be quiet                             & 2.35 & 3.44 & 3.26 & 2.53 & 1.34 & 1.12 & 1.27 & 0.96 \\
17 & I am someone who feels little sympathy for others              & 3.71 & 4.47 & 3.52 & 2.93 & 1.45 & 0.79 & 1.54 & 1.01 \\
18 & I am someone who is systematic, likes to keep things in order  & 3.96 & 2.46 & 2.57 & 2.45 & 1.06 & 1.09 & 1.37 & 0.83 \\
19 & I am someone who can be tense                                 & 3.29 & 3.18 & 3.06 & 2.73 & 1.27 & 1.08 & 1.21 & 0.94 \\
20 & I am someone who is fascinated by art, music, or literature    & 4.06 & 2.99 & 2.89 & 2.81 & 1.16 & 1.44 & 1.36 & 1.10 \\
21 & I am someone who is dominant, acts as a leader                 & 2.90 & 2.46 & 2.51 & 2.06 & 1.35 & 1.15 & 1.49 & 0.59 \\
22 & I am someone who starts arguments with others                 & 4.23 & 4.56 & 3.68 & 4.35 & 1.05 & 0.77 & 1.54 & 0.59 \\
23 & I am someone who has difficulty getting started on tasks       & 3.31 & 3.85 & 3.21 & 2.81 & 1.39 & 0.86 & 1.50 & 0.70 \\
24 & I am someone who feels secure, comfortable with self          & 2.26 & 2.24 & 2.67 & 2.70 & 1.27 & 1.08 & 1.50 & 0.92 \\
25 & I am someone who avoids intellectual, philosophical discussions& 3.94 & 3.36 & 3.12 & 2.62 & 1.21 & 1.36 & 1.56 & 1.01 \\
26 & I am someone who is less active than other people             & 3.35 & 3.86 & 3.14 & 2.68 & 1.31 & 1.04 & 1.31 & 0.83 \\
27 & I am someone who has a forgiving nature                      & 3.79 & 3.48 & 3.26 & 3.61 & 1.20 & 0.90 & 1.38 & 0.77 \\
28 & I am someone who can be somewhat careless                     & 3.56 & 3.19 & 2.99 & 3.18 & 1.25 & 1.11 & 1.38 & 1.01 \\
29 & I am someone who is emotionally stable, not easily upset      & 2.44 & 2.66 & 2.89 & 2.68 & 1.27 & 1.01 & 1.48 & 1.09 \\
30 & I am someone who has little creativity                       & 3.95 & 4.36 & 3.42 & 2.85 & 1.20 & 0.69 & 1.32 & 1.02 \\
31 & I am someone who is sometimes shy, introverted                & 2.35 & 3.43 & 3.33 & 2.72 & 1.36 & 1.11 & 1.31 & 0.98 \\
32 & I am someone who is helpful and unselfish with others         & 4.18 & 3.88 & 3.22 & 4.12 & 0.88 & 1.05 & 1.51 & 0.73 \\
33 & I am someone who keeps things neat and tidy                   & 3.71 & 2.47 & 2.61 & 2.71 & 1.24 & 1.00 & 1.35 & 0.70 \\
34 & I am someone who worries a lot                                & 3.30 & 2.59 & 2.53 & 3.08 & 1.47 & 1.08 & 1.34 & 1.05 \\
35 & I am someone who values art and beauty                       & 4.16 & 3.23 & 3.05 & 2.88 & 1.07 & 1.30 & 1.35 & 1.06 \\
36 & I am someone who finds it hard to influence people           & 3.09 & 3.44 & 2.97 & 2.27 & 1.20 & 1.04 & 1.37 & 0.70 \\
37 & I am someone who is sometimes rude to others                 & 3.79 & 4.43 & 3.61 & 4.13 & 1.25 & 0.91 & 1.50 & 0.99 \\
38 & I am someone who is efficient, gets things done              & 4.23 & 3.45 & 3.10 & 3.09 & 0.88 & 1.12 & 1.47 & 0.92 \\
39 & I am someone who often feels sad                             & 2.71 & 2.27 & 2.44 & 2.68 & 1.43 & 0.98 & 1.27 & 0.90 \\
40 & I am someone who is complex, a deep thinker                  & 4.02 & 3.59 & 3.15 & 2.75 & 1.08 & 1.06 & 1.26 & 0.99 \\
41 & I am someone who is full of energy                           & 3.14 & 3.39 & 2.86 & 2.39 & 1.30 & 1.24 & 1.35 & 0.74 \\
42 & I am someone who is suspicious of others’ intentions         & 2.75 & 3.91 & 3.38 & 3.61 & 1.30 & 0.83 & 1.50 & 0.68 \\
43 & I am someone who is reliable, can always be counted on        & 4.32 & 3.76 & 3.15 & 3.76 & 0.88 & 1.09 & 1.60 & 0.77 \\
44 & I am someone who keeps their emotions under control          & 2.22 & 2.92 & 2.92 & 2.36 & 1.14 & 1.08 & 1.45 & 1.05 \\
45 & I am someone who has difficulty imagining things             & 4.17 & 4.08 & 3.43 & 2.13 & 1.06 & 0.53 & 1.26 & 0.52 \\
46 & I am someone who is talkative                                & 2.87 & 2.53 & 2.49 & 2.36 & 1.38 & 1.11 & 1.28 & 0.92 \\
47 & I am someone who can be cold and uncaring                    & 3.84 & 4.44 & 3.46 & 4.21 & 1.23 & 0.87 & 1.49 & 0.96 \\
48 & I am someone who leaves a mess, doesn’t clean up             & 4.15 & 4.25 & 3.23 & 4.06 & 1.12 & 0.95 & 1.65 & 0.83 \\
49 & I am someone who rarely feels anxious or afraid             & 3.44 & 3.12 & 2.73 & 3.66 & 1.37 & 1.12 & 1.38 & 0.88 \\
50 & I am someone who thinks poetry and plays are boring          & 3.66 & 3.85 & 3.14 & 2.31 & 1.37 & 1.31 & 1.54 & 0.70 \\
51 & I am someone who prefers to have others take charge          & 2.96 & 4.33 & 3.21 & 3.86 & 1.28 & 0.76 & 1.57 & 0.51 \\
52 & I am someone who is polite, courteous to others              & 4.50 & 4.13 & 3.54 & 4.38 & 0.70 & 0.97 & 1.45 & 0.79 \\
53 & I am someone who is persistent, works until the task is finished & 4.28 & 3.87 & 3.12 & 3.76 & 0.91 & 1.13 & 1.48 & 0.95 \\
54 & I am someone who tends to feel depressed, blue               & 2.65 & 2.01 & 2.39 & 2.76 & 1.46 & 1.07 & 1.38 & 0.98 \\
55 & I am someone who has little interest in abstract ideas      & 3.83 & 3.40 & 2.92 & 2.17 & 1.22 & 1.08 & 1.36 & 0.65 \\
56 & I am someone who shows a lot of enthusiasm                   & 3.35 & 4.08 & 2.79 & 2.58 & 1.22 & 1.09 & 1.51 & 0.93 \\
57 & I am someone who assumes the best about people               & 3.37 & 3.40 & 3.11 & 3.63 & 1.28 & 0.98 & 1.47 & 0.75 \\
58 & I am someone who sometimes behaves irresponsibly             & 3.64 & 3.44 & 3.20 & 3.57 & 1.30 & 1.13 & 1.40 & 0.96 \\
59 & I am someone who is temperamental, gets emotional easily     & 2.53 & 2.33 & 2.66 & 2.04 & 1.33 & 1.08 & 1.52 & 0.89 \\
60 & I am someone who is original, comes up with new ideas        & 3.84 & 3.25 & 2.88 & 2.39 & 1.07 & 1.13 & 1.40 & 0.72 \\
\midrule
\multicolumn{10}{c}{\textbf{Facet Level}} \\
\midrule
1  & Sociability          & 2.65 & 3.04 & 2.88 & 2.44 & 1.17 & 0.98 & 1.00 & 0.73 \\
2  & Assertiveness        & 2.98 & 3.33 & 2.87 & 2.61 & 1.04 & 0.82 & 1.22 & 0.41 \\
3  & Energy Level         & 3.35 & 3.88 & 2.94 & 2.51 & 0.95 & 0.88 & 1.19 & 0.64 \\
4  & Compassion           & 4.02 & 4.16 & 3.34 & 3.80 & 0.81 & 0.81 & 1.37 & 0.76 \\
5  & Respectfulness       & 4.29 & 4.39 & 3.61 & 4.27 & 0.70 & 0.74 & 1.34 & 0.68 \\
6  & Trust                & 3.31 & 3.78 & 3.30 & 3.64 & 1.00 & 0.67 & 1.35 & 0.61 \\
7  & Organization         & 3.86 & 3.24 & 2.93 & 3.13 & 1.01 & 0.79 & 1.19 & 0.68 \\
8  & Productiveness       & 3.86 & 3.85 & 3.22 & 3.36 & 0.91 & 0.87 & 1.29 & 0.69 \\
9  & Responsibility       & 3.96 & 3.52 & 3.11 & 3.62 & 0.84 & 0.95 & 1.26 & 0.74 \\
10 & Anxiety              & 3.15 & 2.96 & 2.91 & 3.11 & 1.11 & 0.89 & 1.06 & 0.75 \\
11 & Depression           & 2.51 & 2.20 & 2.61 & 2.61 & 1.14 & 0.89 & 1.28 & 0.80 \\
12 & Emotional Volatility & 2.44 & 2.61 & 2.77 & 2.35 & 1.10 & 0.92 & 1.26 & 0.88 \\
13 & Intellectual Curiosity& 4.05 & 3.51 & 3.08 & 2.65 & 0.83 & 0.96 & 1.24 & 0.71 \\
14 & Aesthetic Sensitivity & 3.89 & 3.47 & 3.00 & 2.68 & 1.01 & 1.09 & 1.26 & 0.83 \\
15 & Creative Imagination  & 3.96 & 3.73 & 3.16 & 2.43 & 0.88 & 0.74 & 1.17 & 0.59 \\
\midrule
\multicolumn{10}{c}{\textbf{Domain Level}} \\
\midrule
1 & Extraversion      & 2.99 & 3.42 & 2.90 & 2.52 & 0.85 & 0.81 & 1.00 & 0.52 \\
2 & Agreeableness     & 3.87 & 4.11 & 3.41 & 3.90 & 0.69 & 0.70 & 1.29 & 0.62 \\
3 & Conscientiousness & 3.89 & 3.54 & 3.08 & 3.37 & 0.80 & 0.79 & 1.16 & 0.65 \\
4 & Neuroticism       & 2.70 & 2.59 & 2.77 & 2.69 & 1.02 & 0.81 & 1.10 & 0.72 \\
5 & Openness          & 3.97 & 3.57 & 3.08 & 2.59 & 0.76 & 0.78 & 1.15 & 0.59 \\

\end{longtable}

\begin{longtable}{c p{4.15cm} *{8}{S}}
\caption{GPT-4o-mini descriptive statistics results.}
\label{tab:GPT-4o_mini} \\

\toprule
 & & \multicolumn{4}{c}{\textbf{$\mu$}} & \multicolumn{4}{c}{\textbf{$\sigma$}} \\
\cmidrule(lr){3-6}\cmidrule(lr){7-10}
{\textbf{No.}} & {\textbf{Content}} & {Human} & {Persona} & {Shape} & {PSI} & {Human} & {Persona} & {Shape} & {PSI} \\
\midrule
\endfirsthead

\multicolumn{10}{@{}l}{\tablename\ \thetable{} -- continued from previous page} \\
\toprule
 & & \multicolumn{4}{c}{\textbf{$\mu$}} & \multicolumn{4}{c}{\textbf{$\sigma$}} \\
\cmidrule(lr){3-6}\cmidrule(lr){7-10}
{\textbf{No.}} & {\textbf{Content}} & {Human} & {Persona} & {Shape} & {PSI} & {Human} & {Persona} & {Shape} & {PSI} \\
\midrule
\endhead

\midrule
\multicolumn{10}{r@{}}{{Continued on next page}} \\
\endfoot

\bottomrule
\endlastfoot

\multicolumn{10}{c}{\textbf{Item Level}} \\
\midrule
1  & I am someone who is outgoing, sociable                          & 3.02 & 2.90 & 3.15 & 2.64 & 1.40 & 0.98 & 1.31 & 0.96 \\
2  & I am someone who is compassionate, has a soft heart             & 4.34 & 3.94 & 3.58 & 4.11 & 0.90 & 0.86 & 1.23 & 0.96 \\
3  & I am someone who tends to be disorganized                       & 3.62 & 3.42 & 3.05 & 3.12 & 1.36 & 0.83 & 1.34 & 0.69 \\
4  & I am someone who is relaxed, handles stress well                & 2.56 & 2.82 & 2.63 & 2.96 & 1.30 & 0.78 & 1.25 & 0.90 \\
5  & I am someone who has few artistic interests                    & 3.68 & 4.05 & 3.14 & 2.94 & 1.33 & 0.89 & 1.34 & 1.10 \\
6  & I am someone who has an assertive personality                  & 2.96 & 2.98 & 3.12 & 2.55 & 1.38 & 0.76 & 1.22 & 0.74 \\
7  & I am someone who is respectful, treats others with respect     & 4.62 & 4.70 & 4.06 & 4.56 & 0.61 & 0.59 & 1.21 & 0.73 \\
8  & I am someone who tends to be lazy                              & 3.62 & 3.94 & 3.29 & 3.56 & 1.31 & 0.72 & 1.36 & 0.73 \\
9  & I am someone who stays optimistic after experiencing a setback & 2.43 & 2.34 & 2.36 & 2.26 & 1.24 & 0.70 & 1.25 & 0.92 \\
10 & I am someone who is curious about many different things        & 4.43 & 4.43 & 3.86 & 3.70 & 0.81 & 0.74 & 1.26 & 0.96 \\
11 & I am someone who rarely feels excited or eager                 & 3.54 & 4.06 & 3.09 & 3.20 & 1.26 & 0.73 & 1.46 & 0.80 \\
12 & I am someone who tends to find fault with others               & 3.34 & 4.00 & 3.45 & 3.79 & 1.32 & 0.44 & 1.37 & 0.57 \\
13 & I am someone who is dependable, steady                         & 4.33 & 3.78 & 3.62 & 4.27 & 0.86 & 0.84 & 1.36 & 0.78 \\
14 & I am someone who is moody, has up and down mood swings         & 2.56 & 2.80 & 2.63 & 3.02 & 1.36 & 0.66 & 1.38 & 0.78 \\
15 & I am someone who is inventive, finds clever ways to do things  & 3.90 & 3.62 & 3.45 & 3.07 & 1.03 & 0.67 & 1.20 & 0.63 \\
16 & I am someone who tends to be quiet                             & 2.35 & 3.21 & 2.84 & 2.22 & 1.34 & 0.89 & 1.15 & 0.97 \\
17 & I am someone who feels little sympathy for others              & 3.71 & 4.68 & 3.51 & 4.23 & 1.45 & 0.60 & 1.52 & 0.63 \\
18 & I am someone who is systematic, likes to keep things in order  & 3.96 & 3.06 & 3.21 & 3.16 & 1.06 & 0.96 & 1.33 & 0.94 \\
19 & I am someone who can be tense                                 & 3.29 & 3.49 & 3.30 & 3.78 & 1.27 & 0.60 & 1.00 & 0.77 \\
20 & I am someone who is fascinated by art, music, or literature    & 4.06 & 3.49 & 3.53 & 3.38 & 1.16 & 1.04 & 1.23 & 1.18 \\
21 & I am someone who is dominant, acts as a leader                 & 2.90 & 2.58 & 2.93 & 2.39 & 1.35 & 0.80 & 1.25 & 0.66 \\
22 & I am someone who starts arguments with others                 & 4.23 & 4.87 & 4.05 & 4.48 & 1.05 & 0.38 & 1.34 & 0.57 \\
23 & I am someone who has difficulty getting started on tasks       & 3.31 & 3.04 & 2.85 & 2.85 & 1.39 & 0.70 & 1.25 & 0.50 \\
24 & I am someone who feels secure, comfortable with self          & 2.26 & 2.63 & 2.37 & 2.65 & 1.27 & 0.78 & 1.20 & 0.92 \\
25 & I am someone who avoids intellectual, philosophical discussions& 3.94 & 3.61 & 2.95 & 2.69 & 1.21 & 0.87 & 1.45 & 0.94 \\
26 & I am someone who is less active than other people             & 3.35 & 3.36 & 2.79 & 2.55 & 1.31 & 0.95 & 1.20 & 0.83 \\
27 & I am someone who has a forgiving nature                      & 3.79 & 3.58 & 3.51 & 3.56 & 1.20 & 0.63 & 1.11 & 0.82 \\
28 & I am someone who can be somewhat careless                     & 3.56 & 3.32 & 2.87 & 3.07 & 1.25 & 0.72 & 1.11 & 0.65 \\
29 & I am someone who is emotionally stable, not easily upset      & 2.44 & 2.69 & 2.31 & 2.71 & 1.27 & 0.81 & 1.22 & 1.03 \\
30 & I am someone who has little creativity                       & 3.95 & 4.38 & 3.29 & 3.39 & 1.20 & 0.69 & 1.43 & 0.75 \\
31 & I am someone who is sometimes shy, introverted                & 2.35 & 2.85 & 2.86 & 2.11 & 1.36 & 0.81 & 1.11 & 0.98 \\
32 & I am someone who is helpful and unselfish with others         & 4.18 & 3.96 & 3.75 & 4.34 & 0.88 & 0.72 & 1.24 & 0.77 \\
33 & I am someone who keeps things neat and tidy                   & 3.71 & 3.03 & 3.42 & 3.10 & 1.24 & 0.91 & 1.26 & 0.71 \\
34 & I am someone who worries a lot                                & 3.30 & 2.97 & 2.93 & 3.72 & 1.47 & 0.78 & 1.25 & 1.07 \\
35 & I am someone who values art and beauty                       & 4.16 & 3.64 & 3.61 & 3.36 & 1.07 & 0.99 & 1.25 & 0.94 \\
36 & I am someone who finds it hard to influence people           & 3.09 & 3.06 & 2.68 & 2.68 & 1.20 & 0.68 & 1.08 & 0.58 \\
37 & I am someone who is sometimes rude to others                 & 3.79 & 4.26 & 3.68 & 3.91 & 1.25 & 0.66 & 1.39 & 0.66 \\
38 & I am someone who is efficient, gets things done              & 4.23 & 3.67 & 3.68 & 3.82 & 0.88 & 0.84 & 1.17 & 0.83 \\
39 & I am someone who often feels sad                             & 2.71 & 2.56 & 2.67 & 3.39 & 1.43 & 0.71 & 1.14 & 0.82 \\
40 & I am someone who is complex, a deep thinker                  & 4.02 & 3.66 & 3.53 & 3.45 & 1.08 & 0.69 & 0.97 & 0.87 \\
41 & I am someone who is full of energy                           & 3.14 & 3.31 & 3.09 & 2.75 & 1.30 & 0.94 & 1.29 & 0.80 \\
42 & I am someone who is suspicious of others’ intentions         & 2.75 & 3.54 & 3.19 & 3.35 & 1.30 & 0.68 & 1.32 & 0.93 \\
43 & I am someone who is reliable, can always be counted on        & 4.32 & 3.84 & 3.85 & 4.21 & 0.88 & 0.83 & 1.20 & 0.81 \\
44 & I am someone who keeps their emotions under control          & 2.22 & 2.55 & 2.45 & 2.46 & 1.14 & 0.70 & 1.24 & 0.78 \\
45 & I am someone who has difficulty imagining things             & 4.17 & 3.96 & 3.20 & 3.10 & 1.06 & 0.59 & 1.32 & 0.73 \\
46 & I am someone who is talkative                                & 2.87 & 2.89 & 3.03 & 3.07 & 1.38 & 0.78 & 1.21 & 0.95 \\
47 & I am someone who can be cold and uncaring                    & 3.84 & 3.93 & 3.33 & 3.98 & 1.23 & 0.64 & 1.41 & 0.84 \\
48 & I am someone who leaves a mess, doesn’t clean up             & 4.15 & 4.37 & 3.18 & 3.63 & 1.12 & 0.91 & 1.65 & 0.53 \\
49 & I am someone who rarely feels anxious or afraid             & 3.44 & 3.47 & 2.41 & 3.58 & 1.37 & 0.73 & 1.16 & 0.89 \\
50 & I am someone who thinks poetry and plays are boring          & 3.66 & 3.77 & 3.17 & 2.76 & 1.37 & 0.79 & 1.41 & 0.61 \\
51 & I am someone who prefers to have others take charge          & 2.96 & 3.34 & 2.49 & 3.21 & 1.28 & 0.79 & 1.27 & 0.72 \\
52 & I am someone who is polite, courteous to others              & 4.50 & 4.31 & 3.92 & 4.20 & 0.70 & 0.81 & 1.24 & 0.81 \\
53 & I am someone who is persistent, works until the task is finished & 4.28 & 4.15 & 3.67 & 4.24 & 0.91 & 0.83 & 1.18 & 0.84 \\
54 & I am someone who tends to feel depressed, blue               & 2.65 & 2.45 & 2.56 & 3.40 & 1.46 & 0.72 & 1.26 & 0.89 \\
55 & I am someone who has little interest in abstract ideas      & 3.83 & 3.31 & 2.89 & 2.63 & 1.22 & 1.10 & 1.39 & 1.12 \\
56 & I am someone who shows a lot of enthusiasm                   & 3.35 & 3.77 & 3.42 & 2.81 & 1.22 & 0.93 & 1.35 & 0.92 \\
57 & I am someone who assumes the best about people               & 3.37 & 3.54 & 3.72 & 3.48 & 1.28 & 0.75 & 1.23 & 0.95 \\
58 & I am someone who sometimes behaves irresponsibly             & 3.64 & 3.27 & 3.00 & 3.12 & 1.30 & 0.73 & 1.14 & 0.66 \\
59 & I am someone who is temperamental, gets emotional easily     & 2.53 & 2.86 & 2.71 & 2.82 & 1.33 & 0.71 & 1.32 & 0.84 \\
60 & I am someone who is original, comes up with new ideas        & 3.84 & 3.55 & 3.56 & 3.07 & 1.07 & 0.73 & 1.19 & 0.68 \\
\midrule
\multicolumn{10}{c}{\textbf{Facet Level}} \\
\midrule
1  & Sociability          & 2.65 & 2.96 & 2.97 & 2.51 & 1.17 & 0.71 & 0.90 & 0.87 \\
2  & Assertiveness        & 2.98 & 2.99 & 2.81 & 2.71 & 1.04 & 0.63 & 0.93 & 0.54 \\
3  & Energy Level         & 3.35 & 3.62 & 3.10 & 2.83 & 0.95 & 0.77 & 1.09 & 0.72 \\
4  & Compassion           & 4.02 & 4.13 & 3.54 & 4.17 & 0.81 & 0.59 & 1.16 & 0.70 \\
5  & Respectfulness       & 4.29 & 4.54 & 3.93 & 4.29 & 0.70 & 0.50 & 1.10 & 0.56 \\
6  & Trust                & 3.31 & 3.66 & 3.47 & 3.55 & 1.00 & 0.51 & 1.05 & 0.70 \\
7  & Organization         & 3.86 & 3.47 & 3.21 & 3.25 & 1.01 & 0.74 & 0.99 & 0.61 \\
8  & Productiveness       & 3.86 & 3.70 & 3.37 & 3.62 & 0.91 & 0.66 & 1.01 & 0.61 \\
9  & Responsibility       & 3.96 & 3.55 & 3.34 & 3.67 & 0.84 & 0.67 & 0.94 & 0.62 \\
10 & Anxiety              & 3.15 & 3.19 & 2.82 & 3.51 & 1.11 & 0.55 & 0.79 & 0.78 \\
11 & Depression           & 2.51 & 2.49 & 2.49 & 2.92 & 1.14 & 0.61 & 0.97 & 0.78 \\
12 & Emotional Volatility & 2.44 & 2.73 & 2.53 & 2.75 & 1.10 & 0.59 & 0.98 & 0.74 \\
13 & Intellectual Curiosity& 4.05 & 3.75 & 3.31 & 3.12 & 0.83 & 0.67 & 0.97 & 0.84 \\
14 & Aesthetic Sensitivity & 3.89 & 3.74 & 3.36 & 3.11 & 1.01 & 0.79 & 1.13 & 0.83 \\
15 & Creative Imagination  & 3.96 & 3.88 & 3.38 & 3.16 & 0.88 & 0.54 & 1.01 & 0.59 \\
\midrule
\multicolumn{10}{c}{\textbf{Domain Level}} \\
\midrule
1 & Extraversion      & 2.99 & 3.19 & 2.96 & 2.68 & 0.85 & 0.64 & 0.86 & 0.63 \\
2 & Agreeableness     & 3.87 & 4.11 & 3.65 & 4.00 & 0.69 & 0.50 & 1.05 & 0.60 \\
3 & Conscientiousness & 3.89 & 3.58 & 3.31 & 3.51 & 0.80 & 0.64 & 0.92 & 0.56 \\
4 & Neuroticism       & 2.70 & 2.80 & 2.61 & 3.06 & 1.02 & 0.52 & 0.83 & 0.70 \\
5 & Openness          & 3.97 & 3.79 & 3.35 & 3.13 & 0.76 & 0.59 & 1.00 & 0.65 \\

\end{longtable}

\begin{longtable}{c p{4.15cm} *{8}{S}}
\caption{GPT-4o descriptive statistics results.}
\label{tab:GPT-4o} \\

\toprule
 & & \multicolumn{4}{c}{\textbf{$\mu$}} & \multicolumn{4}{c}{\textbf{$\sigma$}} \\
\cmidrule(lr){3-6}\cmidrule(lr){7-10}
{\textbf{No.}} & {\textbf{Content}} & {Human} & {Persona} & {Shape} & {PSI} & {Human} & {Persona} & {Shape} & {PSI} \\
\midrule
\endfirsthead

\multicolumn{10}{@{}l}{\tablename\ \thetable{} -- continued from previous page} \\
\toprule
 & & \multicolumn{4}{c}{\textbf{$\mu$}} & \multicolumn{4}{c}{\textbf{$\sigma$}} \\
\cmidrule(lr){3-6}\cmidrule(lr){7-10}
{\textbf{No.}} & {\textbf{Content}} & {Human} & {Persona} & {Shape} & {PSI} & {Human} & {Persona} & {Shape} & {PSI} \\
\midrule
\endhead

\midrule
\multicolumn{10}{r@{}}{{Continued on next page}} \\
\endfoot

\bottomrule
\endlastfoot

\multicolumn{10}{c}{\textbf{Item Level}} \\
\midrule
1  & I am someone who is outgoing, sociable                          & 3.02 & 3.41 & 2.76 & 2.67 & 1.40 & 0.87 & 1.22 & 0.96 \\
2  & I am someone who is compassionate, has a soft heart             & 4.34 & 3.51 & 3.10 & 4.07 & 0.90 & 0.84 & 1.22 & 0.93 \\
3  & I am someone who tends to be disorganized                       & 3.62 & 3.39 & 3.32 & 3.34 & 1.36 & 0.78 & 1.17 & 0.76 \\
4  & I am someone who is relaxed, handles stress well                & 2.56 & 3.20 & 3.47 & 2.92 & 1.30 & 0.79 & 1.16 & 0.94 \\
5  & I am someone who has few artistic interests                    & 3.68 & 3.72 & 3.27 & 2.79 & 1.33 & 1.04 & 1.25 & 1.38 \\
6  & I am someone who has an assertive personality                  & 2.96 & 3.08 & 2.82 & 2.57 & 1.38 & 0.79 & 1.18 & 0.77 \\
7  & I am someone who is respectful, treats others with respect     & 4.62 & 3.38 & 3.48 & 4.15 & 0.61 & 0.89 & 1.23 & 0.76 \\
8  & I am someone who tends to be lazy                              & 3.62 & 4.00 & 3.51 & 3.56 & 1.31 & 0.87 & 1.28 & 0.79 \\
9  & I am someone who stays optimistic after experiencing a setback & 2.43 & 2.74 & 2.95 & 2.19 & 1.24 & 0.84 & 1.34 & 0.91 \\
10 & I am someone who is curious about many different things        & 4.43 & 4.43 & 3.86 & 3.70 & 0.81 & 0.74 & 1.26 & 0.96 \\
11 & I am someone who rarely feels excited or eager                 & 3.54 & 4.06 & 3.09 & 3.20 & 1.26 & 0.73 & 1.46 & 0.80 \\
12 & I am someone who tends to find fault with others               & 3.34 & 4.00 & 3.45 & 3.79 & 1.32 & 0.44 & 1.37 & 0.57 \\
13 & I am someone who is dependable, steady                         & 4.33 & 3.78 & 3.62 & 4.27 & 0.86 & 0.84 & 1.36 & 0.78 \\
14 & I am someone who is moody, has up and down mood swings         & 2.56 & 2.80 & 2.63 & 3.02 & 1.36 & 0.66 & 1.38 & 0.78 \\
15 & I am someone who is inventive, finds clever ways to do things  & 3.90 & 3.62 & 3.45 & 3.07 & 1.03 & 0.67 & 1.20 & 0.63 \\
16 & I am someone who tends to be quiet                             & 2.35 & 3.21 & 2.84 & 2.22 & 1.34 & 0.89 & 1.15 & 0.97 \\
17 & I am someone who feels little sympathy for others              & 3.71 & 4.68 & 3.51 & 4.24 & 1.45 & 0.60 & 1.52 & 0.63 \\
18 & I am someone who is systematic, likes to keep things in order  & 3.96 & 3.06 & 3.21 & 3.16 & 1.06 & 0.96 & 1.33 & 0.94 \\
19 & I am someone who can be tense                                 & 3.29 & 3.49 & 3.30 & 3.78 & 1.27 & 0.60 & 1.00 & 0.77 \\
20 & I am someone who is fascinated by art, music, or literature    & 4.06 & 3.49 & 3.53 & 3.38 & 1.16 & 1.04 & 1.23 & 1.18 \\
21 & I am someone who is dominant, acts as a leader                 & 2.90 & 2.94 & 2.77 & 2.42 & 1.35 & 0.74 & 1.13 & 0.70 \\
22 & I am someone who starts arguments with others                 & 4.23 & 3.65 & 3.65 & 4.43 & 1.05 & 0.72 & 1.24 & 0.71 \\
23 & I am someone who has difficulty getting started on tasks       & 3.31 & 3.29 & 3.16 & 3.11 & 1.39 & 0.69 & 1.26 & 0.73 \\
24 & I am someone who feels secure, comfortable with self          & 2.26 & 2.75 & 3.14 & 2.67 & 1.27 & 0.88 & 1.42 & 0.98 \\
25 & I am someone who avoids intellectual, philosophical discussions& 3.94 & 3.40 & 3.23 & 2.83 & 1.21 & 0.79 & 1.25 & 1.11 \\
26 & I am someone who is less active than other people             & 3.35 & 3.80 & 3.36 & 2.65 & 1.31 & 0.99 & 1.24 & 0.97 \\
27 & I am someone who has a forgiving nature                      & 3.79 & 3.05 & 3.05 & 3.51 & 1.20 & 0.57 & 1.20 & 0.86 \\
28 & I am someone who can be somewhat careless                     & 3.56 & 3.30 & 3.16 & 3.06 & 1.25 & 0.83 & 1.18 & 0.67 \\
29 & I am someone who is emotionally stable, not easily upset      & 2.44 & 3.16 & 3.17 & 2.72 & 1.27 & 0.70 & 1.31 & 1.06 \\
30 & I am someone who has little creativity                       & 3.95 & 4.10 & 3.59 & 3.38 & 1.20 & 0.76 & 1.22 & 0.84 \\
31 & I am someone who is sometimes shy, introverted                & 2.35 & 2.85 & 2.86 & 2.13 & 1.36 & 0.87 & 1.07 & 1.04 \\
32 & I am someone who is helpful and unselfish with others         & 4.18 & 3.32 & 3.27 & 4.07 & 0.88 & 0.80 & 1.21 & 0.74 \\
33 & I am someone who keeps things neat and tidy                   & 3.71 & 2.99 & 3.07 & 3.06 & 1.24 & 0.64 & 1.09 & 0.74 \\
34 & I am someone who worries a lot                                & 3.30 & 2.96 & 2.79 & 3.65 & 1.47 & 0.70 & 1.23 & 1.18 \\
35 & I am someone who values art and beauty                       & 4.16 & 3.65 & 3.61 & 3.36 & 1.07 & 0.82 & 1.18 & 1.00 \\
36 & I am someone who finds it hard to influence people           & 3.09 & 3.32 & 3.25 & 2.74 & 1.20 & 0.68 & 1.26 & 0.69 \\
37 & I am someone who is sometimes rude to others                 & 3.79 & 3.24 & 3.50 & 3.68 & 1.25 & 0.66 & 1.31 & 0.78 \\
38 & I am someone who is efficient, gets things done              & 4.23 & 3.31 & 3.07 & 3.70 & 0.88 & 0.82 & 1.25 & 0.82 \\
39 & I am someone who often feels sad                             & 2.71 & 2.56 & 2.67 & 3.39 & 1.43 & 0.71 & 1.14 & 0.82 \\
40 & I am someone who is complex, a deep thinker                  & 4.02 & 3.66 & 3.53 & 3.45 & 1.08 & 0.69 & 0.97 & 0.87 \\
41 & I am someone who is full of energy                           & 3.14 & 3.45 & 2.99 & 2.83 & 1.30 & 0.94 & 1.23 & 0.87 \\
42 & I am someone who is suspicious of others’ intentions         & 2.75 & 3.14 & 3.27 & 3.42 & 1.30 & 0.58 & 1.26 & 0.99 \\
43 & I am someone who is reliable, can always be counted on        & 4.32 & 3.84 & 3.85 & 4.21 & 0.88 & 0.83 & 1.20 & 0.81 \\
44 & I am someone who keeps their emotions under control          & 2.22 & 2.55 & 2.45 & 2.46 & 1.14 & 0.70 & 1.24 & 0.78 \\
45 & I am someone who has difficulty imagining things             & 4.17 & 4.06 & 3.57 & 3.23 & 1.06 & 0.77 & 1.25 & 0.95 \\
46 & I am someone who is talkative                                & 2.87 & 3.11 & 2.87 & 3.31 & 1.38 & 0.70 & 1.04 & 1.09 \\
47 & I am someone who can be cold and uncaring                    & 3.84 & 3.95 & 3.60 & 4.08 & 1.23 & 0.80 & 1.35 & 0.79 \\
48 & I am someone who leaves a mess, doesn’t clean up             & 4.15 & 3.47 & 3.39 & 3.31 & 1.12 & 0.71 & 1.26 & 0.55 \\
49 & I am someone who rarely feels anxious or afraid             & 3.44 & 3.20 & 2.85 & 3.50 & 1.37 & 0.83 & 1.15 & 1.02 \\
50 & I am someone who thinks poetry and plays are boring          & 3.66 & 3.54 & 3.44 & 2.93 & 1.37 & 0.75 & 1.38 & 0.89 \\
51 & I am someone who prefers to have others take charge          & 2.96 & 3.54 & 3.39 & 3.51 & 1.28 & 0.69 & 1.27 & 0.71 \\
52 & I am someone who is polite, courteous to others              & 4.50 & 3.28 & 3.33 & 3.96 & 0.70 & 0.65 & 1.14 & 0.75 \\
53 & I am someone who is persistent, works until the task is finished & 4.28 & 3.56 & 3.01 & 4.09 & 0.91 & 0.87 & 1.26 & 0.80 \\
54 & I am someone who tends to feel depressed, blue               & 2.65 & 2.70 & 2.60 & 3.28 & 1.46 & 0.79 & 1.31 & 1.07 \\
55 & I am someone who has little interest in abstract ideas      & 3.83 & 3.25 & 3.08 & 2.55 & 1.22 & 0.96 & 1.22 & 1.17 \\
56 & I am someone who shows a lot of enthusiasm                   & 3.35 & 3.86 & 2.81 & 2.70 & 1.22 & 0.85 & 1.34 & 0.85 \\
57 & I am someone who assumes the best about people               & 3.37 & 3.54 & 3.72 & 3.48 & 1.28 & 0.75 & 1.23 & 0.95 \\
58 & I am someone who sometimes behaves irresponsibly             & 3.64 & 3.27 & 3.00 & 3.12 & 1.30 & 0.74 & 1.14 & 0.66 \\
59 & I am someone who is temperamental, gets emotional easily     & 2.53 & 2.86 & 2.71 & 2.82 & 1.33 & 0.71 & 1.32 & 0.84 \\
60 & I am someone who is original, comes up with new ideas        & 3.84 & 3.28 & 2.96 & 3.07 & 1.07 & 0.73 & 1.19 & 0.68 \\
\midrule
\multicolumn{10}{c}{\textbf{Facet Level}} \\
\midrule
1  & Sociability          & 2.65 & 3.26 & 2.94 & 2.57 & 1.17 & 0.69 & 0.95 & 0.91 \\
2  & Assertiveness        & 2.98 & 3.22 & 3.06 & 2.81 & 1.04 & 0.55 & 1.10 & 0.57 \\
3  & Energy Level         & 3.35 & 3.85 & 3.09 & 2.79 & 0.95 & 0.75 & 1.12 & 0.78 \\
4  & Compassion           & 4.02 & 3.63 & 3.42 & 4.11 & 0.81 & 0.69 & 1.18 & 0.68 \\
5  & Respectfulness       & 4.29 & 3.39 & 3.49 & 4.05 & 0.70 & 0.58 & 1.13 & 0.62 \\
6  & Trust                & 3.31 & 3.15 & 3.21 & 3.49 & 1.00 & 0.44 & 1.17 & 0.78 \\
7  & Organization         & 3.86 & 3.21 & 3.18 & 3.22 & 1.01 & 0.57 & 0.98 & 0.63 \\
8  & Productiveness       & 3.86 & 3.54 & 3.19 & 3.62 & 0.91 & 0.65 & 1.12 & 0.67 \\
9  & Responsibility       & 3.96 & 3.24 & 3.03 & 3.59 & 0.84 & 0.59 & 1.10 & 0.63 \\
10 & Anxiety              & 3.15 & 3.15 & 3.06 & 3.44 & 1.11 & 0.54 & 0.95 & 0.89 \\
11 & Depression           & 2.51 & 2.76 & 2.86 & 2.85 & 1.14 & 0.56 & 1.16 & 0.88 \\
12 & Emotional Volatility & 2.44 & 3.07 & 2.92 & 2.82 & 1.10 & 0.43 & 1.06 & 0.78 \\
13 & Intellectual Curiosity& 4.05 & 3.55 & 3.13 & 3.09 & 0.83 & 0.68 & 1.04 & 0.89 \\
14 & Aesthetic Sensitivity & 3.89 & 3.63 & 3.25 & 3.15 & 1.01 & 0.75 & 1.12 & 0.97 \\
15 & Creative Imagination  & 3.96 & 3.72 & 3.29 & 3.19 & 0.88 & 0.57 & 1.05 & 0.67 \\
\midrule
\multicolumn{10}{c}{\textbf{Domain Level}} \\
\midrule
1 & Extraversion      & 2.99 & 3.44 & 3.03 & 2.72 & 0.85 & 0.57 & 0.95 & 0.67 \\
2 & Agreeableness     & 3.87 & 3.39 & 3.37 & 3.89 & 0.69 & 0.51 & 1.11 & 0.62 \\
3 & Conscientiousness & 3.89 & 3.33 & 3.13 & 3.47 & 0.80 & 0.53 & 1.00 & 0.58 \\
4 & Neuroticism       & 2.70 & 2.99 & 2.94 & 3.04 & 1.02 & 0.43 & 0.97 & 0.77 \\
5 & Openness          & 3.97 & 3.63 & 3.22 & 3.14 & 0.76 & 0.58 & 1.03 & 0.73 \\

\end{longtable}

\begingroup
\sisetup{detect-weight,detect-family,mode=text,reset-text-series=false,reset-text-family=false}

\begin{table*}[h]
\centering
\setlength{\tabcolsep}{6pt}
\renewcommand{\arraystretch}{0.95}
\begin{tabular}{l l
                S S
                S S
                S S}
\toprule
\multirow{2}{*}{\textbf{Model}} &
\multirow{2}{*}{\textbf{Method}} &
\multicolumn{2}{c}{\textbf{Item}} &
\multicolumn{2}{c}{\textbf{Facet}} &
\multicolumn{2}{c}{\textbf{Domain}} \\
\cmidrule(lr){3-4} \cmidrule(lr){5-6} \cmidrule(lr){7-8}
 &  & {$\mu$\textsubscript{MAE}~$\downarrow$} & {$\sigma$\textsubscript{MAE}~$\downarrow$} 
    & {$\mu$\textsubscript{MAE}~$\downarrow$} & {$\sigma$\textsubscript{MAE}~$\downarrow$} 
    & {$\mu$\textsubscript{MAE}~$\downarrow$} & {$\sigma$\textsubscript{MAE}~$\downarrow$} \\
\midrule
\multirow{3}{*}{Mistral-7B}
 & Persona & 0.42 & 0.46 & 0.31 & 0.41 & 0.28 & 0.36 \\
 & Shape   & 0.40 & 0.25 & 0.31 & 0.18 & 0.25 & 0.13 \\
 & PSI  & 0.51 & 0.43 & 0.32 & 0.33 & 0.29 & 0.26 \\
\midrule
\multirow{3}{*}{Gemma-2-9B}
 & Persona & 0.51 & 0.62 & 0.48 & 0.49 & 0.49 & 0.41 \\
 & Shape   & 0.58 & 0.21 & 0.52 & 0.22 & 0.47 & 0.21 \\
 & PSI  & 0.68 & 0.50 & 0.65 & 0.45 & 0.64 & 0.37 \\
\midrule
\multirow{3}{*}{Gemma-2-27B}
 & Persona & 0.42 & 0.59 & 0.37 & 0.47 & 0.35 & 0.39 \\
 & Shape   & 0.53 & 0.20 & 0.47 & 0.19 & 0.44 & 0.19 \\
 & PSI  & 0.63 & 0.48 & 0.59 & 0.38 & 0.56 & 0.31 \\
\midrule
\multirow{3}{*}{Llama3-8B}
 & Persona & 0.78 & 0.45 & 0.65 & 0.53 & 0.57 & 0.48 \\
 & Shape   & 0.89 & 0.23 & 0.82 & 0.29 & 0.75 & 0.23 \\
 & PSI  & 0.64 & 0.53 & 0.54 & 0.44 & 0.54 & 0.36 \\
\midrule
\multirow{3}{*}{Llama3-70B}
 & Persona & 0.51 & 0.26 & 0.33 & 0.15 & 0.31 & 0.06 \\
 & Shape   & 0.63 & 0.25 & 0.52 & 0.29 & 0.46 & 0.32 \\
 & PSI  & 0.66 & 0.36 & 0.53 & 0.27 & 0.48 & 0.20 \\
\midrule
\multirow{3}{*}{GPT-4o-mini}
 & Persona & 0.33 & 0.44 & 0.21 & 0.33 & 0.21 & 0.25 \\
 & Shape   & 0.42 & 0.18 & 0.39 & 0.17 & 0.31 & 0.18 \\
 & PSI  & 0.45 & 0.40 & 0.40 & 0.27 & 0.40 & 0.20 \\
\midrule
\multirow{3}{*}{GPT-4o}
 & Persona & 0.48 & 0.44 & 0.42 & 0.36 & 0.42 & 0.30 \\
 & Shape   & 0.58 & 0.16 & 0.50 & 0.18 & 0.46 & 0.21 \\
 & PSI  & 0.47 & 0.34 & 0.40 & 0.22 & 0.38 & 0.15 \\
\bottomrule
\end{tabular}
\caption{MAE results of $\mu$ and $\sigma$ for Item, Facet, and Domain across different Backbone Model + Method combinations.}
\label{tab:mae_results}
\end{table*}

\endgroup

\clearpage

\begin{longtable}{
  l
  >{\centering\arraybackslash}p{2cm}
  >{\centering\arraybackslash}p{0.8cm}
  >{\centering\arraybackslash}p{1.4cm}
  >{\centering\arraybackslash}p{1.4cm}
  >{\centering\arraybackslash}p{1.4cm}
  >{\centering\arraybackslash}p{1.4cm}
}

\caption{Model fit indices for BFI-2 TFMs of each domain and FFM.}
\label{tab:modelfit} \\

\toprule
\textbf{Model} & \textbf{Chi‑square} & \textbf{\textit{df}} & \textbf{CFI} & \textbf{TLI} & \textbf{RMSEA} & \textbf{SRMR} \\
\midrule
\endfirsthead

\multicolumn{7}{@{}l}{\tablename\ \thetable{} -- continued from previous page} \\
\toprule
\textbf{Model} & \textbf{Chi‑square} & \textbf{\textit{df}} & \textbf{CFI} & \textbf{TLI} & \textbf{RMSEA} & \textbf{SRMR} \\
\midrule
\endhead

\midrule
\multicolumn{7}{r}{{Continued on next page}} \\
\endfoot

\bottomrule
\endlastfoot

\multicolumn{7}{c}{\textbf{Extraversion TFM}} \\
\midrule
Mistral-7B+Persona & 188.882 & 51 & 0.873 & 0.836 & 0.095 & 0.066 \\
Mistral-7B+Shape   & 676.095 & 51 & 0.781 & 0.717 & 0.202 & 0.095 \\
Mistral-7B+PSI     & 395.981 & 51 & 0.828 & 0.778 & 0.138 & 0.079 \\
Gemma-2-9B+Persona       & 254.304 & 51 & 0.880 & 0.845 & 0.115 & 0.074 \\
Gemma-2-9B+Shape         &1084.518 & 51 & 0.695 & 0.605 & 0.260 & 0.113 \\
Gemma-2-9B+PSI           & 267.081 & 51 & 0.884 & 0.850 & 0.109 & 0.080 \\
Gemma-2-27B+Persona      & 353.266 & 51 & 0.860 & 0.818 & 0.141 & 0.073 \\
Gemma-2-27B+Shape        &1101.801 & 51 & 0.744 & 0.668 & 0.262 & 0.154 \\
Gemma-2-27B+PSI          & 290.205 & 51 & 0.900 & 0.871 & 0.115 & 0.057 \\
Llama3-8B+Persona  & 430.846 & 51 & 0.556 & 0.425 & 0.158 & 0.166 \\
Llama3-8B+Shape    & 553.196 & 51 & 0.607 & 0.492 & 0.181 & 0.151 \\
Llama3-8B+PSI      & 275.953 & 51 & 0.899 & 0.869 & 0.111 & 0.055 \\
Llama3-70B+Persona & 316.715 & 51 & 0.885 & 0.851 & 0.132 & 0.060 \\
Llama3-70B+Shape   & 944.452 & 51 & 0.681 & 0.588 & 0.242 & 0.130 \\
Llama3-70B+PSI     & 367.869 & 51 & 0.815 & 0.761 & 0.132 & 0.074 \\
GPT-4o-mini+Persona         & 362.974 & 51 & 0.874 & 0.837 & 0.143 & 0.065 \\
GPT-4o-mini+Shape           &1674.008 & 51 & 0.504 & 0.359 & 0.326 & 0.264 \\
GPT-4o-mini+PSI             & 369.695 & 51 & 0.898 & 0.868 & 0.132 & 0.063 \\
GPT-4o+Persona               & 261.099 & 51 & 0.886 & 0.852 & 0.117 & 0.058 \\
GPT-4o+Shape                 & 663.947 & 51 & 0.823 & 0.771 & 0.200 & 0.101 \\
GPT-4o+PSI                   & 376.669 & 51 & 0.895 & 0.864 & 0.134 & 0.055 \\
Human                & 994.226 & 51 & 0.892 & 0.861 & 0.109 & 0.057 \\

\midrule
\multicolumn{7}{c}{\textbf{Agreeableness TFM}} \\
\midrule
Mistral-7B+Persona & 294.949 & 51 & 0.818 & 0.764 & 0.126 & 0.091 \\
Mistral-7B+Shape   & 578.669 & 51 & 0.802 & 0.744 & 0.186 & 0.102 \\
Mistral-7B+PSI     & 298.885 & 51 & 0.908 & 0.881 & 0.117 & 0.061 \\
Gemma-2-9B+Persona       & 355.402 & 51 & 0.853 & 0.810 & 0.141 & 0.073 \\
Gemma-2-9B+Shape         & 771.602 & 51 & 0.881 & 0.845 & 0.217 & 0.042 \\
Gemma-2-9B+PSI           & 257.858 & 51 & 0.914 & 0.889 & 0.107 & 0.067 \\
Gemma-2-27B+Persona      & 377.813 & 51 & 0.863 & 0.823 & 0.146 & 0.068 \\
Gemma-2-27B+Shape        & 865.382 & 51 & 0.863 & 0.823 & 0.231 & 0.058 \\
Gemma-2-27B+PSI          & 347.315 & 51 & 0.900 & 0.870 & 0.128 & 0.065 \\
Llama3-8B+Persona  & 588.254 & 51 & 0.655 & 0.554 & 0.187 & 0.179 \\
Llama3-8B+Shape    &1183.880 & 51 & 0.557 & 0.426 & 0.272 & 0.222 \\
Llama3-8B+PSI      & 241.972 & 51 & 0.898 & 0.868 & 0.102 & 0.071 \\
Llama3-70B+Persona & 616.199 & 51 & 0.798 & 0.739 & 0.192 & 0.088 \\
Llama3-70B+Shape   & 884.948 & 51 & 0.841 & 0.794 & 0.233 & 0.069 \\
Llama3-70B+PSI     & 295.615 & 51 & 0.924 & 0.901 & 0.116 & 0.048 \\
GPT-4o-mini+Persona         & 340.628 & 51 & 0.884 & 0.849 & 0.138 & 0.075 \\
GPT-4o-mini+Shape           &1213.078 & 51 & 0.731 & 0.652 & 0.276 & 0.154 \\
GPT-4o-mini+PSI             & 212.855 & 51 & 0.948 & 0.933 & 0.094 & 0.046 \\
GPT-4o+Persona               & 210.863 & 51 & 0.900 & 0.870 & 0.102 & 0.062 \\
GPT-4o+Shape                 & 525.799 & 51 & 0.909 & 0.882 & 0.176 & 0.039 \\
GPT-4o+PSI                   & 182.956 & 51 & 0.957 & 0.945 & 0.085 & 0.045 \\
Human                & 903.288 & 51 & 0.875 & 0.838 & 0.104 & 0.060 \\

\midrule
\multicolumn{7}{c}{\textbf{Conscientiousness TFM}} \\
\midrule
Mistral-7B+Persona & 415.371 & 51 & 0.809 & 0.753 & 0.154 & 0.103 \\
Mistral-7B+Shape   & 979.589 & 51 & 0.730 & 0.650 & 0.246 & 0.144 \\
Mistral-7B+PSI     & 501.680 & 51 & 0.811 & 0.755 & 0.157 & 0.092 \\
Gemma-2-9B+Persona       & 580.705 & 51 & 0.751 & 0.678 & 0.186 & 0.114 \\
Gemma-2-9B+Shape         &1592.597 & 51 & 0.653 & 0.551 & 0.317 & 0.122 \\
Gemma-2-9B+PSI           & 330.247 & 51 & 0.848 & 0.804 & 0.124 & 0.087 \\
Gemma-2-27B+Persona      & 445.890 & 51 & 0.837 & 0.789 & 0.161 & 0.080 \\
Gemma-2-27B+Shape        &1666.873 & 51 & 0.678 & 0.584 & 0.325 & 0.114 \\
Gemma-2-27B+PSI          & 354.743 & 51 & 0.861 & 0.821 & 0.129 & 0.075 \\
Llama3-8B+Persona  & 357.786 & 51 & 0.783 & 0.719 & 0.142 & 0.117 \\
Llama3-8B+Shape    & 987.900 & 51 & 0.568 & 0.441 & 0.247 & 0.215 \\
Llama3-8B+PSI      & 437.070 & 51 & 0.805 & 0.748 & 0.146 & 0.114 \\
Llama3-70B+Persona & 705.359 & 51 & 0.768 & 0.699 & 0.207 & 0.109 \\
Llama3-70B+Shape   &1498.095 & 51 & 0.645 & 0.541 & 0.308 & 0.192 \\
Llama3-70B+PSI     & 467.824 & 51 & 0.850 & 0.806 & 0.151 & 0.062 \\
GPT-4o-mini+Persona         & 625.986 & 51 & 0.810 & 0.755 & 0.194 & 0.080 \\
GPT-4o-mini+Shape           &1215.114 & 51 & 0.614 & 0.500 & 0.276 & 0.200 \\
GPT-4o-mini+PSI             & 396.471 & 51 & 0.882 & 0.848 & 0.138 & 0.056 \\
GPT-4o+Persona               & 385.339 & 51 & 0.780 & 0.716 & 0.148 & 0.088 \\
GPT-4o+Shape                 & 705.340 & 51 & 0.826 & 0.775 & 0.207 & 0.069 \\
GPT-4o+PSI                   & 411.654 & 51 & 0.883 & 0.849 & 0.141 & 0.057 \\
Human                &1045.113 & 51 & 0.897 & 0.867 & 0.112 & 0.058 \\

\midrule
\multicolumn{7}{c}{\textbf{Neuroticism TFM}} \\
\midrule
Mistral-7B+Persona & 530.516 & 51 & 0.585 & 0.462 & 0.177 & 0.158 \\
Mistral-7B+Shape   & 557.616 & 51 & 0.810 & 0.754 & 0.182 & 0.095 \\
Mistral-7B+PSI     & 431.937 & 51 & 0.841 & 0.794 & 0.145 & 0.066 \\
Gemma-2-9B+Persona       & 469.967 & 51 & 0.816 & 0.762 & 0.165 & 0.096 \\
Gemma-2-9B+Shape         &1072.848 & 51 & 0.718 & 0.635 & 0.258 & 0.134 \\
Gemma-2-9B+PSI           & 611.373 & 51 & 0.787 & 0.724 & 0.175 & 0.094 \\
Gemma-2-27B+Persona      & 598.476 & 51 & 0.743 & 0.668 & 0.189 & 0.118 \\
Gemma-2-27B+Shape        & 759.167 & 51 & 0.839 & 0.791 & 0.215 & 0.085 \\
Gemma-2-27B+PSI          & 478.337 & 51 & 0.836 & 0.787 & 0.153 & 0.082 \\
Llama3-8B+Persona  & 540.289 & 51 & 0.586 & 0.464 & 0.179 & 0.150 \\
Llama3-8B+Shape    & 956.346 & 51 & 0.545 & 0.412 & 0.243 & 0.257 \\
Llama3-8B+PSI      & 522.249 & 51 & 0.814 & 0.760 & 0.161 & 0.081 \\
Llama3-70B+Persona & 346.207 & 51 & 0.884 & 0.850 & 0.139 & 0.059 \\
Llama3-70B+Shape   & 808.502 & 51 & 0.776 & 0.710 & 0.223 & 0.099 \\
Llama3-70B+PSI     & 525.423 & 51 & 0.855 & 0.812 & 0.161 & 0.065 \\
GPT-4o-mini+Persona         & 477.940 & 51 & 0.803 & 0.745 & 0.167 & 0.076 \\
GPT-4o-mini+Shape           & 856.807 & 51 & 0.667 & 0.569 & 0.229 & 0.151 \\
GPT-4o-mini+PSI             & 539.536 & 51 & 0.872 & 0.834 & 0.164 & 0.070 \\
GPT-4o+Persona               & 304.611 & 51 & 0.756 & 0.684 & 0.129 & 0.091 \\
GPT-4o+Shape                 & 960.437 & 51 & 0.720 & 0.638 & 0.244 & 0.099 \\
GPT-4o+PSI                   & 541.928 & 51 & 0.874 & 0.837 & 0.164 & 0.069 \\
Human                & 927.971 & 51 & 0.931 & 0.911 & 0.105 & 0.052 \\

\midrule
\multicolumn{7}{c}{\textbf{Openness TFM}} \\
\midrule
Mistral-7B+Persona & 316.958 & 51 & 0.748 & 0.674 & 0.132 & 0.119 \\
Mistral-7B+Shape   & 962.242 & 51 & 0.653 & 0.551 & 0.244 & 0.149 \\
Mistral-7B+PSI     & 455.750 & 51 & 0.722 & 0.640 & 0.149 & 0.114 \\
Gemma-2-9B+Persona       & 313.753 & 51 & 0.772 & 0.704 & 0.131 & 0.103 \\
Gemma-2-9B+Shape         & 737.946 & 51 & 0.837 & 0.789 & 0.212 & 0.055 \\
Gemma-2-9B+PSI           & 380.816 & 51 & 0.748 & 0.674 & 0.135 & 0.105 \\
Gemma-2-27B+Persona      & 381.298 & 51 & 0.823 & 0.771 & 0.147 & 0.103 \\
Gemma-2-27B+Shape        & 702.555 & 51 & 0.862 & 0.821 & 0.206 & 0.051 \\
Gemma-2-27B+PSI          & 471.557 & 51 & 0.831 & 0.781 & 0.152 & 0.116 \\
Llama3-8B+Persona  & 211.861 & 51 & 0.806 & 0.749 & 0.103 & 0.087 \\
Llama3-8B+Shape    & 825.535 & 51 & 0.637 & 0.530 & 0.225 & 0.209 \\
Llama3-8B+PSI      & 318.777 & 51 & 0.889 & 0.856 & 0.121 & 0.070 \\
Llama3-70B+Persona & 256.032 & 51 & 0.904 & 0.876 & 0.116 & 0.077 \\
Llama3-70B+Shape   & 562.061 & 51 & 0.864 & 0.823 & 0.183 & 0.057 \\
Llama3-70B+PSI     & 408.648 & 51 & 0.853 & 0.810 & 0.140 & 0.117 \\
GPT-4o-mini+Persona         & 257.955 & 51 & 0.899 & 0.869 & 0.116 & 0.068 \\
GPT-4o-mini+Shape           &1464.309 & 51 & 0.656 & 0.555 & 0.304 & 0.201 \\
GPT-4o-mini+PSI             & 307.850 & 51 & 0.914 & 0.889 & 0.119 & 0.067 \\
GPT-4o+Persona               & 226.649 & 51 & 0.910 & 0.883 & 0.107 & 0.057 \\
GPT-4o+Shape                 & 502.104 & 51 & 0.878 & 0.842 & 0.172 & 0.055 \\
GPT-4o+PSI                   & 350.639 & 51 & 0.909 & 0.882 & 0.128 & 0.069 \\
Human                & 904.843 & 51 & 0.900 & 0.870 & 0.104 & 0.064 \\

\midrule
\multicolumn{7}{c}{\textbf{FFM}} \\
\midrule
Mistral-7B+Persona &  560.227 & 80 & 0.802 & 0.740 & 0.141 & 0.088 \\
Mistral-7B+Shape   & 1457.055 & 80 & 0.756 & 0.680 & 0.240 & 0.115 \\
Mistral-7B+PSI     & 1067.765 & 80 & 0.773 & 0.701 & 0.186 & 0.111 \\
Gemma-2-9B+Persona       &  731.270 & 80 & 0.790 & 0.725 & 0.165 & 0.107 \\
Gemma-2-9B+Shape         & 2336.944 & 80 & 0.696 & 0.601 & 0.307 & 0.144 \\
Gemma-2-9B+PSI           &  627.697 & 80 & 0.836 & 0.784 & 0.138 & 0.084 \\
Gemma-2-27B+Persona      &  756.777 & 80 & 0.821 & 0.765 & 0.168 & 0.093 \\
Gemma-2-27B+Shape        & 2325.784 & 80 & 0.714 & 0.624 & 0.306 & 0.139 \\
Gemma-2-27B+PSI          &  706.399 & 80 & 0.836 & 0.784 & 0.148 & 0.101 \\
Llama3-8B+Persona  &  302.351 & 80 & 0.849 & 0.802 & 0.096 & 0.078 \\
Llama3-8B+Shape    &  839.663 & 80 & 0.784 & 0.716 & 0.178 & 0.113 \\
Llama3-8B+PSI      &  868.954 & 80 & 0.782 & 0.714 & 0.166 & 0.117 \\
Llama3-70B+Persona &  954.848 & 80 & 0.780 & 0.711 & 0.191 & 0.120 \\
Llama3-70B+Shape   & 1996.472 & 80 & 0.720 & 0.633 & 0.283 & 0.148 \\
Llama3-70B+PSI     &  901.061 & 80 & 0.788 & 0.722 & 0.170 & 0.112 \\
GPT-4o-mini+Persona         & 1079.430 & 80 & 0.779 & 0.710 & 0.204 & 0.108 \\
GPT-4o-mini+Shape           & 1350.813 & 80 & 0.797 & 0.734 & 0.230 & 0.104 \\
GPT-4o-mini+PSI             & 1144.935 & 80 & 0.760 & 0.685 & 0.193 & 0.120 \\
GPT-4o+Persona               &  551.459 & 80 & 0.819 & 0.762 & 0.140 & 0.097 \\
GPT-4o+Shape                 & 2132.208 & 80 & 0.711 & 0.621 & 0.292 & 0.143 \\
GPT-4o+PSI                   & 1029.123 & 80 & 0.776 & 0.706 & 0.182 & 0.120 \\
Human                & 1747.433 & 80 & 0.852 & 0.806 & 0.116 & 0.080 \\

\end{longtable}

\begin{longtable}{l c c c}

\caption{Tucker’s congruence coefficient for BFI-2 TFM of Extraversion.} 
\label{tcc_tfm_e}\\
\toprule
\textbf{Model} & \textbf{Sociability} & \textbf{Assertiveness} & \textbf{Energy Level} \\
\midrule
\endfirsthead
\multicolumn{4}{@{}l}{\tablename\ \thetable{} -- continued from previous page} \\
\toprule
\textbf{Model} & \textbf{Sociability} & \textbf{Assertiveness} & \textbf{Energy Level} \\
\midrule
\endhead
\midrule
\multicolumn{4}{r}{{Continued on next page}} \\
\endfoot
\bottomrule
\endlastfoot
Mistral-7B+Persona & 0.91 & 0.92 & 0.98 \\
Mistral-7B+Shape   & 0.99 & 0.94 & 0.95 \\
Mistral-7B+PSI    & 1.00 & 0.96 & 0.99 \\
Gemma-2-9B+Persona       & 1.00 & 0.97 & 1.00 \\
Gemma-2-9B+Shape         & 0.97 & 0.86 & 0.97 \\
Gemma-2-9B+PSI          & 1.00 & 0.97 & 0.86 \\
Gemma-2-27B+Persona      & 1.00 & 0.97 & 0.98 \\
Gemma-2-27B+Shape        & 1.00 & 0.97 & 0.97 \\
Gemma-2-27B+PSI         & 1.00 & 0.92 & 0.98 \\
Llama3-8B+Persona  & -0.78 & 0.55 & 0.71 \\
Llama3-8B+Shape    & -0.81 & 0.79 & -0.80 \\
Llama3-8B+PSI     & 1.00 & 0.98 & 0.98 \\
Llama3-70B+Persona & 0.99 & 0.98 & 0.98 \\
Llama3-70B+Shape   & 0.99 & 0.97 & 0.96 \\
Llama3-70B+PSI    & 1.00 & 0.95 & 0.99 \\
GPT-4o-mini+Persona          & 1.00 & 0.99 & 0.99 \\
GPT-4o-mini+Shape            & 0.91 & 0.95 & 1.00 \\
GPT-4o-mini+PSI             & 1.00 & 0.99 & 0.98 \\
GPT-4o+Persona               & 1.00 & 0.97 & 0.98 \\
GPT-4o+Shape                 & 1.00 & 0.99 & 0.97 \\
GPT-4o+PSI                  & 1.00 & 0.98 & 0.98 \\
\end{longtable}

\begin{longtable}{l c c c}
\caption{Tucker’s congruence coefficient for BFI-2 TFM of Agreeableness.}
\label{tcc_tfm_a}\\
\toprule
\textbf{Model} & \textbf{Compassion} & \textbf{Respectfulness} & \textbf{Trust} \\
\midrule
\endfirsthead
\multicolumn{4}{@{}l}{\tablename\ \thetable{} -- continued} \\ \toprule
\textbf{Model} & \textbf{Compassion} & \textbf{Respectfulness} & \textbf{Trust} \\ \midrule
\endhead
\midrule\multicolumn{4}{r}{{Continued on next page}} \\ \endfoot
\bottomrule\endlastfoot
Mistral-7B+Persona & 0.95 & 0.98 & 0.92 \\
Mistral-7B+Shape   & 0.92 & 0.97 & 0.96 \\
Mistral-7B+PSI    & 0.99 & 0.97 & 0.98 \\
Gemma-2-9B+Persona       & 0.97 & 0.99 & 0.97 \\
Gemma-2-9B+Shape         & 0.96 & 1.00 & 1.00 \\
Gemma-2-9B+PSI          & 0.99 & 0.90 & -0.87 \\
Gemma-2-27B+Persona      & 0.97 & 0.99 & 1.00 \\
Gemma-2-27B+Shape        & 0.97 & 1.00 & 0.99 \\
Gemma-2-27B+PSI         & 0.96 & 0.98 & 1.00 \\
Llama3-8B+Persona  & 0.86 & 0.71 & -0.60 \\
Llama3-8B+Shape    & 0.96 & 0.93 & 0.94 \\
Llama3-8B+PSI     & 0.99 & 0.94 & 0.97 \\
Llama3-70B+Persona & 0.96 & 1.00 & 0.99 \\
Llama3-70B+Shape   & 0.95 & 1.00 & 0.99 \\
Llama3-70B+PSI    & 1.00 & 0.99 & 1.00 \\
GPT-4o-mini+Persona          & 0.98 & 0.99 & 0.98 \\
GPT-4o-mini+Shape            & 0.91 & 0.98 & 0.94 \\
GPT-4o-mini+PSI             & 0.97 & 0.97 & 1.00 \\
GPT-4o+Persona               & 0.95 & 1.00 & 0.98 \\
GPT-4o+Shape                 & 0.97 & 1.00 & 1.00 \\
GPT-4o+PSI                  & 0.98 & 0.98 & 1.00 \\
\end{longtable}

\begin{longtable}{l c c c}
\caption{Tucker’s congruence coefficient for BFI-2 TFM of Conscientiousness.} 
\label{tcc_tfm_c} \\
\toprule
\textbf{Model} & \textbf{Organization} & \textbf{Productiveness} & \textbf{Responsibility} \\
\midrule
\endfirsthead
\multicolumn{4}{@{}l}{\tablename\ \thetable{} -- continued} \\ \toprule
\textbf{Model} & \textbf{Organization} & \textbf{Productiveness} & \textbf{Responsibility} \\ \midrule
\endhead
\midrule\multicolumn{4}{r}{{Continued on next page}} \\ \endfoot
\bottomrule\endlastfoot
Mistral-7B+Persona & 0.98 & 0.99 & 0.95 \\
Mistral-7B+Shape   & 0.93 & 0.99 & 0.97 \\
Mistral-7B+PSI    & 0.98 & 0.99 & 0.98 \\
Gemma-2-9B+Persona       & 0.98 & 0.95 & 0.98 \\
Gemma-2-9B+Shape         & 0.99 & 1.00 & 0.99 \\
Gemma-2-9B+PSI          & 0.81 & 0.98 & 0.94 \\
Gemma-2-27B+Persona      & 0.99 & 0.99 & 0.99 \\
Gemma-2-27B+Shape        & 0.98 & 1.00 & 0.99 \\
Gemma-2-27B+PSI         & 0.99 & 1.00 & 0.97 \\
Llama3-8B+Persona  & -0.43 & 0.33 & 0.51 \\
Llama3-8B+Shape    & 0.81 & 0.82 & 0.86 \\
Llama3-8B+PSI     & 0.89 & 0.99 & 0.97 \\
Llama3-70B+Persona & 1.00 & 0.99 & 0.97 \\
Llama3-70B+Shape   & 0.96 & 0.99 & 0.96 \\
Llama3-70B+PSI    & 1.00 & 0.97 & 1.00 \\
GPT-4o-mini+Persona          & 1.00 & 0.99 & 0.99 \\
GPT-4o-mini+Shape            & 0.90 & 0.98 & 0.96 \\
GPT-4o-mini+PSI             & 1.00 & 1.00 & 1.00 \\
GPT-4o+Persona               & 0.99 & 0.99 & 0.99 \\
GPT-4o+Shape                 & 0.99 & 0.99 & 0.99 \\
GPT-4o+PSI                  & 0.99 & 1.00 & 1.00 \\
\end{longtable}

\begin{longtable}{l c c c}
\caption{Tucker’s congruence coefficient for BFI-2 TFM of Neuroticism.} 
\label{tcc_tfm_n} \\
\toprule
\textbf{Model} & \textbf{Anxiety} & \textbf{Depression} & \textbf{Emotional Volatility} \\
\midrule
\endfirsthead
\multicolumn{4}{@{}l}{\tablename\ \thetable{} -- continued} \\ \toprule
\textbf{Model} & \textbf{Anxiety} & \textbf{Depression} & \textbf{Emotional Volatility} \\ \midrule
\endhead
\midrule\multicolumn{4}{r}{{Continued on next page}} \\ \endfoot
\bottomrule\endlastfoot
Mistral-7B+Persona & 0.91 & 0.93 & 0.88 \\
Mistral-7B+Shape   & 0.86 & 1.00 & 0.93 \\
Mistral-7B+PSI    & 0.99 & 0.99 & 1.00 \\
Gemma-2-9B+Persona       & 0.99 & 1.00 & 0.99 \\
Gemma-2-9B+Shape         & 0.92 & 1.00 & 0.91 \\
Gemma-2-9B+PSI          & 0.99 & 1.00 & 0.98 \\
Gemma-2-27B+Persona      & 0.97 & 1.00 & 1.00 \\
Gemma-2-27B+Shape        & 0.85 & 0.99 & 0.97 \\
Gemma-2-27B+PSI         & 0.99 & 0.99 & 0.99 \\
Llama3-8B+Persona  & 0.17 & 0.03 & -0.42 \\
Llama3-8B+Shape    & -0.15 & 0.60 & 0.32 \\
Llama3-8B+PSI     & 0.99 & 1.00 & 0.99 \\
Llama3-70B+Persona & 0.99 & 0.99 & 1.00 \\
Llama3-70B+Shape   & 0.99 & 0.99 & 0.99 \\
Llama3-70B+PSI    & 1.00 & 1.00 & 1.00 \\
GPT-4o-mini+Persona          & 1.00 & 0.99 & 0.99 \\
GPT-4o-mini+Shape            & 0.85 & 0.99 & 0.96 \\
GPT-4o-mini+PSI             & 1.00 & 1.00 & 1.00 \\
GPT-4o+Persona               & 0.99 & 0.97 & 0.97 \\
GPT-4o+Shape                 & 0.98 & 0.99 & 0.99 \\
GPT-4o+PSI                  & 1.00 & 1.00 & 1.00 \\
\end{longtable}

\begin{longtable}{l c c c}
\caption{Tucker’s congruence coefficient for BFI-2 TFM of Openness.}  
\label{tcc_tfm_o} \\
\toprule
\textbf{Model} & \textbf{Intellectual Curiosity} & \textbf{Aesthetic Sensitivity} & \textbf{Creative Imagination} \\
\midrule
\endfirsthead
\multicolumn{4}{@{}l}{\tablename\ \thetable{} -- continued} \\ \toprule
\textbf{Model} & \textbf{Intellectual Curiosity} & \textbf{Aesthetic Sensitivity} & \textbf{Creative Imagination} \\ \midrule
\endhead
\midrule\multicolumn{4}{r}{{Continued on next page}} \\ \endfoot
\bottomrule\endlastfoot
Mistral-7B+Persona & 1.00 & 0.95 & 0.94 \\
Mistral-7B+Shape   & 0.96 & 0.99 & 0.98 \\
Mistral-7B+PSI    & 0.81 & 0.94 & 0.86 \\
Gemma-2-9B+Persona       & 0.96 & 0.94 & 0.87 \\
Gemma-2-9B+Shape         & 0.99 & 0.99 & 0.99 \\
Gemma-2-9B+PSI          & 0.96 & -0.62 & 0.70 \\
Gemma-2-27B+Persona      & 0.98 & 0.98 & 0.99 \\
Gemma-2-27B+Shape        & 0.95 & 0.99 & 0.99 \\
Gemma-2-27B+PSI         & 0.98 & 0.98 & 1.00 \\
Llama3-8B+Persona  & 0.46 & -0.70 & 0.34 \\
Llama3-8B+Shape    & 0.69 & 0.94 & 0.72 \\
Llama3-8B+PSI     & 1.00 & 0.98 & 0.98 \\
Llama3-70B+Persona & 1.00 & 0.98 & 0.98 \\
Llama3-70B+Shape   & 0.98 & 0.98 & 1.00 \\
Llama3-70B+PSI    & 0.98 & 0.97 & 0.96 \\
GPT-4o-mini+Persona          & 0.99 & 1.00 & 0.99 \\
GPT-4o-mini+Shape            & 0.93 & 0.93 & 0.91 \\
GPT-4o-mini+PSI             & 1.00 & 0.98 & 1.00 \\
GPT-4o+Persona               & 1.00 & 0.99 & 1.00 \\
GPT-4o+Shape                 & 0.99 & 0.99 & 1.00 \\
GPT-4o+PSI                  & 1.00 & 0.99 & 0.99 \\
\end{longtable}

\begin{table*}[h]
\caption{Tucker’s congruence coefficient for BFI-2 FFM.}
\label{tab:tcc_ffm}
\centering

\begin{tabular}{l*{5}{c}}

\toprule
\textbf{Model}
 & \textbf{Extraversion}  & \textbf{Agreeableness} & \textbf{Conscientiousness} & \textbf{Neuroticism} & \textbf{Openness} \\
\midrule
Mistral-7B+Persona & 0.99 & 1.00 & 1.00 & 0.99 & 1.00 \\
Mistral-7B+Shape   & 0.99 & 0.99 & 0.99 & 0.99 & 0.99 \\
Mistral-7B+PSI     & 1.00 & 1.00 & 1.00 & 0.99 & 0.99 \\
Gemma-2-9B+Persona       & 0.99 & 1.00 & 0.98 & 0.99 & 0.99 \\
Gemma-2-9B+Shape         & 1.00 & 0.99 & 0.98 & 0.99 & 1.00 \\
Gemma-2-9B+PSI           & 0.99 & 1.00 & 0.99 & 0.99 & 0.96 \\
Gemma-2-27B+Persona      & 1.00 & 1.00 & 0.99 & 1.00 & 0.99 \\
Gemma-2-27B+Shape        & 1.00 & 0.99 & 0.99 & 0.99 & 1.00 \\
Gemma-2-27B+PSI          & 1.00 & 1.00 & 1.00 & 0.99 & 1.00 \\
Llama3-8B+Persona  & 0.99 & 1.00 & 0.97 & 0.96 & 0.99 \\
Llama3-8B+Shape    & 0.98 & 0.99 & 0.99 & 0.96 & 0.99 \\
Llama3-8B+PSI      & 1.00 & 0.99 & 1.00 & 1.00 & 1.00 \\
Llama3-70B+Persona & 0.99 & 1.00 & 0.98 & 1.00 & 0.99 \\
Llama3-70B+Shape   & 1.00 & 0.99 & 0.99 & 1.00 & 0.99 \\
Llama3-70B+PSI     & 1.00 & 1.00 & 0.99 & 1.00 & 0.99 \\
GPT-4o-mini+Persona         & 1.00 & 1.00 & 0.99 & 0.99 & 1.00 \\
GPT-4o-mini+Shape           & 0.99 & 0.99 & 0.99 & 1.00 & 1.00 \\
GPT-4o-mini+PSI             & 1.00 & 1.00 & 0.99 & 0.99 & 1.00 \\
GPT-4o+Persona               & 1.00 & 0.99 & 0.99 & 0.99 & 1.00 \\
GPT-4o+Shape                 & 1.00 & 1.00 & 0.99 & 0.99 & 1.00 \\
GPT-4o+PSI                   & 1.00 & 1.00 & 0.99 & 0.99 & 1.00 \\
\bottomrule
\end{tabular}

\end{table*}

\begin{longtable}{l c c c}
\caption{Factor loading MAEs for BFI-2 TFM of Extraversion.}
\label{tab:mae_extraversion} \\
\toprule
\textbf{Model} & \textbf{Sociability} & \textbf{Assertiveness} & \textbf{Energy Level} \\
\midrule
\endfirsthead
\multicolumn{4}{@{}l}{\tablename\ \thetable{} -- continued} \\ \toprule
\textbf{Model} & \textbf{Sociability} & \textbf{Assertiveness} & \textbf{Energy Level} \\ \midrule
\endhead
\midrule\multicolumn{4}{r}{{Continued on next page}} \\ \endfoot
\bottomrule\endlastfoot
Mistral-7B+Persona & 0.09 & 0.08 & 0.04 \\
Mistral-7B+Shape   & 0.03 & 0.07 & 0.07 \\
Mistral-7B+PSI    & 0.02 & 0.07 & 0.02 \\
Gemma-2-9B+Persona       & 0.01 & 0.05 & 0.02 \\
Gemma-2-9B+Shape         & 0.04 & 0.10 & 0.07 \\
Gemma-2-9B+PSI          & 0.01 & 0.05 & 0.09 \\
Gemma-2-27B+Persona      & 0.02 & 0.06 & 0.05 \\
Gemma-2-27B+Shape        & 0.02 & 0.06 & 0.08 \\
Gemma-2-27B+PSI         & 0.02 & 0.09 & 0.04 \\
Llama3-8B+Persona  & 0.41 & 0.16 & 0.14 \\
Llama3-8B+Shape    & 0.40 & 0.11 & 0.33 \\
Llama3-8B+PSI     & 0.02 & 0.04 & 0.04 \\
Llama3-70B+Persona & 0.03 & 0.05 & 0.05 \\
Llama3-70B+Shape   & 0.05 & 0.05 & 0.06 \\
Llama3-70B+PSI    & 0.03 & 0.09 & 0.03 \\
GPT-4o-mini+Persona          & 0.01 & 0.04 & 0.05 \\
GPT-4o-mini+Shape            & 0.09 & 0.07 & 0.03 \\
GPT-4o-mini+PSI             & 0.02 & 0.03 & 0.05 \\
GPT-4o+Persona               & 0.01 & 0.06 & 0.05 \\
GPT-4o+Shape                 & 0.02 & 0.05 & 0.06 \\
GPT-4o+PSI                  & 0.02 & 0.04 & 0.06 \\
\end{longtable}

\begin{longtable}{l c c c}
\caption{Factor loading MAEs for BFI-2 TFM of Agreeableness.} 
\label{tab:mae_agreeableness} \\
\toprule
\textbf{Model} & \textbf{Compassion} & \textbf{Respectfulness} & \textbf{Trust} \\
\midrule
\endfirsthead
\multicolumn{4}{@{}l}{\tablename\ \thetable{} -- continued} \\ \toprule
\textbf{Model} & \textbf{Compassion} & \textbf{Respectfulness} & \textbf{Trust} \\ \midrule
\endhead
\midrule\multicolumn{4}{r}{{Continued on next page}} \\ \endfoot
\bottomrule\endlastfoot
Mistral-7B+Persona & 0.06 & 0.04 & 0.07 \\
Mistral-7B+Shape   & 0.08 & 0.06 & 0.07 \\
Mistral-7B+PSI    & 0.03 & 0.06 & 0.04 \\
Gemma-2-9B+Persona       & 0.06 & 0.02 & 0.05 \\
Gemma-2-9B+Shape         & 0.10 & 0.08 & 0.07 \\
Gemma-2-9B+PSI          & 0.05 & 0.09 & 0.42 \\
Gemma-2-27B+Persona      & 0.05 & 0.04 & 0.01 \\
Gemma-2-27B+Shape        & 0.10 & 0.08 & 0.06 \\
Gemma-2-27B+PSI         & 0.05 & 0.05 & 0.03 \\
Llama3-8B+Persona  & 0.09 & 0.12 & 0.35 \\
Llama3-8B+Shape    & 0.04 & 0.07 & 0.07 \\
Llama3-8B+PSI     & 0.04 & 0.07 & 0.05 \\
Llama3-70B+Persona & 0.06 & 0.03 & 0.03 \\
Llama3-70B+Shape   & 0.09 & 0.07 & 0.06 \\
Llama3-70B+PSI    & 0.04 & 0.05 & 0.03 \\
GPT-4o-mini+Persona          & 0.06 & 0.04 & 0.04 \\
GPT-4o-mini+Shape            & 0.09 & 0.06 & 0.08 \\
GPT-4o-mini+PSI             & 0.07 & 0.05 & 0.03 \\
GPT-4o+Persona               & 0.06 & 0.02 & 0.05 \\
GPT-4o+Shape                 & 0.10 & 0.07 & 0.07 \\
GPT-4o+PSI                  & 0.07 & 0.05 & 0.04 \\
\end{longtable}

\begin{longtable}{l c c c}
\caption{Factor loading MAEs for BFI-2 TFM of Conscientiousness.} 
\label{tab:mae_conscientiousness} \\
\toprule
\textbf{Model} & \textbf{Organization} & \textbf{Productiveness} & \textbf{Responsibility} \\
\midrule
\endfirsthead
\multicolumn{4}{@{}l}{\tablename\ \thetable{} -- continued} \\ \toprule
\textbf{Model} & \textbf{Organization} & \textbf{Productiveness} & \textbf{Responsibility} \\ \midrule
\endhead
\midrule\multicolumn{4}{r}{{Continued on next page}} \\ \endfoot
\bottomrule\endlastfoot
Mistral-7B+Persona & 0.08 & 0.04 & 0.08 \\
Mistral-7B+Shape   & 0.08 & 0.04 & 0.06 \\
Mistral-7B+PSI    & 0.07 & 0.02 & 0.06 \\
Gemma-2-9B+Persona       & 0.06 & 0.06 & 0.05 \\
Gemma-2-9B+Shape         & 0.03 & 0.06 & 0.04 \\
Gemma-2-9B+PSI          & 0.15 & 0.04 & 0.08 \\
Gemma-2-27B+Persona      & 0.03 & 0.03 & 0.05 \\
Gemma-2-27B+Shape        & 0.05 & 0.06 & 0.06 \\
Gemma-2-27B+PSI         & 0.06 & 0.03 & 0.06 \\
Llama3-8B+Persona  & 0.35 & 0.19 & 0.19 \\
Llama3-8B+Shape    & 0.13 & 0.11 & 0.11 \\
Llama3-8B+PSI     & 0.11 & 0.03 & 0.06 \\
Llama3-70B+Persona & 0.03 & 0.04 & 0.06 \\
Llama3-70B+Shape   & 0.07 & 0.04 & 0.07 \\
Llama3-70B+PSI    & 0.02 & 0.04 & 0.04 \\
GPT-4o-mini+Persona          & 0.02 & 0.03 & 0.04 \\
GPT-4o-mini+Shape            & 0.09 & 0.04 & 0.06 \\
GPT-4o-mini+PSI             & 0.02 & 0.03 & 0.03 \\
GPT-4o+Persona               & 0.05 & 0.02 & 0.03 \\
GPT-4o+Shape                 & 0.03 & 0.04 & 0.06 \\
GPT-4o+PSI                  & 0.02 & 0.02 & 0.04 \\
\end{longtable}

\begin{longtable}{l c c c}
\caption{Factor loading MAEs for BFI-2 TFM of Neuroticism.} 
\label{tab:mae_neuroticism} \\
\toprule
\textbf{Model} & \textbf{Anxiety} & \textbf{Depression} & \textbf{Emotional Volatility} \\
\midrule
\endfirsthead
\multicolumn{4}{@{}l}{\tablename\ \thetable{} -- continued} \\ \toprule
\textbf{Model} & \textbf{Anxiety} & \textbf{Depression} & \textbf{Emotional Volatility} \\ \midrule
\endhead
\midrule\multicolumn{4}{r}{{Continued on next page}} \\ \endfoot
\bottomrule\endlastfoot
Mistral-7B+Persona & 0.10 & 0.08 & 0.12 \\
Mistral-7B+Shape   & 0.11 & 0.01 & 0.09 \\
Mistral-7B+PSI    & 0.04 & 0.03 & 0.02 \\
Gemma-2-9B+Persona       & 0.02 & 0.01 & 0.04 \\
Gemma-2-9B+Shape         & 0.08 & 0.03 & 0.09 \\
Gemma-2-9B+PSI          & 0.02 & 0.01 & 0.05 \\
Gemma-2-27B+Persona      & 0.06 & 0.01 & 0.02 \\
Gemma-2-27B+Shape        & 0.10 & 0.05 & 0.06 \\
Gemma-2-27B+PSI         & 0.05 & 0.03 & 0.03 \\
Llama3-8B+Persona  & 0.23 & 0.23 & 0.34 \\
Llama3-8B+Shape    & 0.28 & 0.17 & 0.23 \\
Llama3-8B+PSI     & 0.02 & 0.02 & 0.04 \\
Llama3-70B+Persona & 0.03 & 0.03 & 0.01 \\
Llama3-70B+Shape   & 0.03 & 0.03 & 0.03 \\
Llama3-70B+PSI    & 0.02 & 0.02 & 0.02 \\
GPT-4o-mini+Persona          & 0.03 & 0.04 & 0.03 \\
GPT-4o-mini+Shape            & 0.11 & 0.02 & 0.07 \\
GPT-4o-mini+PSI             & 0.02 & 0.02 & 0.01 \\
GPT-4o+Persona               & 0.06 & 0.07 & 0.09 \\
GPT-4o+Shape                 & 0.04 & 0.03 & 0.04 \\
GPT-4o+PSI                  & 0.02 & 0.02 & 0.01 \\
\end{longtable}

\begin{longtable}{l c c c}
\caption{Factor loading MAEs for BFI-2 TFM of Openness.} 
\label{tab:mae_openness} \\
\toprule
\textbf{Model} & \textbf{Intellectual Curiosity} & \textbf{Aesthetic Sensitivity} & \textbf{Creative Imagination} \\
\midrule
\endfirsthead
\multicolumn{4}{@{}l}{\tablename\ \thetable{} -- continued} \\ \toprule
\textbf{Model} & \textbf{Intellectual Curiosity} & \textbf{Aesthetic Sensitivity} & \textbf{Creative Imagination} \\ \midrule
\endhead
\midrule\multicolumn{4}{r}{{Continued on next page}} \\ \endfoot
\bottomrule\endlastfoot
Mistral-7B+Persona & 0.02 & 0.07 & 0.07 \\
Mistral-7B+Shape   & 0.05 & 0.02 & 0.05 \\
Mistral-7B+PSI    & 0.10 & 0.07 & 0.11 \\
Gemma-2-9B+Persona       & 0.06 & 0.06 & 0.09 \\
Gemma-2-9B+Shape         & 0.06 & 0.04 & 0.04 \\
Gemma-2-9B+PSI          & 0.05 & 0.36 & 0.13 \\
Gemma-2-27B+Persona      & 0.03 & 0.04 & 0.02 \\
Gemma-2-27B+Shape        & 0.09 & 0.04 & 0.06 \\
Gemma-2-27B+PSI         & 0.04 & 0.05 & 0.03 \\
Llama3-8B+Persona  & 0.15 & 0.37 & 0.19 \\
Llama3-8B+Shape    & 0.14 & 0.07 & 0.14 \\
Llama3-8B+PSI     & 0.03 & 0.04 & 0.04 \\
Llama3-70B+Persona & 0.04 & 0.04 & 0.05 \\
Llama3-70B+Shape   & 0.07 & 0.05 & 0.02 \\
Llama3-70B+PSI    & 0.04 & 0.05 & 0.06 \\
GPT-4o-mini+Persona          & 0.02 & 0.01 & 0.03 \\
GPT-4o-mini+Shape            & 0.07 & 0.09 & 0.10 \\
GPT-4o-mini+PSI             & 0.05 & 0.03 & 0.02 \\
GPT-4o+Persona               & 0.02 & 0.02 & 0.02 \\
GPT-4o+Shape                 & 0.05 & 0.04 & 0.02 \\
GPT-4o+PSI                  & 0.06 & 0.02 & 0.03 \\
\end{longtable}

\begin{table*}[h]
\caption{Factor loading MAEs for BFI-2 FFM.}
\label{tab:mae_ffm}
\centering
\begin{tabular}{l*{5}{c}}
\toprule
\textbf{Model}
 & \textbf{Extraversion}  & \textbf{Agreeableness} & \textbf{Conscientiousness} & \textbf{Neuroticism} & \textbf{Openness} \\
\midrule
Mistral-7B+Persona & 0.02 & 0.01 & 0.01 & 0.03 & 0.01 \\
Mistral-7B+Shape   & 0.04 & 0.03 & 0.03 & 0.03 & 0.03 \\
Mistral-7B+PSI     & 0.03 & 0.03 & 0.01 & 0.02 & 0.01 \\
Gemma-2-9B+Persona       & 0.02 & 0.03 & 0.03 & 0.02 & 0.02 \\
Gemma-2-9B+Shape         & 0.05 & 0.03 & 0.02 & 0.03 & 0.03 \\
Gemma-2-9B+PSI           & 0.02 & 0.02 & 0.02 & 0.02 & 0.04 \\
Gemma-2-27B+Persona      & 0.03 & 0.04 & 0.03 & 0.01 & 0.02 \\
Gemma-2-27B+Shape        & 0.04 & 0.03 & 0.02 & 0.03 & 0.04 \\
Gemma-2-27B+PSI          & 0.03 & 0.02 & 0.02 & 0.01 & 0.02 \\
Llama3-8B+Persona  & 0.02 & 0.03 & 0.05 & 0.03 & 0.03 \\
Llama3-8B+Shape    & 0.03 & 0.04 & 0.03 & 0.04 & 0.02 \\
Llama3-8B+PSI      & 0.03 & 0.02 & 0.02 & 0.01 & 0.02 \\
Llama3-70B+Persona & 0.04 & 0.04 & 0.03 & 0.02 & 0.02 \\
Llama3-70B+Shape   & 0.03 & 0.04 & 0.03 & 0.01 & 0.03 \\
Llama3-70B+PSI     & 0.03 & 0.03 & 0.02 & 0.01 & 0.02 \\
GPT-4o-mini+Persona         & 0.03 & 0.04 & 0.03 & 0.02 & 0.01 \\
GPT-4o-mini+Shape           & 0.02 & 0.04 & 0.03 & 0.02 & 0.04 \\
GPT-4o-mini+PSI             & 0.03 & 0.03 & 0.02 & 0.02 & 0.01 \\
GPT-4o+Persona               & 0.02 & 0.02 & 0.02 & 0.03 & 0.01 \\
GPT-4o+Shape                 & 0.04 & 0.03 & 0.04 & 0.03 & 0.02 \\
GPT-4o+PSI                   & 0.03 & 0.02 & 0.02 & 0.02 & 0.01 \\
\bottomrule
\end{tabular}
\end{table*}

\begin{longtable}{l c c c c}
\caption{Standardized factor loadings for Sociability items in BFI-2 TFM of Extraversion.}%
\label{tab:fc_sociability} \\
\toprule
\textbf{Model} 
 & \textbf{Item 1} & \textbf{Item 16} & \textbf{Item 31} & \textbf{Item 46} \\
\midrule
\endfirsthead
\multicolumn{5}{@{}l}{\tablename\ \thetable{} -- continued} \\ \toprule
\textbf{Model} 
 & \textbf{Item 1} & \textbf{Item 16} & \textbf{Item 31} & \textbf{Item 46} \\ \midrule
\endhead
\midrule\multicolumn{5}{r}{{Continued on next page}} \\ \endfoot
\bottomrule\endlastfoot
Mistral-7B+Persona  & 0.88 & 0.46 & 0.22 & 0.77 \\
Mistral-7B+Shape    & 0.64 & 0.87 & 0.65 & 0.78 \\
Mistral-7B+PSI     & 0.77 & 0.73 & 0.72 & 0.81 \\
Gemma-2-9B+Persona        & 0.75 & 0.81 & 0.79 & 0.71 \\
Gemma-2-9B+Shape          & 0.71 & 0.90 & 0.80 & 0.41 \\
Gemma-2-9B+PSI           & 0.79 & 0.83 & 0.71 & 0.77 \\
Gemma-2-27B+Persona       & 0.75 & 0.80 & 0.70 & 0.76 \\
Gemma-2-27B+Shape         & 0.84 & 0.87 & 0.77 & 0.80 \\
Gemma-2-27B+PSI          & 0.70 & 0.91 & 0.84 & 0.80 \\
Llama3-8B+Persona   & -0.15 & -0.78 & -0.75 & 0.01 \\
Llama3-8B+Shape     & -0.71 & -0.18 & -0.08 & -0.67 \\
Llama3-8B+PSI      & 0.82 & 0.71 & 0.77 & 0.78 \\
Llama3-70B+Persona  & 0.72 & 0.88 & 0.87 & 0.64 \\
Llama3-70B+Shape    & 0.55 & 0.73 & 0.79 & 0.58 \\
Llama3-70B+PSI     & 0.65 & 0.86 & 0.74 & 0.69 \\
GPT-4o-mini+Persona           & 0.79 & 0.78 & 0.72 & 0.74 \\
GPT-4o-mini+Shape             & 0.88 & 0.33 & 0.37 & 0.86 \\
GPT-4o-mini+PSI              & 0.85 & 0.88 & 0.88 & 0.82 \\
GPT-4o+Persona                & 0.77 & 0.79 & 0.79 & 0.74 \\
GPT-4o+Shape                  & 0.85 & 0.74 & 0.73 & 0.85 \\
GPT-4o+PSI                   & 0.87 & 0.84 & 0.90 & 0.76 \\
Human                 & 0.80 & 0.82 & 0.82 & 0.76 \\
\end{longtable}

\begin{longtable}{l c c c c}
\caption{Standardized factor loadings for Assertiveness items in BFI-2 TFM of Extraversion.}%
\label{tab:fc_assertiveness} \\
\toprule
\textbf{Model} 
 & \textbf{Item 6} & \textbf{Item 21} & \textbf{Item 36} & \textbf{Item 51} \\
\midrule
\endfirsthead
\multicolumn{5}{@{}l}{\tablename\ \thetable{} -- continued} \\ \toprule
\textbf{Model} 
 & \textbf{Item 6} & \textbf{Item 21} & \textbf{Item 36} & \textbf{Item 51} \\ \midrule
\endhead
\midrule\multicolumn{5}{r}{{Continued on next page}} \\ \endfoot
\bottomrule\endlastfoot
Mistral-7B+Persona  & 0.71 & 0.33 & 0.59 & 0.35 \\
Mistral-7B+Shape    & 0.76 & 0.54 & 0.86 & 0.80 \\
Mistral-7B+PSI     & 0.85 & 0.63 & 0.34 & 0.35 \\
Gemma-2-9B+Persona        & 0.90 & 0.84 & 0.33 & 0.44 \\
Gemma-2-9B+Shape          & 0.70 & 0.20 & 0.81 & 0.81 \\
Gemma-2-9B+PSI           & 0.71 & 0.86 & 0.37 & 0.35 \\
Gemma-2-27B+Persona       & 0.83 & 0.67 & 0.76 & 0.52 \\
Gemma-2-27B+Shape         & 0.99 & 0.84 & 0.80 & 0.52 \\
Gemma-2-27B+PSI          & 0.58 & 0.41 & 0.74 & 0.45 \\
Llama3-8B+Persona   & 0.78 & 0.74 & -0.32 & -0.23 \\
Llama3-8B+Shape     & 0.79 & 0.83 & -0.28 & 0.31 \\
Llama3-8B+PSI      & 0.80 & 0.78 & 0.50 & 0.43 \\
Llama3-70B+Persona  & 0.84 & 0.73 & 0.72 & 0.54 \\
Llama3-70B+Shape    & 0.86 & 0.70 & 0.88 & 0.72 \\
Llama3-70B+PSI     & 0.49 & 0.53 & 0.59 & 0.28 \\
GPT-4o-mini+Persona           & 0.87 & 0.80 & 0.74 & 0.65 \\
GPT-4o-mini+Shape             & 0.99 & 0.86 & 0.28 & 0.37 \\
GPT-4o-mini+PSI              & 0.78 & 0.76 & 0.72 & 0.63 \\
GPT-4o+Persona                & 0.66 & 0.61 & 0.73 & 0.59 \\
GPT-4o+Shape                  & 0.94 & 0.90 & 0.84 & 0.80 \\
GPT-4o+PSI                   & 0.76 & 0.79 & 0.78 & 0.56 \\
Human                 & 0.76 & 0.89 & 0.53 & 0.69 \\
\end{longtable}

\begin{longtable}{l c c c c}
\caption{Standardized factor loadings for Energy Level items in BFI-2 TFM of Extraversion.}%
\label{tab:fc_energy} \\
\toprule
\textbf{Model} 
 & \textbf{Item 11} & \textbf{Item 26} & \textbf{Item 41} & \textbf{Item 56} \\
\midrule
\endfirsthead
\multicolumn{5}{@{}l}{\tablename\ \thetable{} -- continued} \\ \toprule
\textbf{Model} 
 & \textbf{Item 11} & \textbf{Item 26} & \textbf{Item 41} & \textbf{Item 56} \\ \midrule
\endhead
\midrule\multicolumn{5}{r}{{Continued on next page}} \\ \endfoot
\bottomrule\endlastfoot
Mistral-7B+Persona  & 0.60 & 0.74 & 0.79 & 0.64 \\
Mistral-7B+Shape    & 0.87 & 0.88 & 0.73 & 0.73 \\
Mistral-7B+PSI     & 0.31 & 0.54 & 0.87 & 0.79 \\
Gemma-2-9B+Persona        & 0.47 & 0.64 & 0.88 & 0.77 \\
Gemma-2-9B+Shape          & 0.84 & 0.92 & 0.84 & 0.86 \\
Gemma-2-9B+PSI           & 0.06 & -0.00 & 0.88 & 0.82 \\
Gemma-2-27B+Persona       & 0.72 & 0.83 & 0.86 & 0.79 \\
Gemma-2-27B+Shape         & 0.91 & 0.84 & 0.90 & 0.90 \\
Gemma-2-27B+PSI          & 0.80 & 0.63 & 0.85 & 0.75 \\
Llama3-8B+Persona   & -0.13 & -0.14 & 0.57 & 0.65 \\
Llama3-8B+Shape     & -0.04 & 0.14 & -0.69 & -0.76 \\
Llama3-8B+PSI      & 0.77 & 0.62 & 0.83 & 0.80 \\
Llama3-70B+Persona  & 0.73 & 0.75 & 0.87 & 0.78 \\
Llama3-70B+Shape    & 0.88 & 0.77 & 0.76 & 0.77 \\
Llama3-70B+PSI     & 0.64 & 0.50 & 0.72 & 0.84 \\
GPT-4o-mini+Persona           & 0.70 & 0.78 & 0.88 & 0.88 \\
GPT-4o-mini+Shape             & 0.55 & 0.48 & 0.89 & 0.87 \\
GPT-4o-mini+PSI              & 0.76 & 0.75 & 0.89 & 0.82 \\
GPT-4o+Persona                & 0.82 & 0.69 & 0.81 & 0.82 \\
GPT-4o+Shape                  & 0.91 & 0.72 & 0.82 & 0.88 \\
GPT-4o+PSI                   & 0.82 & 0.75 & 0.88 & 0.82 \\
Human                 & 0.45 & 0.54 & 0.83 & 0.77 \\
\end{longtable}

\begin{longtable}{l c c c c}
\caption{Standardized factor loadings for Compassion items in BFI-2 TFM of Agreeableness.}%
\label{tab:fc_compassion} \\
\toprule
\textbf{Model} 
 & \textbf{Item 2} & \textbf{Item 17} & \textbf{Item 32} & \textbf{Item 47} \\
\midrule
\endfirsthead
\multicolumn{5}{@{}l}{\tablename\ \thetable{} -- continued} \\ \toprule
\textbf{Model} 
 & \textbf{Item 2} & \textbf{Item 17} & \textbf{Item 32} & \textbf{Item 47} \\ \midrule
\endhead
\midrule\multicolumn{5}{r}{{Continued on next page}} \\ \endfoot
\bottomrule\endlastfoot
Mistral-7B+Persona & 0.68 & 0.11 & 0.76 & 0.48 \\
Mistral-7B+Shape   & 0.46 & 0.82 & 0.69 & 0.89 \\
Mistral-7B+PSI    & 0.78 & 0.44 & 0.86 & 0.73 \\
Gemma-2-9B+Persona       & 0.80 & 0.76 & 0.80 & 0.77 \\
Gemma-2-9B+Shape         & 0.86 & 0.92 & 0.93 & 0.96 \\
Gemma-2-9B+PSI          & 0.83 & 0.55 & 0.87 & 0.80 \\
Gemma-2-27B+Persona      & 0.70 & 0.68 & 0.86 & 0.79 \\
Gemma-2-27B+Shape        & 0.85 & 0.90 & 0.93 & 0.96 \\
Gemma-2-27B+PSI         & 0.74 & 0.81 & 0.73 & 0.81 \\
Llama3-8B+Persona  & 0.71 & 0.04 & 0.88 & 0.23 \\
Llama3-8B+Shape    & 0.77 & 0.25 & 0.71 & 0.43 \\
Llama3-8B+PSI     & 0.80 & 0.46 & 0.75 & 0.62 \\
Llama3-70B+Persona & 0.75 & 0.86 & 0.74 & 0.85 \\
Llama3-70B+Shape   & 0.74 & 0.97 & 0.86 & 0.98 \\
Llama3-70B+PSI    & 0.84 & 0.48 & 0.79 & 0.86 \\
GPT-4o-mini+Persona          & 0.82 & 0.70 & 0.84 & 0.76 \\
GPT-4o-mini+Shape            & 0.52 & 0.96 & 0.69 & 0.98 \\
GPT-4o-mini+PSI             & 0.84 & 0.79 & 0.81 & 0.89 \\
GPT-4o+Persona               & 0.76 & 0.84 & 0.74 & 0.67 \\
GPT-4o+Shape                 & 0.90 & 0.93 & 0.87 & 0.96 \\
GPT-4o+PSI                  & 0.83 & 0.77 & 0.79 & 0.89 \\
Human                & 0.73 & 0.34 & 0.62 & 0.75 \\
\end{longtable}

\begin{longtable}{l c c c c}
\caption{Standardized factor loadings for Respectfulness items in BFI-2 TFM of Agreeableness.}%
\label{tab:fc_respectfulness} \\
\toprule
\textbf{Model} 
 & \textbf{Item 7} & \textbf{Item 22} & \textbf{Item 37} & \textbf{Item 52} \\
\midrule
\endfirsthead
\multicolumn{5}{@{}l}{\tablename\ \thetable{} -- continued} \\ \toprule
\textbf{Model} 
 & \textbf{Item 7} & \textbf{Item 22} & \textbf{Item 37} & \textbf{Item 52} \\ \midrule
\endhead
\midrule\multicolumn{5}{r}{{Continued on next page}} \\ \endfoot
\bottomrule\endlastfoot
Mistral-7B+Persona & 0.77 & 0.53 & 0.54 & 0.88 \\
Mistral-7B+Shape   & 0.60 & 0.77 & 0.88 & 0.48 \\
Mistral-7B+PSI    & 0.92 & 0.36 & 0.70 & 0.93 \\
Gemma-2-9B+Persona       & 0.77 & 0.60 & 0.63 & 0.86 \\
Gemma-2-9B+Shape         & 0.98 & 0.83 & 0.91 & 0.97 \\
Gemma-2-9B+PSI          & 0.90 & 0.10 & 0.50 & 0.89 \\
Gemma-2-27B+Persona      & 0.82 & 0.50 & 0.64 & 0.87 \\
Gemma-2-27B+Shape        & 0.96 & 0.83 & 0.87 & 0.99 \\
Gemma-2-27B+PSI         & 0.93 & 0.42 & 0.70 & 0.89 \\
Llama3-8B+Persona  & 0.73 & 0.12 & -0.13 & 0.77 \\
Llama3-8B+Shape    & 0.94 & 0.50 & 0.28 & 0.75 \\
Llama3-8B+PSI     & 0.90 & 0.26 & 0.51 & 0.82 \\
Llama3-70B+Persona & 0.80 & 0.68 & 0.79 & 0.80 \\
Llama3-70B+Shape   & 0.85 & 0.87 & 0.94 & 0.84 \\
Llama3-70B+PSI    & 0.92 & 0.48 & 0.84 & 0.90 \\
GPT-4o-mini+Persona          & 0.86 & 0.45 & 0.67 & 0.85 \\
GPT-4o-mini+Shape            & 0.64 & 0.84 & 0.97 & 0.62 \\
GPT-4o-mini+PSI             & 0.85 & 0.35 & 0.76 & 0.89 \\
GPT-4o+Persona               & 0.82 & 0.54 & 0.74 & 0.70 \\
GPT-4o+Shape                 & 0.93 & 0.80 & 0.94 & 0.88 \\
GPT-4o+PSI                  & 0.91 & 0.44 & 0.78 & 0.87 \\
Human                & 0.71 & 0.60 & 0.70 & 0.70 \\
\end{longtable}

\begin{longtable}{l c c c c}
\caption{Standardized factor loadings for Trust items in BFI-2 TFM of Agreeableness.}%
\label{tab:fc_trust} \\
\toprule
\textbf{Model} 
 & \textbf{Item 12} & \textbf{Item 27} & \textbf{Item 42} & \textbf{Item 57} \\
\midrule
\endfirsthead
\multicolumn{5}{@{}l}{\tablename\ \thetable{} -- continued} \\ \toprule
\textbf{Model} 
 & \textbf{Item 12} & \textbf{Item 27} & \textbf{Item 42} & \textbf{Item 57} \\ \midrule
\endhead
\midrule\multicolumn{5}{r}{{Continued on next page}} \\ \endfoot
\bottomrule\endlastfoot
Mistral-7B+Persona & 0.34 & 0.75 & 0.21 & 0.78 \\
Mistral-7B+Shape   & 0.94 & 0.50 & 0.90 & 0.58 \\
Mistral-7B+PSI    & 0.40 & 0.78 & 0.67 & 0.84 \\
Gemma-2-9B+Persona       & 0.35 & 0.62 & 0.56 & 0.88 \\
Gemma-2-9B+Shape         & 0.95 & 0.87 & 0.95 & 0.90 \\
Gemma-2-9B+PSI          & -0.01 & -0.81 & -0.51 & -0.85 \\
Gemma-2-27B+Persona      & 0.66 & 0.70 & 0.68 & 0.88 \\
Gemma-2-27B+Shape        & 0.96 & 0.80 & 0.96 & 0.86 \\
Gemma-2-27B+PSI         & 0.80 & 0.71 & 0.81 & 0.88 \\
Llama3-8B+Persona  & 0.18 & -0.79 & 0.12 & -0.83 \\
Llama3-8B+Shape    & 0.46 & 0.80 & 0.27 & 0.85 \\
Llama3-8B+PSI     & 0.38 & 0.79 & 0.56 & 0.86 \\
Llama3-70B+Persona & 0.57 & 0.80 & 0.63 & 0.82 \\
Llama3-70B+Shape   & 0.94 & 0.84 & 0.95 & 0.85 \\
Llama3-70B+PSI    & 0.66 & 0.85 & 0.80 & 0.87 \\
GPT-4o-mini+Persona          & 0.49 & 0.85 & 0.66 & 0.90 \\
GPT-4o-mini+Shape            & 0.94 & 0.61 & 0.94 & 0.44 \\
GPT-4o-mini+PSI             & 0.69 & 0.80 & 0.81 & 0.91 \\
GPT-4o+Persona               & 0.63 & 0.60 & 0.35 & 0.64 \\
GPT-4o+Shape                 & 0.96 & 0.84 & 0.95 & 0.90 \\
GPT-4o+PSI                  & 0.76 & 0.80 & 0.80 & 0.89 \\
Human                & 0.69 & 0.68 & 0.67 & 0.78 \\
\end{longtable}

\begin{longtable}{l c c c c}
\caption{Standardized factor loadings for Organization items in BFI-2 TFM of Conscientiousness.}%
\label{tab:fc_organization} \\
\toprule
\textbf{Model} 
 & \textbf{Item 3} & \textbf{Item 18} & \textbf{Item 33} & \textbf{Item 48} \\
\midrule
\endfirsthead
\multicolumn{5}{@{}l}{\tablename\ \thetable{} -- continued} \\ \toprule
\textbf{Model} 
 & \textbf{Item 3} & \textbf{Item 18} & \textbf{Item 33} & \textbf{Item 48} \\ \midrule
\endhead
\midrule\multicolumn{5}{r}{{Continued on next page}} \\ \endfoot
\bottomrule\endlastfoot
Mistral-7B+Persona  & 0.47 & 0.59 & 0.67 & 0.41 \\
Mistral-7B+Shape    & 0.81 & 0.41 & 0.36 & 0.87 \\
Mistral-7B+PSI     & 0.73 & 0.58 & 0.42 & 0.60 \\
Gemma-2-9B+Persona        & 0.71 & 0.83 & 0.69 & 0.50 \\
Gemma-2-9B+Shape          & 0.77 & 0.65 & 0.66 & 0.78 \\
Gemma-2-9B+PSI           & 0.25 & 0.82 & 0.39 & 0.13 \\
Gemma-2-27B+Persona       & 0.63 & 0.70 & 0.73 & 0.69 \\
Gemma-2-27B+Shape         & 0.93 & 0.54 & 0.64 & 0.88 \\
Gemma-2-27B+PSI          & 0.80 & 0.60 & 0.63 & 0.42 \\
Llama3-8B+Persona   & 0.41 & -0.72 & -0.78 & 0.06 \\
Llama3-8B+Shape     & 0.11 & 0.89 & 0.81 & 0.22 \\
Llama3-8B+PSI      & 0.33 & 0.89 & 0.73 & 0.26 \\
Llama3-70B+Persona  & 0.75 & 0.66 & 0.68 & 0.74 \\
Llama3-70B+Shape    & 0.49 & 0.90 & 0.95 & 0.56 \\
Llama3-70B+PSI     & 0.83 & 0.65 & 0.75 & 0.76 \\
GPT-4o-mini+Persona           & 0.80 & 0.72 & 0.76 & 0.76 \\
GPT-4o-mini+Shape             & 0.85 & 0.27 & 0.32 & 0.85 \\
GPT-4o-mini+PSI              & 0.84 & 0.79 & 0.75 & 0.73 \\
GPT-4o+Persona                & 0.64 & 0.62 & 0.63 & 0.70 \\
GPT-4o+Shape                  & 0.75 & 0.73 & 0.73 & 0.88 \\
GPT-4o+PSI                   & 0.84 & 0.84 & 0.75 & 0.72 \\
Human                 & 0.85 & 0.70 & 0.86 & 0.74 \\
\end{longtable}

\begin{longtable}{l c c c c}
\caption{Standardized factor loadings for Productiveness items in BFI-2 TFM of Conscientiousness.}%
\label{tab:fc_productiveness} \\
\toprule
\textbf{Model} 
 & \textbf{Item 8} & \textbf{Item 23} & \textbf{Item 38} & \textbf{Item 53} \\
\midrule
\endfirsthead
\multicolumn{5}{@{}l}{\tablename\ \thetable{} -- continued} \\ \toprule
\textbf{Model} 
 & \textbf{Item 8} & \textbf{Item 23} & \textbf{Item 38} & \textbf{Item 53} \\ \midrule
\endhead
\midrule\multicolumn{5}{r}{{Continued on next page}} \\ \endfoot
\bottomrule\endlastfoot
Mistral-7B+Persona  & 0.54 & 0.59 & 0.82 & 0.71 \\
Mistral-7B+Shape    & 0.96 & 0.89 & 0.60 & 0.73 \\
Mistral-7B+PSI     & 0.74 & 0.60 & 0.70 & 0.74 \\
Gemma-2-9B+Persona        & 0.68 & 0.36 & 0.85 & 0.86 \\
Gemma-2-9B+Shape          & 0.84 & 0.95 & 0.90 & 0.91 \\
Gemma-2-9B+PSI           & 0.47 & 0.57 & 0.80 & 0.72 \\
Gemma-2-27B+Persona       & 0.70 & 0.67 & 0.86 & 0.81 \\
Gemma-2-27B+Shape         & 0.82 & 0.92 & 0.91 & 0.92 \\
Gemma-2-27B+PSI          & 0.64 & 0.59 & 0.75 & 0.60 \\
Llama3-8B+Persona   & -0.13 & -0.42 & 0.71 & 0.52 \\
Llama3-8B+Shape     & 0.16 & 0.13 & 0.78 & 0.73 \\
Llama3-8B+PSI      & 0.65 & 0.62 & 0.79 & 0.73 \\
Llama3-70B+Persona  & 0.68 & 0.65 & 0.88 & 0.82 \\
Llama3-70B+Shape    & 0.67 & 0.72 & 0.83 & 0.95 \\
Llama3-70B+PSI     & 0.79 & 0.43 & 0.74 & 0.78 \\
GPT-4o-mini+Persona           & 0.75 & 0.67 & 0.89 & 0.84 \\
GPT-4o-mini+Shape             & 0.89 & 0.81 & 0.48 & 0.64 \\
GPT-4o-mini+PSI              & 0.82 & 0.68 & 0.86 & 0.75 \\
GPT-4o+Persona                & 0.72 & 0.60 & 0.71 & 0.81 \\
GPT-4o+Shape                  & 0.80 & 0.73 & 0.86 & 0.95 \\
GPT-4o+PSI                   & 0.79 & 0.79 & 0.87 & 0.71 \\
Human                 & 0.72 & 0.75 & 0.73 & 0.70 \\
\end{longtable}

\begin{longtable}{l c c c c}
\caption{Standardized factor loadings for Responsibility items in BFI-2 TFM of Conscientiousness.}%
\label{tab:fc_responsibility} \\
\toprule
\textbf{Model} 
 & \textbf{Item 13} & \textbf{Item 28} & \textbf{Item 43} & \textbf{Item 58} \\
\midrule
\endfirsthead
\multicolumn{5}{@{}l}{\tablename\ \thetable{} -- continued} \\ \toprule
\textbf{Model} 
 & \textbf{Item 13} & \textbf{Item 28} & \textbf{Item 43} & \textbf{Item 58} \\ \midrule
\endhead
\midrule\multicolumn{5}{r}{{Continued on next page}} \\ \endfoot
\bottomrule\endlastfoot
Mistral-7B+Persona  & 0.93 & 0.46 & 0.97 & 0.51 \\
Mistral-7B+Shape    & 0.50 & 0.90 & 0.60 & 0.88 \\
Mistral-7B+PSI     & 0.93 & 0.60 & 0.93 & 0.61 \\
Gemma-2-9B+Persona        & 0.88 & 0.58 & 0.88 & 0.60 \\
Gemma-2-9B+Shape          & 0.83 & 0.72 & 0.77 & 0.92 \\
Gemma-2-9B+PSI           & 0.95 & 0.35 & 0.94 & 0.63 \\
Gemma-2-27B+Persona       & 0.91 & 0.66 & 0.93 & 0.68 \\
Gemma-2-27B+Shape         & 0.94 & 0.74 & 0.92 & 0.85 \\
Gemma-2-27B+PSI          & 0.92 & 0.51 & 0.93 & 0.65 \\
Llama3-8B+Persona   & 0.90 & -0.26 & 0.95 & -0.20 \\
Llama3-8B+Shape     & 0.91 & 0.16 & 0.90 & 0.31 \\
Llama3-8B+PSI      & 0.88 & 0.52 & 0.89 & 0.50 \\
Llama3-70B+Persona  & 0.94 & 0.58 & 0.95 & 0.66 \\
Llama3-70B+Shape    & 0.92 & 0.48 & 0.96 & 0.64 \\
Llama3-70B+PSI     & 0.81 & 0.74 & 0.82 & 0.84 \\
GPT-4o-mini+Persona           & 0.89 & 0.72 & 0.92 & 0.74 \\
GPT-4o-mini+Shape             & 0.46 & 0.86 & 0.52 & 0.85 \\
GPT-4o-mini+PSI              & 0.81 & 0.77 & 0.79 & 0.78 \\
GPT-4o+Persona                & 0.66 & 0.58 & 0.64 & 0.47 \\
GPT-4o+Shape                  & 0.92 & 0.68 & 0.94 & 0.80 \\
GPT-4o+PSI                   & 0.83 & 0.78 & 0.84 & 0.78 \\
Human                 & 0.68 & 0.73 & 0.70 & 0.65 \\
\end{longtable}

\begin{longtable}{l c c c c}
\caption{Standardized factor loadings for Anxiety items in BFI-2 TFM of Neuroticism.}%
\label{tab:fc_anxiety} \\
\toprule
\textbf{Model}
 & \textbf{Item 4} & \textbf{Item 19} & \textbf{Item 34} & \textbf{Item 49} \\
\midrule
\endfirsthead
\multicolumn{5}{@{}l}{\tablename\ \thetable{} -- continued} \\ \toprule
\textbf{Model}
 & \textbf{Item 4} & \textbf{Item 19} & \textbf{Item 34} & \textbf{Item 49} \\ \midrule
\endhead
\midrule\multicolumn{5}{r}{{Continued on next page}} \\ \endfoot
\bottomrule\endlastfoot
Mistral-7B+Persona  & 0.25 & 0.55 & 0.71 & 0.25 \\
Mistral-7B+Shape    & 0.35 & 0.96 & 0.79 & 0.14 \\
Mistral-7B+PSI     & 0.69 & 0.59 & 0.63 & 0.75 \\
Gemma-2-9B+Persona        & 0.76 & 0.72 & 0.82 & 0.80 \\
Gemma-2-9B+Shape          & 0.58 & 0.83 & 0.88 & 0.18 \\
Gemma-2-9B+PSI           & 0.69 & 0.71 & 0.74 & 0.73 \\
Gemma-2-27B+Persona       & 0.75 & 0.85 & 0.69 & 0.35 \\
Gemma-2-27B+Shape         & 0.66 & 0.91 & 0.89 & -0.08 \\
Gemma-2-27B+PSI          & 0.68 & 0.78 & 0.63 & 0.46 \\
Llama3-8B+Persona   & 0.56 & -0.48 & -0.31 & 0.53 \\
Llama3-8B+Shape     & 0.51 & -0.72 & -0.71 & 0.52 \\
Llama3-8B+PSI      & 0.71 & 0.78 & 0.77 & 0.64 \\
Llama3-70B+Persona  & 0.72 & 0.61 & 0.81 & 0.74 \\
Llama3-70B+Shape    & 0.63 & 0.72 & 0.80 & 0.71 \\
Llama3-70B+PSI     & 0.79 & 0.64 & 0.76 & 0.59 \\
GPT-4o-mini+Persona           & 0.74 & 0.58 & 0.67 & 0.64 \\
GPT-4o-mini+Shape             & 0.27 & 0.79 & 0.92 & 0.15 \\
GPT-4o-mini+PSI              & 0.84 & 0.86 & 0.84 & 0.70 \\
GPT-4o+Persona                & 0.52 & 0.65 & 0.59 & 0.51 \\
GPT-4o+Shape                  & 0.55 & 0.80 & 0.78 & 0.75 \\
GPT-4o+PSI                   & 0.85 & 0.87 & 0.80 & 0.68 \\
Human                 & 0.85 & 0.70 & 0.79 & 0.66 \\
\end{longtable}

\begin{longtable}{l c c c c}
\caption{Standardized factor loadings for Depression items in BFI-2 TFM of Neuroticism.}%
\label{tab:fc_depression} \\
\toprule
\textbf{Model}
 & \textbf{Item 9} & \textbf{Item 24} & \textbf{Item 39} & \textbf{Item 54} \\
\midrule
\endfirsthead
\multicolumn{5}{@{}l}{\tablename\ \thetable{} -- continued} \\ \toprule
\textbf{Model}
 & \textbf{Item 9} & \textbf{Item 24} & \textbf{Item 39} & \textbf{Item 54} \\ \midrule
\endhead
\midrule\multicolumn{5}{r}{{Continued on next page}} \\ \endfoot
\bottomrule\endlastfoot
Mistral-7B+Persona  & 0.19 & 0.27 & 0.79 & 0.91 \\
Mistral-7B+Shape    & 0.68 & 0.69 & 0.83 & 0.92 \\
Mistral-7B+PSI     & 0.64 & 0.79 & 0.76 & 0.82 \\
Gemma-2-9B+Persona        & 0.52 & 0.66 & 0.88 & 0.90 \\
Gemma-2-9B+Shape          & 0.74 & 0.80 & 0.85 & 0.95 \\
Gemma-2-9B+PSI           & 0.57 & 0.63 & 0.89 & 0.95 \\
Gemma-2-27B+Persona       & 0.67 & 0.63 & 0.88 & 0.86 \\
Gemma-2-27B+Shape         & 0.88 & 0.91 & 0.93 & 0.98 \\
Gemma-2-27B+PSI          & 0.42 & 0.63 & 0.92 & 0.95 \\
Llama3-8B+Persona   & 0.57 & 0.58 & -0.50 & -0.28 \\
Llama3-8B+Shape     & 0.67 & 0.93 & -0.09 & 0.15 \\
Llama3-8B+PSI      & 0.56 & 0.67 & 0.84 & 0.86 \\
Llama3-70B+Persona  & 0.68 & 0.86 & 0.81 & 0.88 \\
Llama3-70B+Shape    & 0.78 & 0.86 & 0.87 & 0.95 \\
Llama3-70B+PSI     & 0.61 & 0.80 & 0.89 & 0.94 \\
GPT-4o-mini+Persona           & 0.71 & 0.82 & 0.76 & 0.80 \\
GPT-4o-mini+Shape             & 0.50 & 0.54 & 0.90 & 0.90 \\
GPT-4o-mini+PSI              & 0.63 & 0.81 & 0.92 & 0.95 \\
GPT-4o+Persona                & 0.27 & 0.72 & 0.70 & 0.67 \\
GPT-4o+Shape                  & 0.75 & 0.83 & 0.84 & 0.89 \\
GPT-4o+PSI                   & 0.69 & 0.83 & 0.93 & 0.95 \\
Human                 & 0.64 & 0.67 & 0.90 & 0.91 \\
\end{longtable}

\begin{longtable}{l c c c c}
\caption{Standardized factor loadings for Emotional Volatility items in BFI-2 TFM of Neuroticism.}%
\label{tab:fc_emotional_volatility} \\
\toprule
\textbf{Model}
 & \textbf{Item 14} & \textbf{Item 29} & \textbf{Item 44} & \textbf{Item 59} \\
\midrule
\endfirsthead
\multicolumn{5}{@{}l}{\tablename\ \thetable{} -- continued} \\ \toprule
\textbf{Model}
 & \textbf{Item 14} & \textbf{Item 29} & \textbf{Item 44} & \textbf{Item 59} \\ \midrule
\endhead
\midrule\multicolumn{5}{r}{{Continued on next page}} \\ \endfoot
\bottomrule\endlastfoot
Mistral-7B+Persona  & 0.64 & 0.30 & 0.23 & 0.90 \\
Mistral-7B+Shape    & 0.92 & 0.62 & 0.25 & 0.93 \\
Mistral-7B+PSI     & 0.77 & 0.87 & 0.63 & 0.69 \\
Gemma-2-9B+Persona        & 0.72 & 0.90 & 0.62 & 0.62 \\
Gemma-2-9B+Shape          & 0.96 & 0.57 & 0.20 & 0.89 \\
Gemma-2-9B+PSI           & 0.72 & 0.79 & 0.69 & 0.40 \\
Gemma-2-27B+Persona       & 0.76 & 0.84 & 0.74 & 0.63 \\
Gemma-2-27B+Shape         & 0.99 & 0.71 & 0.54 & 0.91 \\
Gemma-2-27B+PSI          & 0.88 & 0.81 & 0.71 & 0.63 \\
Llama3-8B+Persona   & 0.23 & -0.75 & -0.70 & 0.34 \\
Llama3-8B+Shape     & 0.90 & -0.41 & -0.44 & 0.90 \\
Llama3-8B+PSI      & 0.63 & 0.87 & 0.57 & 0.67 \\
Llama3-70B+Persona  & 0.86 & 0.86 & 0.73 & 0.80 \\
Llama3-70B+Shape    & 0.88 & 0.85 & 0.60 & 0.84 \\
Llama3-70B+PSI     & 0.86 & 0.91 & 0.83 & 0.74 \\
GPT-4o-mini+Persona           & 0.77 & 0.90 & 0.58 & 0.66 \\
GPT-4o-mini+Shape             & 0.85 & 0.57 & 0.38 & 0.74 \\
GPT-4o-mini+PSI              & 0.80 & 0.93 & 0.77 & 0.69 \\
GPT-4o+Persona                & 0.71 & 0.44 & 0.40 & 0.59 \\
GPT-4o+Shape                  & 0.92 & 0.80 & 0.58 & 0.75 \\
GPT-4o+PSI                   & 0.78 & 0.92 & 0.75 & 0.72 \\
Human                 & 0.80 & 0.88 & 0.77 & 0.80 \\
\end{longtable}

\begin{longtable}{l c c c c}
\caption{Standardized factor loadings for Intellectual Curiosity items in BFI-2 TFM of Openness.}%
\label{tab:fc_intellectual_curiosity} \\
\toprule
\textbf{Model} 
 & \textbf{Item 10} & \textbf{Item 25} & \textbf{Item 40} & \textbf{Item 55} \\
\midrule
\endfirsthead
\multicolumn{5}{@{}l}{\tablename\ \thetable{} -- continued} \\ \toprule
\textbf{Model} 
 & \textbf{Item 10} & \textbf{Item 25} & \textbf{Item 40} & \textbf{Item 55} \\ \midrule
\endhead
\midrule\multicolumn{5}{r}{{Continued on next page}} \\ \endfoot
\bottomrule\endlastfoot
Mistral-7B+Persona & 0.60 & 0.63 & 0.54 & 0.65 \\
Mistral-7B+Shape   & 0.81 & 0.62 & 0.39 & 0.58 \\
Mistral-7B+PSI    & 0.78 & 0.19 & 0.68 & 0.14 \\
Gemma-2-9B+Persona       & 0.69 & 0.42 & 0.62 & 0.40 \\
Gemma-2-9B+Shape         & 0.90 & 0.90 & 0.67 & 0.89 \\
Gemma-2-9B+PSI          & 0.80 & 0.55 & 0.65 & 0.44 \\
Gemma-2-27B+Persona      & 0.65 & 0.80 & 0.48 & 0.77 \\
Gemma-2-27B+Shape        & 0.87 & 0.95 & 0.35 & 0.93 \\
Gemma-2-27B+PSI         & 0.81 & 0.67 & 0.73 & 0.54 \\
Llama3-8B+Persona  & 0.64 & -0.10 & 0.55 & -0.21 \\
Llama3-8B+Shape    & 0.93 & -0.05 & 0.61 & 0.11 \\
Llama3-8B+PSI     & 0.72 & 0.77 & 0.73 & 0.79 \\
Llama3-70B+Persona & 0.81 & 0.76 & 0.70 & 0.82 \\
Llama3-70B+Shape   & 0.91 & 0.88 & 0.56 & 0.89 \\
Llama3-70B+PSI    & 0.81 & 0.64 & 0.78 & 0.57 \\
GPT-4o-mini+Persona          & 0.79 & 0.63 & 0.68 & 0.68 \\
GPT-4o-mini+Shape            & 0.61 & 0.87 & 0.23 & 0.86 \\
GPT-4o-mini+PSI             & 0.80 & 0.77 & 0.81 & 0.86 \\
GPT-4o+Persona               & 0.71 & 0.68 & 0.78 & 0.75 \\
GPT-4o+Shape                 & 0.90 & 0.87 & 0.62 & 0.82 \\
GPT-4o+PSI                  & 0.76 & 0.84 & 0.81 & 0.89 \\
Human                & 0.60 & 0.63 & 0.65 & 0.74 \\
\end{longtable}

\begin{longtable}{l c c c c}
\caption{Standardized factor loadings for Aesthetic Sensitivity items in BFI-2 TFM of Openness.}%
\label{tab:fc_aesthetic_sensitivity} \\
\toprule
\textbf{Model} 
 & \textbf{Item 5} & \textbf{Item 20} & \textbf{Item 35} & \textbf{Item 50} \\
\midrule
\endfirsthead
\multicolumn{5}{@{}l}{\tablename\ \thetable{} -- continued} \\ \toprule
\textbf{Model} 
 & \textbf{Item 5} & \textbf{Item 20} & \textbf{Item 35} & \textbf{Item 50} \\ \midrule
\endhead
\midrule\multicolumn{5}{r}{{Continued on next page}} \\ \endfoot
\bottomrule\endlastfoot
Mistral-7B+Persona & 0.35 & 0.76 & 0.95 & 0.25 \\
Mistral-7B+Shape   & 0.65 & 0.85 & 0.73 & 0.73 \\
Mistral-7B+PSI    & 0.41 & 0.89 & 0.79 & 0.18 \\
Gemma-2-9B+Persona       & 0.24 & 0.87 & 0.91 & 0.34 \\
Gemma-2-9B+Shape         & 0.85 & 0.86 & 0.88 & 0.89 \\
Gemma-2-9B+PSI          & 0.29 & -1.01 & -0.63 & 0.10 \\
Gemma-2-27B+Persona      & 0.85 & 0.83 & 0.89 & 0.44 \\
Gemma-2-27B+Shape        & 0.85 & 0.83 & 0.85 & 0.93 \\
Gemma-2-27B+PSI         & 0.88 & 0.88 & 0.95 & 0.40 \\
Llama3-8B+Persona  & 0.17 & -0.79 & -0.86 & 0.07 \\
Llama3-8B+Shape    & 0.19 & 0.86 & 0.80 & 0.38 \\
Llama3-8B+PSI     & 0.82 & 0.88 & 0.91 & 0.41 \\
Llama3-70B+Persona & 0.85 & 0.85 & 0.88 & 0.41 \\
Llama3-70B+Shape   & 0.91 & 0.83 & 0.82 & 0.91 \\
Llama3-70B+PSI    & 0.87 & 0.84 & 0.94 & 0.33 \\
GPT-4o-mini+Persona          & 0.79 & 0.87 & 0.88 & 0.64 \\
GPT-4o-mini+Shape            & 0.91 & 0.56 & 0.60 & 0.96 \\
GPT-4o-mini+PSI             & 0.90 & 0.85 & 0.89 & 0.51 \\
GPT-4o+Persona               & 0.84 & 0.82 & 0.86 & 0.69 \\
GPT-4o+Shape                 & 0.81 & 0.91 & 0.85 & 0.89 \\
GPT-4o+PSI                  & 0.87 & 0.85 & 0.90 & 0.64 \\
Human                & 0.66 & 0.86 & 0.87 & 0.65 \\
\end{longtable}

\begin{longtable}{l c c c c}
\caption{Standardized factor loadings for Creative Imagination items in BFI-2 TFM of Openness.}%
\label{tab:fc_creative_imagination} \\
\toprule
\textbf{Model} 
 & \textbf{Item 15} & \textbf{Item 30} & \textbf{Item 45} & \textbf{Item 60} \\
\midrule
\endfirsthead
\multicolumn{5}{@{}l}{\tablename\ \thetable{} -- continued} \\ \toprule
\textbf{Model} 
 & \textbf{Item 15} & \textbf{Item 30} & \textbf{Item 45} & \textbf{Item 60} \\ \midrule
\endhead
\midrule\multicolumn{5}{r}{{Continued on next page}} \\ \endfoot
\bottomrule\endlastfoot
Mistral-7B+Persona & 0.76 & 0.54 & 0.18 & 0.72 \\
Mistral-7B+Shape   & 0.91 & 0.55 & 0.57 & 0.84 \\
Mistral-7B+PSI    & 0.85 & 0.34 & 0.06 & 0.91 \\
Gemma-2-9B+Persona       & 0.74 & 0.11 & 0.34 & 0.78 \\
Gemma-2-9B+Shape         & 0.82 & 0.93 & 0.86 & 0.78 \\
Gemma-2-9B+PSI          & 0.78 & -0.08 & 0.04 & 0.81 \\
Gemma-2-27B+Persona      & 0.80 & 0.70 & 0.55 & 0.80 \\
Gemma-2-27B+Shape        & 0.87 & 0.97 & 0.95 & 0.75 \\
Gemma-2-27B+PSI         & 0.60 & 0.74 & 0.55 & 0.76 \\
Llama3-8B+Persona  & 0.71 & -0.40 & -0.34 & 0.76 \\
Llama3-8B+Shape    & 0.93 & -0.04 & 0.05 & 0.80 \\
Llama3-8B+PSI     & 0.62 & 0.78 & 0.80 & 0.59 \\
Llama3-70B+Persona & 0.89 & 0.63 & 0.49 & 0.95 \\
Llama3-70B+Shape   & 0.86 & 0.81 & 0.76 & 0.81 \\
Llama3-70B+PSI    & 0.85 & 0.60 & 0.37 & 0.95 \\
GPT-4o-mini+Persona          & 0.80 & 0.74 & 0.50 & 0.90 \\
GPT-4o-mini+Shape            & 0.47 & 0.97 & 0.87 & 0.32 \\
GPT-4o-mini+PSI             & 0.70 & 0.86 & 0.78 & 0.81 \\
GPT-4o+Persona               & 0.65 & 0.81 & 0.68 & 0.70 \\
GPT-4o+Shape                 & 0.79 & 0.84 & 0.79 & 0.79 \\
GPT-4o+PSI                  & 0.66 & 0.88 & 0.88 & 0.79 \\
Human                & 0.72 & 0.77 & 0.66 & 0.80 \\
\end{longtable}

\begin{longtable}{l c c c}
\caption{Standardized factor loadings for five‐factor model: Extraversion subdimension.}%
\label{tab:ffm_extraversion} \\
\toprule
\textbf{Model} 
 & \textbf{Sociability} & \textbf{Assertiveness} & \textbf{Energy Level} \\
\midrule
\endfirsthead
\multicolumn{4}{@{}l}{\tablename\ \thetable{} -- continued} \\ \toprule
\textbf{Model} 
 & \textbf{Sociability} & \textbf{Assertiveness} & \textbf{Energy Level} \\ \midrule
\endhead
\midrule\multicolumn{4}{r}{{Continued on next page}} \\ \endfoot
\bottomrule\endlastfoot
Mistral-7B+Persona & 0.59 & 0.69 & 0.87 \\
Mistral-7B+Shape   & 0.74 & 0.94 & 0.91 \\
Mistral-7B+PSI    & 0.76 & 0.76 & 0.95 \\
Gemma-2-9B+Persona       & 0.66 & 0.83 & 0.81 \\
Gemma-2-9B+Shape         & 0.86 & 0.84 & 0.92 \\
Gemma-2-9B+PSI          & 0.65 & 0.81 & 0.84 \\
Gemma-2-27B+Persona      & 0.76 & 0.82 & 0.90 \\
Gemma-2-27B+Shape        & 0.88 & 0.83 & 0.95 \\
Gemma-2-27B+PSI         & 0.84 & 0.78 & 0.89 \\
Llama3-8B+Persona  & 0.63 & 0.66 & 0.65 \\
Llama3-8B+Shape    & 0.70 & 0.87 & 0.69 \\
Llama3-8B+PSI     & 0.76 & 0.74 & 0.94 \\
Llama3-70B+Persona & 0.85 & 0.86 & 0.86 \\
Llama3-70B+Shape   & 0.87 & 0.74 & 0.85 \\
Llama3-70B+PSI    & 0.77 & 0.68 & 0.90 \\
GPT-4o-mini+Persona          & 0.82 & 0.78 & 0.93 \\
GPT-4o-mini+Shape            & 0.76 & 0.58 & 1.03 \\
GPT-4o-mini+PSI             & 0.82 & 0.69 & 0.94 \\
GPT-4o+Persona               & 0.76 & 0.67 & 0.88 \\
GPT-4o+Shape                 & 0.91 & 0.77 & 0.89 \\
GPT-4o+PSI                  & 0.82 & 0.71 & 0.93 \\
Human                & 0.66 & 0.59 & 0.77 \\
\end{longtable}

\begin{longtable}{l c c c}
\caption{Standardized factor loadings for five‐factor model: Agreeableness subdimension.}%
\label{tab:ffm_agreeableness} \\
\toprule
\textbf{Model} 
 & \textbf{Compassion} & \textbf{Respectfulness} & \textbf{Trust} \\
\midrule
\endfirsthead
\multicolumn{4}{@{}l}{\tablename\ \thetable{} -- continued} \\ \toprule
\textbf{Model} 
 & \textbf{Compassion} & \textbf{Respectfulness} & \textbf{Trust} \\ \midrule
\endhead
\midrule\multicolumn{4}{r}{{Continued on next page}} \\ \endfoot
\bottomrule\endlastfoot
Mistral-7B+Persona & 0.76 & 0.85 & 0.67 \\
Mistral-7B+Shape   & 0.93 & 0.83 & 0.90 \\
Mistral-7B+PSI    & 0.86 & 0.88 & 0.90 \\
Gemma-2-9B+Persona       & 0.86 & 0.92 & 0.81 \\
Gemma-2-9B+Shape         & 0.98 & 0.92 & 0.95 \\
Gemma-2-9B+PSI          & 0.85 & 0.82 & 0.81 \\
Gemma-2-27B+Persona      & 0.88 & 0.90 & 0.91 \\
Gemma-2-27B+Shape        & 1.00 & 0.87 & 0.95 \\
Gemma-2-27B+PSI         & 0.80 & 0.91 & 0.85 \\
Llama3-8B+Persona  & 0.80 & 0.86 & 0.76 \\
Llama3-8B+Shape    & 0.94 & 0.79 & 0.90 \\
Llama3-8B+PSI     & 0.83 & 0.79 & 0.84 \\
Llama3-70B+Persona & 0.90 & 0.94 & 0.87 \\
Llama3-70B+Shape   & 0.99 & 0.86 & 0.95 \\
Llama3-70B+PSI    & 0.79 & 0.95 & 0.84 \\
GPT-4o-mini+Persona          & 0.89 & 0.94 & 0.88 \\
GPT-4o-mini+Shape            & 0.97 & 0.86 & 0.94 \\
GPT-4o-mini+PSI             & 0.85 & 0.89 & 0.86 \\
GPT-4o+Persona               & 0.82 & 0.90 & 0.75 \\
GPT-4o+Shape                 & 0.96 & 0.90 & 0.93 \\
GPT-4o+PSI                  & 0.83 & 0.88 & 0.84 \\
Human                & 0.70 & 0.80 & 0.65 \\
\end{longtable}

\begin{longtable}{l c c c}
\caption{Standardized factor loadings for five‐factor model: Conscientiousness subdimension.}%
\label{tab:ffm_conscientiousness} \\
\toprule
\textbf{Model} 
 & \textbf{Organization} & \textbf{Productiveness} & \textbf{Responsibility} \\
\midrule
\endfirsthead
\multicolumn{4}{@{}l}{\tablename\ \thetable{} -- continued} \\ \toprule
\textbf{Model} 
 & \textbf{Organization} & \textbf{Productiveness} & \textbf{Responsibility} \\ \midrule
\endhead
\midrule\multicolumn{4}{r}{{Continued on next page}} \\ \endfoot
\bottomrule\endlastfoot
Mistral-7B+Persona & 0.75 & 0.85 & 0.85 \\
Mistral-7B+Shape   & 0.96 & 0.88 & 0.90 \\
Mistral-7B+PSI    & 0.78 & 0.85 & 0.86 \\
Gemma-2-9B+Persona       & 0.77 & 0.68 & 0.93 \\
Gemma-2-9B+Shape         & 0.92 & 0.76 & 0.95 \\
Gemma-2-9B+PSI          & 0.66 & 0.79 & 0.89 \\
Gemma-2-27B+Persona      & 0.86 & 0.80 & 0.92 \\
Gemma-2-27B+Shape        & 0.91 & 0.80 & 0.97 \\
Gemma-2-27B+PSI         & 0.85 & 0.84 & 0.86 \\
Llama3-8B+Persona  & 0.62 & 0.50 & 0.77 \\
Llama3-8B+Shape    & 0.83 & 0.75 & 0.92 \\
Llama3-8B+PSI     & 0.79 & 0.80 & 0.86 \\
Llama3-70B+Persona & 0.87 & 0.75 & 0.95 \\
Llama3-70B+Shape   & 0.97 & 0.80 & 0.93 \\
Llama3-70B+PSI    & 0.87 & 0.83 & 0.92 \\
GPT-4o-mini+Persona          & 0.91 & 0.83 & 0.94 \\
GPT-4o-mini+Shape            & 0.96 & 0.86 & 0.90 \\
GPT-4o-mini+PSI             & 0.86 & 0.81 & 0.90 \\
GPT-4o+Persona               & 0.87 & 0.77 & 0.82 \\
GPT-4o+Shape                 & 0.94 & 0.84 & 0.93 \\
GPT-4o+PSI                  & 0.87 & 0.83 & 0.88 \\
Human                & 0.70 & 0.89 & 0.79 \\
\end{longtable}

\begin{longtable}{l c c c}
\caption{Standardized factor loadings for five‐factor model: Neuroticism subdimension.}%
\label{tab:ffm_neuroticism} \\
\toprule
\textbf{Model} 
 & \textbf{Anxiety} & \textbf{Depression} & \textbf{Emotional Volatility} \\
\midrule
\endfirsthead
\multicolumn{4}{@{}l}{\tablename\ \thetable{} -- continued} \\ \toprule
\textbf{Model} 
 & \textbf{Anxiety} & \textbf{Depression} & \textbf{Emotional Volatility} \\ \midrule
\endhead
\midrule\multicolumn{4}{r}{{Continued on next page}} \\ \endfoot
\bottomrule\endlastfoot
Mistral-7B+Persona & 0.51 & 0.80 & 0.75 \\
Mistral-7B+Shape   & 0.67 & 1.00 & 0.76 \\
Mistral-7B+PSI    & 0.66 & 0.96 & 0.81 \\
Gemma-2-9B+Persona       & 0.76 & 0.96 & 0.70 \\
Gemma-2-9B+Shape         & 0.82 & 1.01 & 0.73 \\
Gemma-2-9B+PSI          & 0.72 & 0.96 & 0.78 \\
Gemma-2-27B+Persona      & 0.70 & 0.85 & 0.82 \\
Gemma-2-27B+Shape        & 0.79 & 1.01 & 0.72 \\
Gemma-2-27B+PSI         & 0.62 & 0.90 & 0.82 \\
Llama3-8B+Persona  & 0.44 & 0.91 & 0.60 \\
Llama3-8B+Shape    & 0.42 & 0.91 & 0.65 \\
Llama3-8B+PSI     & 0.80 & 0.87 & 0.81 \\
Llama3-70B+Persona & 0.75 & 0.96 & 0.79 \\
Llama3-70B+Shape   & 0.87 & 0.96 & 0.81 \\
Llama3-70B+PSI    & 0.75 & 0.88 & 0.85 \\
GPT-4o-mini+Persona          & 0.73 & 0.97 & 0.74 \\
GPT-4o-mini+Shape            & 0.78 & 0.97 & 0.78 \\
GPT-4o-mini+PSI             & 0.74 & 0.97 & 0.78 \\
GPT-4o+Persona               & 0.69 & 0.91 & 0.63 \\
GPT-4o+Shape                 & 0.80 & 1.02 & 0.71 \\
GPT-4o+PSI                  & 0.74 & 0.99 & 0.76 \\
Human                & 0.84 & 0.88 & 0.85 \\
\end{longtable}

\begin{longtable}{l c c c}
\caption{Standardized factor loadings for five‐factor model: Openness subdimension.}%
\label{tab:ffm_openness} \\
\toprule
\textbf{Model} 
 & \textbf{Intellectual Curiosity} & \textbf{Aesthetic Sensitivity} & \textbf{Creative Imagination} \\
\midrule
\endfirsthead
\multicolumn{4}{@{}l}{\tablename\ \thetable{} -- continued} \\ \toprule
\textbf{Model} 
 & \textbf{Intellectual Curiosity} & \textbf{Aesthetic Sensitivity} & \textbf{Creative Imagination} \\ \midrule
\endhead
\midrule\multicolumn{4}{r}{{Continued on next page}} \\ \endfoot
\bottomrule\endlastfoot
Mistral-7B+Persona & 0.74 & 0.64 & 0.70 \\
Mistral-7B+Shape   & 0.89 & 0.94 & 0.79 \\
Mistral-7B+PSI    & 0.74 & 0.53 & 0.77 \\
Gemma-2-9B+Persona       & 0.85 & 0.58 & 0.74 \\
Gemma-2-9B+Shape         & 0.96 & 0.94 & 0.87 \\
Gemma-2-9B+PSI          & 0.86 & 0.40 & 0.62 \\
Gemma-2-27B+Persona      & 0.82 & 0.81 & 0.71 \\
Gemma-2-27B+Shape        & 0.94 & 0.96 & 0.89 \\
Gemma-2-27B+PSI         & 0.74 & 0.67 & 0.91 \\
Llama3-8B+Persona  & 0.69 & 0.50 & 0.63 \\
Llama3-8B+Shape    & 0.78 & 0.96 & 0.78 \\
Llama3-8B+PSI     & 0.80 & 0.63 & 0.91 \\
Llama3-70B+Persona & 0.66 & 0.65 & 0.89 \\
Llama3-70B+Shape   & 0.93 & 0.96 & 0.82 \\
Llama3-70B+PSI    & 0.85 & 0.58 & 0.78 \\
GPT-4o-mini+Persona          & 0.86 & 0.77 & 0.80 \\
GPT-4o-mini+Shape            & 0.94 & 0.96 & 0.93 \\
GPT-4o-mini+PSI             & 0.79 & 0.70 & 0.88 \\
GPT-4o+Persona               & 0.77 & 0.73 & 0.91 \\
GPT-4o+Shape                 & 0.97 & 0.95 & 0.90 \\
GPT-4o+PSI                  & 0.75 & 0.70 & 0.96 \\
Human                & 0.74 & 0.71 & 0.81 \\
\end{longtable}

\begin{longtable}{l c c c}
\caption{Inter‐factor correlations for BFI-2 TFM of Extraversion.}%
\label{tab:inter_extraversion} \\
\toprule
\textbf{Model} 
 & \textbf{Soc–Ass} 
 & \textbf{Soc–Ene} 
 & \textbf{Ass–Ene} \\
\midrule
\endfirsthead
\multicolumn{4}{@{}l}{\tablename\ \thetable{} – continued} \\ \toprule
\textbf{Model} 
 & \textbf{Soc–Ass} 
 & \textbf{Soc–Ene} 
 & \textbf{Ass–Ene} \\ \midrule
\endhead
\midrule\multicolumn{4}{r}{{Continued}} \\ \endfoot
\bottomrule\endlastfoot
Mistral-7B+Persona & 0.72 & 0.67 & 0.87 \\
Mistral-7B+Shape   & 0.81 & 0.80 & 1.01 \\
Mistral-7B+PSI    & 0.79 & 0.89 & 0.88 \\
Gemma-2-9B+Persona       & 0.73 & 0.66 & 0.64 \\
Gemma-2-9B+Shape         & 0.91 & 0.88 & 0.97 \\
Gemma-2-9B+PSI          & 0.72 & 0.88 & 0.81 \\
Gemma-2-27B+Persona      & 0.79 & 0.83 & 0.79 \\
Gemma-2-27B+Shape        & 0.71 & 0.92 & 0.68 \\
Gemma-2-27B+PSI         & 0.89 & 0.87 & 0.95 \\
Llama3-8B+Persona  & -0.15 & -0.32 & 0.96 \\
Llama3-8B+Shape    & -0.79 & 1.15 & -0.69 \\
Llama3-8B+PSI     & 0.73 & 0.85 & 0.84 \\
Llama3-70B+Persona & 0.89 & 0.84 & 0.82 \\
Llama3-70B+Shape   & 0.91 & 0.82 & 0.61 \\
Llama3-70B+PSI    & 0.80 & 0.85 & 0.88 \\
GPT-4o-mini+Persona          & 0.81 & 0.90 & 0.78 \\
GPT-4o-mini+Shape            & 0.60 & 0.95 & 0.53 \\
GPT-4o-mini+PSI             & 0.73 & 0.87 & 0.73 \\
GPT-4o+Persona               & 0.72 & 0.79 & 0.66 \\
GPT-4o+Shape                 & 0.77 & 0.91 & 0.65 \\
GPT-4o+PSI                  & 0.71 & 0.86 & 0.74 \\
Human                & 0.59 & 0.64 & 0.47 \\
\end{longtable}

\begin{longtable}{l c c c}
\caption{Inter‐factor correlations for BFI-2 TFM of Agreeableness.}%
\label{tab:inter_agreeableness} \\
\toprule
\textbf{Model}
 & \textbf{Com–Res}
 & \textbf{Com–Tru}
 & \textbf{Res–Tru} \\
\midrule
\endfirsthead
\multicolumn{4}{@{}l}{\tablename\ \thetable{} – continued} \\ \toprule
\textbf{Model}
  & \textbf{Com–Res}
 & \textbf{Com–Tru}
 & \textbf{Res–Tru} \\ \midrule
\endhead
\midrule\multicolumn{4}{r}{{Continued}} \\ \endfoot
\bottomrule\endlastfoot
Mistral-7B+Persona & 1.00 & 0.89 & 0.71 \\
Mistral-7B+Shape   & 0.93 & 0.96 & 0.91 \\
Mistral-7B+PSI    & 0.91 & 0.98 & 0.92 \\
Gemma-2-9B+Persona       & 0.93 & 0.88 & 0.94 \\
Gemma-2-9B+Shape         & 0.90 & 0.98 & 0.91 \\
Gemma-2-9B+PSI          & 0.93 & -0.93 & -0.90 \\
Gemma-2-27B+Persona      & 0.96 & 0.96 & 0.97 \\
Gemma-2-27B+Shape        & 0.88 & 1.00 & 0.84 \\
Gemma-2-27B+PSI         & 0.79 & 0.79 & 0.84 \\
Llama3-8B+Persona  & 1.05 & -0.85 & -0.97 \\
Llama3-8B+Shape    & 0.99 & 1.09 & 0.86 \\
Llama3-8B+PSI     & 0.87 & 0.88 & 0.83 \\
Llama3-70B+Persona & 0.98 & 0.94 & 0.96 \\
Llama3-70B+Shape   & 0.87 & 0.96 & 0.85 \\
Llama3-70B+PSI    & 0.90 & 0.85 & 0.88 \\
GPT-4o-mini+Persona          & 0.99 & 0.91 & 0.94 \\
GPT-4o-mini+Shape            & 0.88 & 0.99 & 0.90 \\
GPT-4o-mini+PSI             & 0.90 & 0.82 & 0.86 \\
GPT-4o+Persona               & 0.90 & 0.83 & 0.95 \\
GPT-4o+Shape                 & 0.92 & 0.94 & 0.93 \\
GPT-4o+PSI                   & 0.86 & 0.80 & 0.85 \\
Human                & 0.81 & 0.70 & 0.58 \\
\end{longtable}

\begin{longtable}{l c c c}
\caption{Inter‐factor correlations for BFI-2 TFM of Conscientiousness.}%
\label{tab:inter_conscientiousness} \\
\toprule
\textbf{Model}
 & \textbf{Org–Pro}
 & \textbf{Org–Res}
 & \textbf{Pro–Res} \\
\midrule
\endfirsthead
\multicolumn{4}{@{}l}{\tablename\ \thetable{} – continued} \\ \toprule
\textbf{Model}
  & \textbf{Org–Pro}
 & \textbf{Org–Res}
 & \textbf{Pro–Res}  \\ \midrule
\endhead
\midrule\multicolumn{4}{r}{{Continued}} \\ \endfoot
\bottomrule\endlastfoot
Mistral-7B+Persona  & 0.93 & 0.85 & 0.87 \\
Mistral-7B+Shape    & 1.01 & 1.04 & 0.90 \\
Mistral-7B+PSI     & 0.96 & 0.83 & 0.81 \\
Gemma-2-9B+Persona        & 0.62 & 0.81 & 0.77 \\
Gemma-2-9B+Shape          & 0.89 & 1.00 & 0.66 \\
Gemma-2-9B+PSI           & 0.77 & 0.73 & 0.82 \\
Gemma-2-27B+Persona       & 0.84 & 0.90 & 0.83 \\
Gemma-2-27B+Shape         & 0.94 & 0.88 & 0.75 \\
Gemma-2-27B+PSI          & 0.97 & 0.85 & 0.84 \\
Llama3-8B+Persona   & -0.98 & -0.75 & 0.99 \\
Llama3-8B+Shape     & 0.87 & 0.87 & 0.91 \\
Llama3-8B+PSI      & 0.76 & 0.76 & 0.74 \\
Llama3-70B+Persona  & 0.81 & 0.85 & 0.87 \\
Llama3-70B+Shape    & 0.69 & 0.86 & 0.87 \\
Llama3-70B+PSI     & 0.88 & 0.91 & 0.88 \\
GPT-4o-mini+Persona           & 0.88 & 0.93 & 0.90 \\
GPT-4o-mini+Shape             & 1.07 & 0.92 & 0.85 \\
GPT-4o-mini+PSI              & 0.85 & 0.90 & 0.83 \\
GPT-4o+Persona                & 0.86 & 1.01 & 0.82 \\
GPT-4o+Shape                  & 0.97 & 0.99 & 0.87 \\
GPT-4o+PSI                    & 0.86 & 0.86 & 0.80 \\
Human                 & 0.75 & 0.70 & 0.86 \\
\end{longtable}

\begin{longtable}{l c c c}
\caption{Inter‐factor correlations for BFI-2 TFM of Neuroticism.}%
\label{tab:inter_neuroticism} \\
\toprule
\textbf{Model}
 & \textbf{Anx–Dep}
 & \textbf{Anx–Emo}
 & \textbf{Dep–Emo} \\
\midrule
\endfirsthead
\multicolumn{4}{@{}l}{\tablename\ \thetable{} – continued} \\ \toprule
\textbf{Model}
 & \textbf{Anx–Dep}
 & \textbf{Anx–Emo}
 & \textbf{Dep–Emo} \\ \midrule
\endhead
\midrule\multicolumn{4}{r}{{Continued}} \\ \endfoot
\bottomrule\endlastfoot
Mistral-7B+Persona & 0.78 & 0.51 & 0.60 \\
Mistral-7B+Shape   & 0.93 & 0.82 & 0.91 \\
Mistral-7B+PSI    & 0.81 & 0.83 & 0.93 \\
Gemma-2-9B+Persona       & 0.84 & 0.79 & 0.68 \\
Gemma-2-9B+Shape         & 1.01 & 0.94 & 0.92 \\
Gemma-2-9B+PSI          & 0.77 & 0.85 & 0.87 \\
Gemma-2-27B+Persona      & 0.66 & 0.77 & 0.63 \\
Gemma-2-27B+Shape        & 0.97 & 0.89 & 0.91 \\
Gemma-2-27B+PSI         & 0.70 & 0.82 & 0.84 \\
Llama3-8B+Persona  & 1.13 & -1.10 & -1.04 \\
Llama3-8B+Shape    & 0.18 & -0.90 & 0.23 \\
Llama3-8B+PSI     & 0.85 & 0.95 & 0.87 \\
Llama3-70B+Persona & 0.88 & 0.79 & 0.85 \\
Llama3-70B+Shape   & 0.96 & 0.87 & 0.84 \\
Llama3-70B+PSI    & 0.77 & 0.83 & 0.83 \\
GPT-4o-mini+Persona          & 0.88 & 0.83 & 0.91 \\
GPT-4o-mini+Shape            & 0.90 & 0.78 & 0.93 \\
GPT-4o-mini+PSI             & 0.80 & 0.89 & 0.86 \\
GPT-4o+Persona               & 0.88 & 0.87 & 0.84 \\
GPT-4o+Shape                 & 0.95 & 0.87 & 0.89 \\
GPT-4o+PSI                   & 0.85 & 0.91 & 0.86 \\
Human                & 0.81 & 0.90 & 0.79 \\
\end{longtable}

\begin{longtable}{l c c c}
\caption{Inter‐factor correlations for BFI-2 TFM of Openness.}%
\label{tab:inter_openness} \\
\toprule
\textbf{Model}
 & \textbf{Int–Aes}
 & \textbf{Int–Cre}
 & \textbf{Aes–Cre} \\
\midrule
\endfirsthead
\multicolumn{4}{@{}l}{\tablename\ \thetable{} – continued} \\ \toprule
\textbf{Model}
  & \textbf{Int–Aes}
 & \textbf{Int–Cre}
 & \textbf{Aes–Cre} \\ \midrule
\endhead
\midrule\multicolumn{4}{r}{{Continued}} \\ \endfoot
\bottomrule\endlastfoot
Mistral-7B+Persona & 0.35 & 0.65 & 0.58 \\
Mistral-7B+Shape   & 1.09 & 0.83 & 0.77 \\
Mistral-7B+PSI    & 0.51 & 0.67 & 0.63 \\
Gemma-2-9B+Persona       & 0.61 & 0.96 & 0.64 \\
Gemma-2-9B+Shape         & 0.97 & 0.94 & 0.90 \\
Gemma-2-9B+PSI          & -0.47 & 0.89 & -0.54 \\
Gemma-2-27B+Persona      & 0.65 & 0.66 & 0.67 \\
Gemma-2-27B+Shape        & 0.98 & 0.94 & 0.93 \\
Gemma-2-27B+PSI         & 0.44 & 0.87 & 0.71 \\
Llama3-8B+Persona  & -0.70 & 1.06 & -0.64 \\
Llama3-8B+Shape    & 0.97 & 0.96 & 0.98 \\
Llama3-8B+PSI     & 0.54 & 0.85 & 0.74 \\
Llama3-70B+Persona & 0.59 & 0.67 & 0.60 \\
Llama3-70B+Shape   & 1.00 & 0.89 & 0.85 \\
Llama3-70B+PSI    & 0.48 & 0.74 & 0.51 \\
GPT-4o-mini+Persona          & 0.81 & 0.82 & 0.64 \\
GPT-4o-mini+Shape            & 1.00 & 0.98 & 1.00 \\
GPT-4o-mini+PSI             & 0.57 & 0.79 & 0.75 \\
GPT-4o+Persona               & 0.68 & 0.85 & 0.79 \\
GPT-4o+Shape                 & 1.02 & 0.99 & 0.93 \\
GPT-4o+PSI                   & 0.54 & 0.80 & 0.78 \\
Human                & 0.66 & 0.74 & 0.63 \\
\end{longtable}

\begin{table*}[h]

\centering
\scriptsize
\resizebox{\linewidth}{!}{%
\begin{tabular}{l*{10}{c}}
\toprule
\textbf{Model} & 
\makecell{\textbf{E\textasciitilde}\textbf{A}} & 
\makecell{\textbf{E\textasciitilde}\textbf{C}} & 
\makecell{\textbf{E\textasciitilde}\textbf{N}} & 
\makecell{\textbf{E\textasciitilde}\textbf{O}} & 
\makecell{\textbf{A\textasciitilde}\textbf{C}} & 
\makecell{\textbf{A\textasciitilde}\textbf{N}} & 
\makecell{\textbf{A\textasciitilde}\textbf{O}} & 
\makecell{\textbf{C\textasciitilde}\textbf{N}} & 
\makecell{\textbf{C\textasciitilde}\textbf{O}} & 
\makecell{\textbf{N\textasciitilde}\textbf{O}} \\
\midrule
Mistral-7B+Persona & 0.57 & 0.74 & -0.68 & 0.42 & 0.86 & -0.59 & 0.49 & -0.78 & 0.33 & -0.07 \\
Mistral-7B+Shape & 0.58 & 0.67 & -0.80 & 0.64 & 0.79 & -0.78 & 0.75 & -0.82 & 0.51 & -0.61 \\
Mistral-7B+PSI & 0.57 & 0.60 & -0.64 & 0.64 & 0.73 & -0.68 & 0.43 & -0.78 & 0.19 & -0.18 \\
Gemma-2-9B+Persona & 0.25 & 0.45 & -0.63 & 0.53 & 0.74 & -0.60 & 0.42 & -0.67 & 0.32 & -0.30 \\
Gemma-2-9B+Shape & 0.66 & 0.42 & -0.77 & 0.83 & 0.66 & -0.80 & 0.82 & -0.73 & 0.43 & -0.62 \\
Gemma-2-9B+PSI & 0.48 & 0.64 & -0.70 & 0.71 & 0.69 & -0.64 & 0.52 & -0.76 & 0.45 & -0.34 \\
Gemma-2-27B+Persona & 0.48 & 0.57 & -0.67 & 0.53 & 0.79 & -0.72 & 0.62 & -0.84 & 0.41 & -0.36 \\
Gemma-2-27B+Shape & 0.72 & 0.47 & -0.83 & 0.82 & 0.77 & -0.82 & 0.85 & -0.80 & 0.48 & -0.71 \\
Gemma-2-27B+PSI & 0.34 & 0.41 & -0.43 & 0.47 & 0.67 & -0.52 & 0.13 & -0.76 & 0.08 & 0.17 \\
Llama3-8B+Persona & 0.30 & 0.30 & -0.64 & 0.42 & 0.62 & -0.63 & 0.40 & -0.59 & 0.03 & -0.25 \\
Llama3-8B+Shape & 0.40 & 0.27 & -0.62 & 0.68 & 0.58 & -0.69 & 0.76 & -0.55 & 0.27 & -0.62 \\
Llama3-8B+PSI & 0.53 & 0.56 & -0.46 & 0.36 & 0.63 & -0.35 & 0.22 & -0.55 & 0.16 & 0.16 \\
Llama3-70B+Persona & 0.10 & 0.26 & -0.59 & 0.51 & 0.77 & -0.54 & 0.25 & -0.63 & 0.16 & -0.23 \\
Llama3-70B+Shape & 0.54 & 0.31 & -0.71 & 0.70 & 0.65 & -0.74 & 0.89 & -0.82 & 0.45 & -0.65 \\
Llama3-70B+PSI & 0.28 & 0.36 & -0.33 & 0.51 & 0.64 & -0.50 & 0.15 & -0.63 & 0.20 & 0.12 \\
GPT-4o-mini+Persona & 0.51 & 0.55 & -0.76 & 0.52 & 0.79 & -0.73 & 0.54 & -0.69 & 0.36 & -0.39 \\
GPT-4o-mini+Shape & 0.81 & 0.59 & -0.61 & 0.87 & 0.75 & -0.79 & 0.91 & -0.75 & 0.58 & -0.62 \\
GPT-4o-mini+PSI & 0.50 & 0.45 & -0.62 & 0.33 & 0.65 & -0.52 & 0.19 & -0.64 & 0.07 & 0.10 \\
GPT-4o+Persona & 0.26 & 0.43 & -0.61 & 0.37 & 0.56 & -0.40 & 0.31 & -0.52 & 0.10 & -0.06 \\
GPT-4o+Shape & 0.59 & 0.36 & -0.80 & 0.72 & 0.65 & -0.66 & 0.79 & -0.67 & 0.27 & -0.48 \\
GPT-4o+PSI & 0.44 & 0.45 & -0.63 & 0.30 & 0.60 & -0.47 & 0.15 & -0.58 & 0.06 & 0.09 \\
Human & 0.28 & 0.56 & -0.61 & 0.35 & 0.47 & -0.43 & 0.27 & -0.61 & 0.17 & -0.17 \\
\bottomrule
\end{tabular}%
} 
\caption{Inter-factor correlations for BFI-2 FFM. E = Extraversion; A = Agreeableness; C = Conscientiousness; N = Neuroticism; O = Openness.}
\label{tab:inter-factor-ffm}
\end{table*}

\begin{longtable}{l c c c}
\caption{Cronbach’s alpha for Extraversion facets.}%
\label{tab:alpha_extraversion} \\
\toprule
\textbf{Model} 
 & \textbf{Sociability} 
 & \textbf{Assertiveness} 
 & \textbf{Energy Level} \\
\midrule
\endfirsthead
\multicolumn{4}{@{}l}{\tablename\ \thetable{} continued} \\ \toprule
\textbf{Model} 
 & \textbf{Sociability} 
 & \textbf{Assertiveness} 
 & \textbf{Energy Level} \\
\midrule
\endhead
\midrule\multicolumn{4}{r}{{Continued on next page}} \\ \endfoot
\bottomrule\endlastfoot
Mistral-7B+Persona   & 0.68 & 0.55 & 0.78 \\
Mistral-7B+Shape     & 0.81 & 0.79 & 0.88 \\
Mistral-7B+PSI      & 0.84 & 0.61 & 0.72 \\
Gemma-2-9B+Persona         & 0.82 & 0.71 & 0.79 \\
Gemma-2-9B+Shape           & 0.79 & 0.74 & 0.92 \\
Gemma-2-9B+PSI            & 0.86 & 0.67 & 0.47 \\
Gemma-2-27B+Persona        & 0.83 & 0.77 & 0.88 \\
Gemma-2-27B+Shape          & 0.89 & 0.86 & 0.93 \\
Gemma-2-27B+PSI           & 0.88 & 0.62 & 0.83 \\
Llama3-8B+Persona    & 0.44 & 0.01 & 0.13 \\
Llama3-8B+Shape      & 0.58 & 0.49 & 0.32 \\
Llama3-8B+PSI       & 0.86 & 0.71 & 0.83 \\
Llama3-70B+Persona   & 0.86 & 0.80 & 0.86 \\
Llama3-70B+Shape     & 0.76 & 0.86 & 0.87 \\
Llama3-70B+PSI      & 0.82 & 0.55 & 0.77 \\
GPT-4o-mini+Persona            & 0.84 & 0.85 & 0.88 \\
GPT-4o-mini+Shape              & 0.74 & 0.77 & 0.84 \\
GPT-4o-mini+PSI               & 0.92 & 0.81 & 0.88 \\
GPT-4o+Persona                 & 0.85 & 0.74 & 0.86 \\
GPT-4o+Shape                   & 0.87 & 0.93 & 0.90 \\
GPT-4o+PSI                    & 0.91 & 0.81 & 0.89 \\
Human                  & 0.87 & 0.81 & 0.73 \\
\end{longtable}

\begin{longtable}{l c c c}
\caption{Cronbach’s alpha for Agreeableness facets.}%
\label{tab:alpha_agreeableness} \\
\toprule
\textbf{Model} 
 & \textbf{Compassion} 
 & \textbf{Respectfulness} 
 & \textbf{Trust} \\
\midrule
\endfirsthead
\multicolumn{4}{@{}l}{\tablename\ \thetable{} continued} \\ \toprule
\textbf{Model} 
 & \textbf{Compassion} 
 & \textbf{Respectfulness} 
 & \textbf{Trust} \\
\midrule
\endhead
\midrule\multicolumn{4}{r}{{Continued on next page}} \\ \endfoot
\bottomrule\endlastfoot
Mistral-7B+Persona   & 0.56 & 0.78 & 0.61 \\
Mistral-7B+Shape     & 0.79 & 0.78 & 0.83 \\
Mistral-7B+PSI      & 0.80 & 0.81 & 0.77 \\
Gemma-2-9B+Persona         & 0.86 & 0.78 & 0.69 \\
Gemma-2-9B+Shape           & 0.95 & 0.96 & 0.96 \\
Gemma-2-9B+PSI            & 0.84 & 0.66 & 0.64 \\
Gemma-2-27B+Persona        & 0.84 & 0.79 & 0.83 \\
Gemma-2-27B+Shape          & 0.95 & 0.95 & 0.95 \\
Gemma-2-27B+PSI           & 0.85 & 0.83 & 0.87 \\
Llama3-8B+Persona    & 0.54 & 0.44 & 0.36 \\
Llama3-8B+Shape      & 0.68 & 0.74 & 0.69 \\
Llama3-8B+PSI       & 0.74 & 0.72 & 0.75 \\
Llama3-70B+Persona   & 0.87 & 0.85 & 0.79 \\
Llama3-70B+Shape     & 0.93 & 0.93 & 0.94 \\
Llama3-70B+PSI      & 0.83 & 0.87 & 0.87 \\
GPT-4o-mini+Persona            & 0.86 & 0.80 & 0.81 \\
GPT-4o-mini+Shape              & 0.88 & 0.87 & 0.85 \\
GPT-4o-mini+PSI               & 0.89 & 0.81 & 0.87 \\
GPT-4o+Persona                 & 0.84 & 0.79 & 0.64 \\
GPT-4o+Shape                   & 0.95 & 0.94 & 0.96 \\
GPT-4o+PSI                    & 0.89 & 0.84 & 0.88 \\
Human                  & 0.67 & 0.74 & 0.80 \\
\end{longtable}

\begin{longtable}{l c c c}
\caption{Cronbach’s alpha for Conscientiousness facets.}%
\label{tab:alpha_conscientiousness} \\
\toprule
\textbf{Model} 
 & \textbf{Organization} 
 & \textbf{Productiveness} 
 & \textbf{Responsibility} \\
\midrule
\endfirsthead
\multicolumn{4}{@{}l}{\tablename\ \thetable{} continued} \\ \toprule
\textbf{Model} 
 & \textbf{Organization} 
 & \textbf{Productiveness} 
 & \textbf{Responsibility} \\
\midrule
\endhead
\midrule\multicolumn{4}{r}{{Continued on next page}} \\ \endfoot
\bottomrule\endlastfoot
Mistral-7B+Persona   & 0.58 & 0.77 & 0.83 \\
Mistral-7B+Shape     & 0.68 & 0.88 & 0.84 \\
Mistral-7B+PSI      & 0.67 & 0.77 & 0.86 \\
Gemma-2-9B+Persona         & 0.75 & 0.78 & 0.80 \\
Gemma-2-9B+Shape           & 0.80 & 0.94 & 0.87 \\
Gemma-2-9B+PSI            & 0.38 & 0.72 & 0.82 \\
Gemma-2-27B+Persona        & 0.77 & 0.84 & 0.86 \\
Gemma-2-27B+Shape          & 0.84 & 0.94 & 0.91 \\
Gemma-2-27B+PSI           & 0.70 & 0.74 & 0.85 \\
Llama3-8B+Persona    & 0.04 & 0.03 & 0.31 \\
Llama3-8B+Shape      & 0.61 & 0.60 & 0.68 \\
Llama3-8B+PSI       & 0.64 & 0.79 & 0.79 \\
Llama3-70B+Persona   & 0.80 & 0.85 & 0.88 \\
Llama3-70B+Shape     & 0.83 & 0.89 & 0.87 \\
Llama3-70B+PSI      & 0.83 & 0.77 & 0.87 \\
GPT-4o-mini+Persona            & 0.84 & 0.87 & 0.89 \\
GPT-4o-mini+Shape              & 0.67 & 0.82 & 0.79 \\
GPT-4o-mini+PSI               & 0.85 & 0.85 & 0.86 \\
GPT-4o+Persona                 & 0.73 & 0.80 & 0.67 \\
GPT-4o+Shape                   & 0.86 & 0.91 & 0.91 \\
GPT-4o+PSI                    & 0.86 & 0.87 & 0.88 \\
Human                  & 0.87 & 0.80 & 0.77 \\
\end{longtable}

\begin{longtable}{l c c c}
\caption{Cronbach’s alpha for Neuroticism facets.}%
\label{tab:alpha_neuroticism} \\
\toprule
\textbf{Model} 
 & \textbf{Anxiety} 
 & \textbf{Depression} 
 & \textbf{Emotional Volatility} \\
\midrule
\endfirsthead
\multicolumn{4}{@{}l}{\tablename\ \thetable{} continued} \\ \toprule
\textbf{Model} 
 & \textbf{Anxiety} 
 & \textbf{Depression} 
 & \textbf{Emotional Volatility} \\
\midrule
\endhead
\midrule\multicolumn{4}{r}{{Continued on next page}} \\ \endfoot
\bottomrule\endlastfoot
Mistral-7B+Persona   & 0.54 & 0.64 & 0.61 \\
Mistral-7B+Shape     & 0.64 & 0.85 & 0.79 \\
Mistral-7B+PSI      & 0.75 & 0.84 & 0.83 \\
Gemma-2-9B+Persona         & 0.85 & 0.80 & 0.80 \\
Gemma-2-9B+Shape           & 0.71 & 0.89 & 0.78 \\
Gemma-2-9B+PSI            & 0.79 & 0.84 & 0.75 \\
Gemma-2-27B+Persona        & 0.74 & 0.83 & 0.83 \\
Gemma-2-27B+Shape          & 0.61 & 0.96 & 0.90 \\
Gemma-2-27B+PSI           & 0.73 & 0.89 & 0.90 \\
Llama3-8B+Persona    & -0.17 & 0.40 & 0.19 \\
Llama3-8B+Shape      & -0.57 & 0.57 & -0.04 \\
Llama3-8B+PSI       & 0.82 & 0.83 & 0.78 \\
Llama3-70B+Persona   & 0.81 & 0.88 & 0.89 \\
Llama3-70B+Shape     & 0.80 & 0.92 & 0.88 \\
Llama3-70B+PSI      & 0.82 & 0.89 & 0.90 \\
GPT-4o-mini+Persona            & 0.76 & 0.86 & 0.84 \\
GPT-4o-mini+Shape              & 0.60 & 0.82 & 0.76 \\
GPT-4o-mini+PSI               & 0.88 & 0.90 & 0.88 \\
GPT-4o+Persona                 & 0.65 & 0.66 & 0.60 \\
GPT-4o+Shape                   & 0.81 & 0.90 & 0.87 \\
GPT-4o+PSI                    & 0.88 & 0.91 & 0.87 \\
Human                  & 0.84 & 0.86 & 0.89 \\
\end{longtable}

\begin{longtable}{l c c c}
\caption{Cronbach’s alpha for Openness facets.}%
\label{tab:alpha_openness} \\
\toprule
\textbf{Model} 
 & \textbf{Intellectual Curiosity} 
 & \textbf{Aesthetic Sensitivity} 
 & \textbf{Creative Imagination} \\
\midrule
\endfirsthead
\multicolumn{4}{@{}l}{\tablename\ \thetable{} continued} \\ \toprule
\textbf{Model} 
 & \textbf{Intellectual Curiosity} 
 & \textbf{Aesthetic Sensitivity} 
 & \textbf{Creative Imagination} \\
\midrule
\endhead
\midrule\multicolumn{4}{r}{{Continued on next page}} \\ \endfoot
\bottomrule\endlastfoot
Mistral-7B+Persona   & 0.69 & 0.60 & 0.62 \\
Mistral-7B+Shape     & 0.66 & 0.80 & 0.81 \\
Mistral-7B+PSI      & 0.52 & 0.50 & 0.62 \\
Gemma-2-9B+Persona         & 0.60 & 0.68 & 0.56 \\
Gemma-2-9B+Shape           & 0.90 & 0.92 & 0.92 \\
Gemma-2-9B+PSI            & 0.70 & 0.04 & 0.44 \\
Gemma-2-27B+Persona        & 0.75 & 0.83 & 0.80 \\
Gemma-2-27B+Shape          & 0.84 & 0.92 & 0.94 \\
Gemma-2-27B+PSI           & 0.77 & 0.86 & 0.74 \\
Llama3-8B+Persona    & 0.19 & 0.27 & 0.16 \\
Llama3-8B+Shape      & 0.54 & 0.65 & 0.51 \\
Llama3-8B+PSI       & 0.83 & 0.84 & 0.79 \\
Llama3-70B+Persona   & 0.85 & 0.83 & 0.82 \\
Llama3-70B+Shape     & 0.89 & 0.93 & 0.88 \\
Llama3-70B+PSI      & 0.80 & 0.84 & 0.77 \\
GPT-4o-mini+Persona            & 0.77 & 0.87 & 0.82 \\
GPT-4o-mini+Shape              & 0.76 & 0.88 & 0.79 \\
GPT-4o-mini+PSI               & 0.88 & 0.86 & 0.87 \\
GPT-4o+Persona                 & 0.82 & 0.87 & 0.80 \\
GPT-4o+Shape                   & 0.87 & 0.92 & 0.88 \\
GPT-4o+PSI                    & 0.89 & 0.88 & 0.88 \\
Human                  & 0.75 & 0.83 & 0.82 \\
\end{longtable}

\begin{table}[h]
\caption{Cronbach’s alpha for BFI-2 each domain.}
\label{tab:alpha_domain}
  \centering
  \scriptsize
  \resizebox{\linewidth}{!}{%
    \begin{tabular}{l *{5}{c}}
      \toprule
      \textbf{Model}
      & \textbf{Extraversion}
      & \textbf{Agreeableness}
      & \textbf{Conscientiousness}
      & \textbf{Neuroticism}
      & \textbf{Openness} \\
      \midrule
      Mistral-7B+Persona   & 0.83 & 0.83 & 0.88 & 0.79 & 0.79 \\
      Mistral-7B+Shape     & 0.93 & 0.92 & 0.93 & 0.90 & 0.91 \\
      Mistral-7B+PSI       & 0.89 & 0.92 & 0.90 & 0.91 & 0.74 \\
      Gemma-2-9B+Persona         & 0.88 & 0.91 & 0.89 & 0.91 & 0.80 \\
      Gemma-2-9B+Shape           & 0.93 & 0.98 & 0.94 & 0.92 & 0.96 \\
      Gemma-2-9B+PSI             & 0.85 & 0.88 & 0.86 & 0.91 & 0.70 \\
      Gemma-2-27B+Persona        & 0.92 & 0.93 & 0.92 & 0.90 & 0.89 \\
      Gemma-2-27B+Shape          & 0.95 & 0.98 & 0.95 & 0.94 & 0.96 \\
      Gemma-2-27B+PSI            & 0.91 & 0.93 & 0.90 & 0.90 & 0.89 \\
      Llama3-8B+Persona    & 0.59 & 0.78 & 0.55 & 0.59 & 0.55 \\
      Llama3-8B+Shape      & 0.76 & 0.89 & 0.85 & 0.59 & 0.84 \\
      Llama3-8B+PSI        & 0.90 & 0.88 & 0.88 & 0.91 & 0.90 \\
      Llama3-70B+Persona   & 0.93 & 0.94 & 0.93 & 0.93 & 0.89 \\
      Llama3-70B+Shape     & 0.91 & 0.97 & 0.94 & 0.94 & 0.96 \\
      Llama3-70B+PSI       & 0.87 & 0.93 & 0.93 & 0.93 & 0.88 \\
      GPT-4o-mini+Persona           & 0.93 & 0.93 & 0.94 & 0.91 & 0.91 \\
      GPT-4o-mini+Shape             & 0.90 & 0.95 & 0.91 & 0.89 & 0.94 \\
      GPT-4o-mini+PSI               & 0.93 & 0.94 & 0.93 & 0.94 & 0.92 \\
      GPT-4o+Persona                 & 0.90 & 0.90 & 0.88 & 0.83 & 0.91 \\
      GPT-4o+Shape                   & 0.95 & 0.98 & 0.95 & 0.94 & 0.96 \\
      GPT-4o+PSI                     & 0.93 & 0.94 & 0.94 & 0.95 & 0.92 \\
      Human                  & 0.87 & 0.85 & 0.90 & 0.94 & 0.89 \\
      \bottomrule
    \end{tabular}%
  }

\end{table}

\begin{table}[h]

  \centering
  \resizebox{\linewidth}{!}{%
    \begin{tabular}{l *{11}{c}}
      \toprule
      \textbf{Model}
      & \textbf{E–A}
      & \textbf{E–C}
      & \textbf{E–N}
      & \textbf{E–O}
      & \textbf{A–C}
      & \textbf{A–N}
      & \textbf{A–O}
      & \textbf{C–N}
      & \textbf{C–O}
      & \textbf{N–O}
      & \textbf{Mean of abs.\ values} \\
      \midrule
      Mistral-7B+Persona & 0.42 & 0.55 & –0.49 & 0.30 & 0.71 & –0.42 & 0.37 & –0.58 & 0.23 & –0.04 & 0.41 \\
      Mistral-7B+Shape   & 0.48 & 0.59 & –0.56 & 0.63 & 0.75 & –0.74 & 0.63 & –0.80 & 0.46 & –0.46 & 0.61 \\
      Mistral-7B+PSI     & 0.45 & 0.46 & –0.45 & 0.43 & 0.64 & –0.54 & 0.29 & –0.66 & 0.06 & –0.03 & 0.40 \\
      Gemma-2-9B+Persona       & 0.23 & 0.43 & –0.46 & 0.38 & 0.62 & –0.53 & 0.39 & –0.56 & 0.30 & –0.19 & 0.41 \\
      Gemma-2-9B+Shape         & 0.57 & 0.54 & –0.54 & 0.78 & 0.67 & –0.72 & 0.75 & –0.78 & 0.49 & –0.42 & 0.63 \\
      Gemma-2-9B+PSI           & 0.37 & 0.50 & –0.49 & 0.46 & 0.57 & –0.53 & 0.40 & –0.64 & 0.36 & –0.18 & 0.45 \\
      Gemma-2-27B+Persona      & 0.41 & 0.54 & –0.56 & 0.44 & 0.70 & –0.61 & 0.53 & –0.71 & 0.37 & –0.27 & 0.51 \\
      Gemma-2-27B+Shape        & 0.57 & 0.56 & –0.62 & 0.78 & 0.78 & –0.78 & 0.74 & –0.87 & 0.54 & –0.55 & 0.68 \\
      Gemma-2-27B+PSI          & 0.32 & 0.36 & –0.31 & 0.34 & 0.57 & –0.39 & 0.16 & –0.63 & 0.05 & 0.24  & 0.34 \\
      Llama3-8B+Persona  & 0.24 & 0.26 & –0.39 & 0.26 & 0.45 & –0.51 & 0.32 & –0.40 & 0.02 & –0.13 & 0.30 \\
      Llama3-8B+Shape    & 0.28 & 0.24 & –0.39 & 0.59 & 0.55 & –0.60 & 0.61 & –0.46 & 0.24 & –0.47 & 0.44 \\
      Llama3-8B+PSI      & 0.38 & 0.44 & –0.40 & 0.19 & 0.53 & –0.29 & 0.19 & –0.46 & 0.07 & 0.26  & 0.32 \\
      Llama3-70B+Persona & 0.12 & 0.33 & –0.42 & 0.28 & 0.66 & –0.54 & 0.31 & –0.60 & 0.20 & –0.08 & 0.35 \\
      Llama3-70B+Shape   & 0.41 & 0.42 & –0.57 & 0.68 & 0.70 & –0.72 & 0.78 & –0.87 & 0.48 & –0.57 & 0.62 \\
      Llama3-70B+PSI     & 0.24 & 0.31 & –0.23 & 0.35 & 0.53 & –0.40 & 0.15 & –0.54 & 0.09 & 0.20  & 0.31 \\
      GPT-4o-mini+Persona         & 0.42 & 0.54 & –0.59 & 0.40 & 0.71 & –0.64 & 0.50 & –0.66 & 0.31 & –0.21 & 0.50 \\
      GPT-4o-mini+Shape           & 0.63 & 0.50 & –0.52 & 0.75 & 0.77 & –0.71 & 0.85 & –0.72 & 0.60 & –0.48 & 0.65 \\
      GPT-4o-mini+PSI             & 0.39 & 0.40 & –0.43 & 0.23 & 0.55 & –0.34 & 0.20 & –0.53 & 0.03 & 0.26  & 0.34 \\
      GPT-4o+Persona               & 0.21 & 0.40 & –0.43 & 0.23 & 0.47 & –0.30 & 0.30 & –0.43 & 0.09 & 0.03  & 0.29 \\
      GPT-4o+Shape                 & 0.48 & 0.46 & –0.58 & 0.66 & 0.67 & –0.64 & 0.71 & –0.74 & 0.29 & –0.31 & 0.55 \\
      GPT-4o+PSI                   & 0.35 & 0.39 & –0.43 & 0.16 & 0.51 & –0.30 & 0.14 & –0.47 & 0.02 & 0.30  & 0.31 \\
      Human                & 0.17 & 0.35 & –0.45 & 0.22 & 0.33 & –0.36 & 0.20 & –0.49 & 0.09 & –0.11 & 0.28 \\
      \bottomrule
    \end{tabular}%
  }
  \caption{Domain level correlation analysis. E = Extraversion; A = Agreeableness; C = Conscientiousness; N = Neuroticism; O = Openness.}
\label{tab:dimension_correlation}
\end{table}

\begin{table*}[htbp]
\centering
\setlength{\tabcolsep}{6pt}
\renewcommand{\arraystretch}{1.1}
\begin{tabular}{l c c c c c}
\toprule
\textbf{Model + Method} & \textbf{EXT\textasciitilde}\textbf{OCB} & \textbf{AGR\textasciitilde}\textbf{OCB} & \textbf{NEU\textasciitilde}\textbf{OCB} & \textbf{CON\textasciitilde}\textbf{OCB} & \textbf{OPE\textasciitilde}\textbf{OCB} \\
\midrule
Mistral-7B+PSI   & 0.58 (\textbf{+0.16}) & 0.63 (\textbf{+0.45}) & -0.44 (\underline{-0.23}) & 0.60 (\textbf{+0.48}) & 0.25 (\textbf{+0.08}) \\
Gemma-2-9B+PSI   & 0.59 (\textbf{+0.17}) & 0.60 (\textbf{+0.42}) & -0.52 (\underline{-0.31}) & 0.58 (\textbf{+0.46}) & 0.35 (\textbf{+0.18}) \\
Gemma-2-27B+PSI  & 0.53 (\textbf{+0.11}) & 0.63 (\textbf{+0.45}) & -0.48 (\underline{-0.27}) & 0.62 (\textbf{+0.50}) & 0.17 (+0.00) \\
Llama3-8B+PSI    & 0.59 (\textbf{+0.17}) & 0.54 (\textbf{+0.36}) & -0.32 (\underline{-0.11}) & 0.55 (\textbf{+0.43}) & 0.15 (\underline{-0.02}) \\
Llama3-70B+PSI   & 0.48 (\textbf{+0.06}) & 0.46 (\textbf{+0.28}) & -0.38 (\underline{-0.17}) & 0.47 (\textbf{+0.35}) & 0.11 (\underline{-0.06}) \\
GPT4o-mini+PSI   & 0.56 (\textbf{+0.14}) & 0.43 (\textbf{+0.25}) & -0.40 (\underline{-0.19}) & 0.52 (\textbf{+0.40}) & 0.08 (\underline{-0.09}) \\
GPT4o+PSI        & 0.54 (\textbf{+0.12}) & 0.36 (\textbf{+0.18}) & -0.35 (\underline{-0.14}) & 0.45 (\textbf{+0.33}) & 0.02 (\underline{-0.15}) \\
\midrule
Human             & 0.42 & 0.18 & -0.21 & 0.12 & 0.17 \\
\bottomrule
\end{tabular}
\caption{OCB correlation with five personality domains. Values in parentheses show the difference from human, \textbf{bold} for positive, \underline{underline} for negative.}
\label{tab:ocb}
\end{table*}

\begin{table*}[htbp]
\centering
\setlength{\tabcolsep}{6pt}
\renewcommand{\arraystretch}{1.1}
\begin{tabular}{l c c c c c}
\toprule
\textbf{Model + Method} & \textbf{EXT\textasciitilde}\textbf{CWB} & \textbf{AGR\textasciitilde}\textbf{CWB} & \textbf{NEU\textasciitilde}\textbf{CWB} & \textbf{CON\textasciitilde}\textbf{CWB} & \textbf{OPE\textasciitilde}\textbf{CWB} \\
\midrule
Mistral-7B+PSI   & -0.05 (\underline{-0.04}) & -0.29 (\textbf{+0.01}) & 0.26 (\textbf{+0.03}) & -0.27 (\textbf{+0.08}) & -0.07 (\textbf{+0.10}) \\
Gemma-2-9B+PSI   & -0.04 (\underline{-0.03}) & -0.58 (\underline{-0.28}) & 0.45 (\textbf{+0.22}) & -0.45 (\underline{-0.10}) & -0.11 (\textbf{+0.06}) \\
Gemma-2-27B+PSI  & -0.02 (\underline{-0.01}) & -0.49 (\underline{-0.19}) & 0.40 (\textbf{+0.17}) & -0.48 (\underline{-0.13}) & 0.10 (\textbf{+0.27}) \\
Llama3-8B+PSI    & 0.04 (\textbf{+0.05}) & -0.20 (\textbf{+0.10}) & 0.07 (\underline{-0.16}) & -0.18 (\textbf{+0.17}) & 0.00 (\textbf{+0.17}) \\
Llama3-70B+PSI   & 0.05 (\textbf{+0.06}) & -0.45 (\underline{-0.15}) & 0.33 (\textbf{+0.10}) & -0.41 (\underline{-0.06}) & 0.00 (\textbf{+0.17}) \\
GPT4o-mini+PSI   & -0.21 (\underline{-0.20}) & -0.56 (\underline{-0.26}) & 0.43 (\textbf{+0.20}) & -0.52 (\underline{-0.17}) & 0.04 (\textbf{+0.21}) \\
GPT4o+PSI        & -0.08 (\underline{-0.07}) & -0.54 (\underline{-0.24}) & 0.36 (\textbf{+0.13}) & -0.47 (\underline{-0.12}) & 0.01 (\textbf{+0.18}) \\
\midrule
Human             & -0.01 & -0.30 & 0.23 & -0.35 & -0.17 \\
\bottomrule
\end{tabular}
\caption{CWB correlation with five personality domains. Values in parentheses show the difference from human, \textbf{bold} for positive, \underline{underline} for negative.}
\label{tab:cwb}
\end{table*}

\end{document}